\documentclass{article}

\usepackage{microtype}
\usepackage{graphicx}
\usepackage{subcaption}
\usepackage{booktabs} 

\usepackage[preprint]{icml2026}

\usepackage[utf8]{inputenc} 
\usepackage[T1]{fontenc}    
\usepackage{url}            
\usepackage{amsfonts}       
\usepackage{nicefrac}       
\usepackage{enumitem}

\usepackage{bm}
\usepackage{algorithm} 
\usepackage{algorithmicx}
\usepackage[noend]{algpseudocode} 

\usepackage{multirow}
\usepackage{adjustbox} 
\usepackage[hidelinks]{hyperref}
\usepackage{xcolor}
\usepackage{bbold}
\definecolor{MyDarkGreen}{rgb}{0.0, 0.5, 0.0}
\usepackage{amsmath}
\usepackage{amssymb}
\usepackage{mathtools}
\usepackage{amsthm}

\theoremstyle{plain}
\newtheorem{theorem}{Theorem}[section]

\newtheorem{corollary}[theorem]{Corollary}
\theoremstyle{definition}
\newtheorem{definition}[theorem]{Definition}

\theoremstyle{remark}

\icmltitlerunning{Early-Exit Graph Neural Networks}

\begin{document}

\twocolumn[
  \icmltitle{Early-Exit Graph Neural Networks}



  \icmlsetsymbol{equal}{*}

  \begin{icmlauthorlist}
    \icmlauthor{Andrea Giuseppe Di Francesco}{equal,yyy,pisa}
    \icmlauthor{Maria Sofia Bucarelli}{yyy,sch}
    \icmlauthor{Franco Maria Nardini}{pisa}
    \icmlauthor{Raffaele Perego}{pisa}
    \icmlauthor{Nicola Tonellotto}{unipi}
    \icmlauthor{Fabrizio Silvestri}{yyy}
  \end{icmlauthorlist}

  \icmlaffiliation{yyy}{Department of Computer Science, Control and Management Engineering, Sapienza University of Rome, Rome, Italy}
  \icmlaffiliation{pisa}{Institute of Information Science and Technologies "Alessandro Faedo" - ISTI-CNR, Pisa, Italy}
  \icmlaffiliation{unipi}{Information Engineering Department, University of Pisa, Pisa, Italy}
  \icmlaffiliation{sch}{CNRS, i3S, Inria}
  \icmlcorrespondingauthor{Andrea Giuseppe Di Francesco}{difrancesco@diag.uniroma1.it}

  \icmlkeywords{Machine Learning}

]



\printAffiliationsAndNotice{}  

\begin{abstract}
Early-exit mechanisms allow deep neural networks to stop inference once prediction confidence is high, reducing latency and energy on easy inputs while retaining full-depth accuracy on harder ones. Similarly, adding early exit mechanisms to Graph Neural Networks (GNNs), the go-to models for graph-structured data, allows for dynamic trading depth for confidence on simple graphs while maintaining full-depth accuracy on harder ones to capture intricate relationships. 
Yet, their potential in deep GNNs, where over-smoothing, over-squashing or more generally vanishing gradients prevent these model to properly learn, remains largely unexplored. To address this, we introduce \textit{Symmetric-Anti-Symmetric GNNs} (SAS-GNN), whose symmetry-based inductive biases yield stable intermediate representations that support safe early exits.
Building on this backbone, we propose Early-Exit GNNs (EEGNNs), which attach confidence-aware exit neural heads which are trainable end-to-end based on the task objective, enabling on-the-fly termination at node or graph level. Experiments show that EEGNNs learn task-driven exit strategies, while achieving competitive results on heterophilic graphs and long-range tasks. Even when not outperforming the strongest baselines, EEGNNs consistently deliver favorable accuracy–efficiency trade-offs thanks to their adaptive and parameter-efficient design. We plan to release the code to reproduce our experiments.
\end{abstract}

\section{Introduction}
Deep learning models are increasingly deployed in latency and energy-constrained settings (e.g., mobile AR, autonomous drones, real-time
recommendation).
Graph Neural Networks (GNNs) inherit these constraints because their message-passing depth directly translates into runtime and energy costs.
In such scenarios, \emph{adapting computational effort to input difficulty} is critical for both efficiency and sustainability.
GNNs process graph-structured data across domains such as text, images, knowledge graphs, and social networks~\citep{NMT_survey, recsystem1, KBcompletion}, excelling in tasks like node/graph classification~\citep{WL1} and link prediction~\citep{linkprediction}.
Most GNNs are Message-Passing Neural Networks (MPNNs)~\citep{messagepassing}, where each layer updates node states by aggregating neighbor features via convolutions, attention, or learned functions~\citep{everythingisconnected}. The number of layers thus determines how far information can propagate: deeper models capture broader context but incur higher cost.
Intuitively, increasing the number of layers should allow for better integration of long-range information, enabling messages to traverse farther across the graph and enriching each node’s representation with broader structural context.
In practice, however, depth is a double-edged sword: on the one hand, too many layers cause \emph{over-smoothing} \citep{oversmoothing2}, where node embeddings become indistinguishable, and \emph{over-squashing}, where information from distant nodes is severely compressed due to topological or computational bottlenecks~\citep{underreaching, oversquashing1, demystifyingcommonbeliefs}, both hindering the success of MPNNs on tasks requiring long-range propagation; on the other hand, too few layers result in \emph{under-reaching}, where messages fail to cover the task-specific \emph{problem radius}~\citep{underreaching}.
Because the problem radius cannot be known \emph{a priori}, the layer count becomes a delicate hyperparameter that should be set to the shallowest depth still able to span the necessary receptive field while keeping model size and training cost as low as possible.

\begin{figure}[t]
    \centering    
\includegraphics[width=\linewidth]{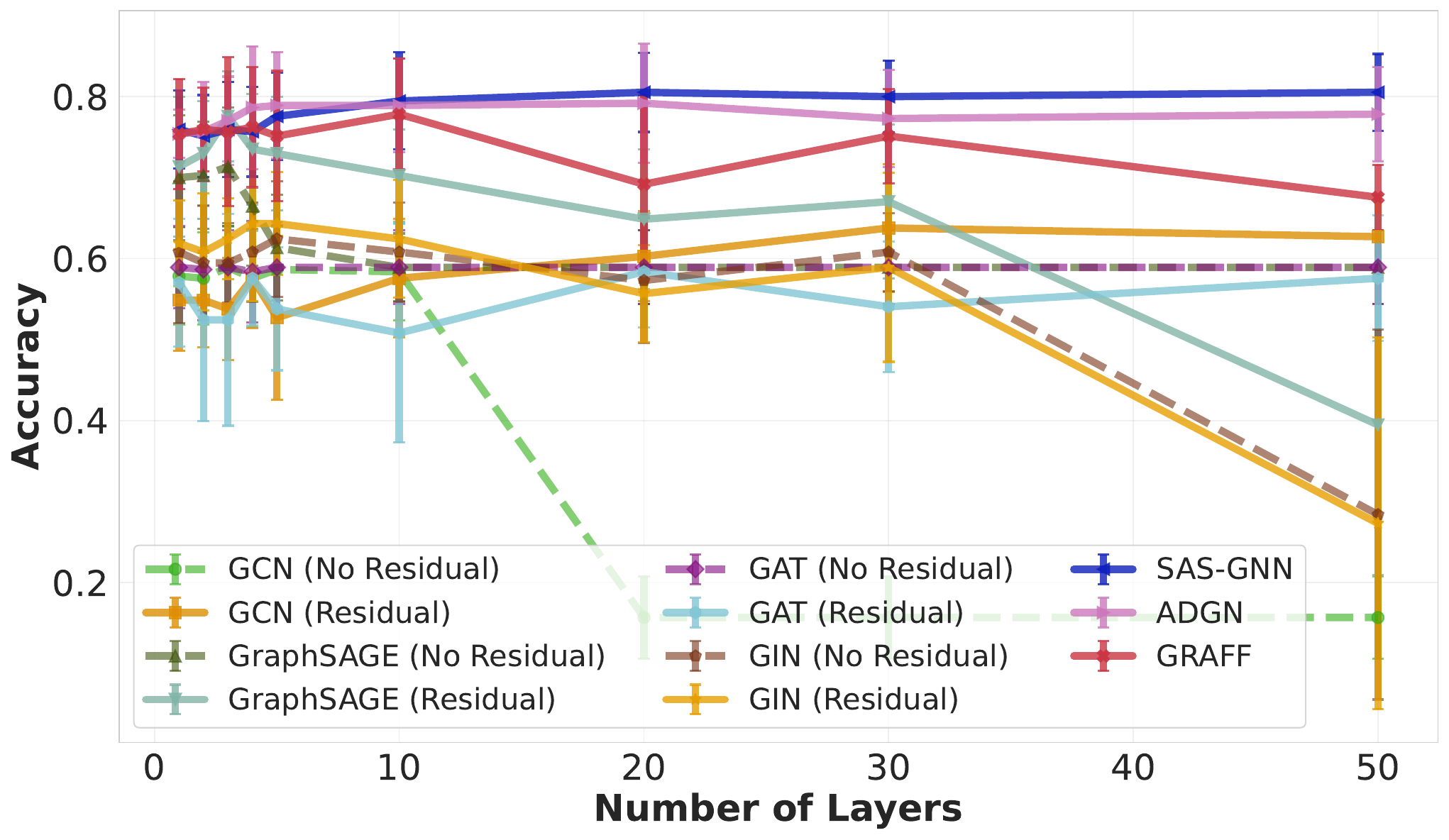}
\caption{Classic MPNNs + residual design is not enough for going deep. \texttt{Texas} Dataset.}
\label{fig:depth_performance_degrade}
\end{figure}

\noindent
\textbf{Our idea}.
Rather than fixing a single “best” depth, we allow GNNs to \emph{decide on-the-fly}: each node or the entire graph can halt message passing once its prediction is sufficiently confident. However, as shown in Figure~\ref{fig:depth_performance_degrade}, classic MPNNs do not reliably scale in depth due to the above issues, even in the presence of residual connections~\citep{residualconnectionsoversmoothing}. As a consequence, early-exit mechanisms cannot be straightforwardly applied to regimes that inherently require deep GNNs, such as long-range reasoning, unless we propose specialized approaches. When depth itself is unreliable, early exits merely act as a safeguard, dedicating time to easy inputs, but ignoring harder samples, which contrasts with the core purpose of early-exit \cite{EEsurvey}. This limitation was already evident in the seminal work of \citet{AdaProp}, where the exit mechanism was learned via maximum budget loss function. Reliable early exits require intermediate representations that remain informative across layers, so that the model can safely continue message passing when it is not confident yet. However, this requirement is not met by current methodologies: the trade-off between exiting early and processing features deeply enough is fundamentally unstable. In fact, we show that existing early-exit, when rely on budget-aware training objectives bias the model toward premature halting focusing more on computational constraints rather than predictive correctness. In contrast to this, we believe that a reliable early-exit mechanism must be agnostic to any predefined budget, and allow for exit coherently with the task objective. To achieve this, we design \emph{Symmetric–Anti-Symmetric GNNs} (SAS-GNNs), whose weight-shared, ODE-inspired message passing yields informative embeddings and constant memory necessary to deploy EEGNN, an early exit model which learns to halt propagation based on the task.
\paragraph{Contributions.}
\begin{enumerate}
\item \textbf{EEGNN}. The first end-to-end \emph{early-exit} GNN: nodes or graphs halt when confident via Gumbel–Softmax heads enabling end-to-end training and removing the necessity for depth or budget tuning.
\item \textbf{SAS-GNN backbone}. A weight-shared, symmetry/anti-symmetry MPNN (ODE-inspired) that provides stable intermediate states for safe early exits with constant memory.
\item \textbf{Theory}. We prove SAS-GNN preserves node information while inducing adaptive attraction/repulsion edge-wise, resulting as a good proxy for deep feature processing and supporting long-range tasks.
\item \textbf{Results}. Extensive experiments on heterophilic and long-range benchmarks demonstrate that EEGNN and SAS-GNN match or surpass complex Attention-based and Asynchronous MPNNs. Notably, they achieve this with significantly fewer parameters, no normalization or dropout, and contained latency, while successfully adapting exit strategies to the specific demands of each task.
\end{enumerate}

\section{Related Work}
\label{sec:related_works}
\textbf{Early-Exit for GNNs.} Our approach shares similarities with \citet{AdaProp}, which introduces the idea of adaptive propagation node-wise for GNNs, allowing nodes to halt their updates during message passing.

While their method discuss depth-related issues like over-smoothing (OST), no solution for designing deep architectures and dealing with long-range tasks has been proposed. Additionally, their design requires auxiliary loss terms to enable differentiable node-level exit decisions. 

Other early-exit methods for GNNs~\citep{learningtopropagatemess, turningacurse, ADMP-GNN} are restricted to node-level tasks and do not investigate deep regimes or OST/OSQ. In contrast, we propose the first fully differentiable mechanism that naturally addresses deep MPNN flaws without auxiliary supervision. Furthermore, we are the first to extend this capability to graph classification and regression tasks.
 


\textbf{Asynchronous / Transformer GNNs}.
Graph Transformers~\citep{GT1} capture long-range dependencies via global attention but incur quadratic computational costs. Asynchronous MPNNs~\citep{coognns, AdaptiveMP} attempt to reduce this cost by adapting topology at each layer at the cost of an increased architectural complexity w.r.t. classic MPNNs. Unlike these approaches, our model adapts depth dynamically at both train and test time without altering the graph topology, sharing the complexity with classic MPNNs.

\textbf{Neural ODE methods}.
Graph Neural ODEs (GraphNODEs)~\citep{GraphNODE} model feature propagation as a continuous-time dynamical system, providing a rigorous framework to analyze stability. While initially used to address over-smoothing (OST)~\citep{graphcon, GRAFF2}, recent works leverage physics-inspired architectures to mitigate over-squashing (OSQ) and enhance long-range propagation~\citep{ADGN, swan, trenta2025sonar}.

In this work, we draw inspiration from prior approaches addressing both OST and OSQ, and we propose EEGNN, the first GraphNODE who natively supports an early-exit mechanism.
We provide an extended related work discussion in Appendix \ref{sec:extended_related_works}.
\looseness=-1 

\section{Methodology}
\label{sec:method}
We use bold fonts for both matrices and vectors, with uppercase letters representing matrices and lowercase letters representing vectors (e.g., $\mathbf{M}$, $\mathbf{v}$). Scalars are denoted by italic letters (e.g., \textit{s}).

\noindent \textbf{Notation}. Let $\mathcal{G} = (\mathcal{V}, \mathcal{E}, \mathbf{X})$ be an undirected graph, where $\mathcal{V}$ is the set of nodes, $\mathcal{E} \subseteq \mathcal{V} \times \mathcal{V}$ is the set of edges and $\mathbf{X} \in \mathbb{R}^{n \times m}$ is the instance matrix containing node features  representations. The $u$-$th$ row of the instance matrix is represented by $\mathbf{x}_u \in \mathbb{R}^m$.
$\mathcal{G}$ has $n=|\mathcal{V}|$ nodes and $|\mathcal{E}|$ edges.
 $\Gamma(u)$ represents the neighborhood of node $u$, and $|\Gamma(u)|$ denotes its degree. 
 The degree matrix $\mathbf{D} \in \mathbb{R}^{n\times n}$ has diagonal entries $D_{uu}=|\Gamma(u)|$. 
The set of edges $\mathcal{E}$ can also be expressed as the adjacency matrix $\mathbf{A} \in \{0, 1\}^{n \times n}$, where $A_{uv} = 1$ if $(u, v) \in \mathcal{E}$, and $A_{uv} = 0$ otherwise.

\noindent \textbf{Graph Neural Networks}. 
Most GNNs follow the message-passing paradigm~\citep{messagepassing}, where node features are iteratively updated.
 Let $\mathbf{H}^0=\mathbf{X}$, or $\mathbf{H}^0=f(\mathbf{X})$ when using a learnable projection (e.g., an MLP). At layer $l$, the hidden matrix is $\mathbf{H}^l$ with node embeddings $\mathbf{h}_u^l \in \mathbb{R}^{m'}$, and the process continues up to $L$, conditioned on $\mathbf{A}$. The final representations $\mathbf{H}^L=\mathbf{Z}\in\mathbb{R}^{n\times m'}$ are used for downstream tasks. 
For node classification, $\hat y_v=g(\mathbf{z}_v)$, and for graph classification, $\hat y=g(\text{Pool}(\mathbf{Z}))$, where $\text{Pool}$ is a permutation-invariant operator (e.g., mean, max).
For datasets $\mathcal{D}={\mathcal{G}_i}$ with graphs $\mathcal{G}_i=(\mathcal{V}_i,\mathcal{E}_i,\mathbf{X}_i)$ and adjacency $\mathbf{A}_i$, we analogously define $\mathbf{H}^l_i$. We use single-graph notation when possible. 
We define a task as transductive when it is performed on a single graph, while the task is inductive when datasets contain more than one graph.

\noindent \textbf{Our approach}. A naïve scheme would let every node draw an $\arg\max$ over the two actions—\textit{exit} or \textit{continue}—at each layer and stop if it chooses the first. Since the hard $\arg\max$ is non-differentiable, the exit policy cannot be learned with standard back-propagation. To retain end-to-end differentiability, we need a \emph{soft}, trainable substitute for the discrete decision.
This requirement leads to two concrete design goals: 
\begin{enumerate}[label=O\arabic*.]
    \item \textbf{Stable backbone.} Design the message-passing backbone so that hidden node features stay \emph{stable and distinct} as layers accumulate, i.e., they neither blow up nor collapse into identical vectors.
    With useful information preserved at every depth, any layer can serve as a trustworthy early-exit point.
    \item \textbf{Contextual exit policy.} Equip each layer with a differentiable confidence head (implemented here with the Gumbel–Softmax trick) that decides, on the fly, whether a node or the whole graph has gathered enough evidence to stop.
\end{enumerate}
\textit{Because we want early-exit heads to act on reliable hidden states, O1 is a prerequisite for O2.}

\noindent \textbf{Symmetric-Anti-Symmetric Graph Neural Network (O1)}.
To accomplish \textbf{O1}, we build upon two message-passing schemes proposed in recent years. One is the \textit{Anti-Symmetric Deep Graph Network} (A-DGN) \citep{ADGN}, and the other is the \textit{Gradient Flow Framework} (GRAFF) \citep{GRAFF2}. A-DGN preserves long-range information by using antisymmetric learnable matrices, yielding a  \emph{stable} and \emph{non-dissipative} dynamical system where gradients remain well-conditioned across layers, resulting into a good proxy for OSQ.
GRAFF, instead, was introduced to deal with node classification in heterophilic graphs\footnote{A graph is heterophilic when adjacent nodes tend to share different class labels.}. Since OST causes nodes to converge to the same representation, its negative effect is enhanced when learning in heterophilic graphs.  GRAFF takes advantage of symmetric learnable matrices, that induce attraction and repulsion edge-wise to prevent adjacent nodes from becoming similar in the limit of many layers. Further discussions on GRAFF and A-DGN are available in Appendix~\ref{sec:extended_preliminaries}.
\looseness=-1 

Building on these ideas, we design the Symmetric–Anti-Symmetric GNN (SAS-GNN), where antisymmetric matrices preserve long-range dependencies and symmetric ones enforce discriminability across classes.
\looseness=-1
Its message-passing rule is 
\begin{equation}
\label{eq:sas-gnn_derivative}
    \dot{\mathbf{H}}^{t} = \sigma_1(- \sigma_2(\mathbf{H}^t \mathbf{\Omega}_{as}) +  \mathbf{\bar{A}} \mathbf{H}^t\mathbf{W}_s),
\end{equation}
$\mathbf{\bar{A}} = \mathbf{D}^{-\frac{1}{2}}\mathbf{A}\mathbf{D}^{-\frac{1}{2}}$ is the normalized adjacency matrix,  $\sigma_1, \sigma_2$ are non-linear activation functions, and finally $\mathbf{\Omega}_{as}$, $\mathbf{W}_s \in \mathbb{R}^{m' \times m'}$ are the antisymmetric and symmetric trainable weight matrices. We formalized Equation \eqref{eq:sas-gnn_derivative} as an ODE since we build upon the Graph Neural ODE framework. We integrate it via the Euler discretization as:
\begin{equation}
 \mathbf{H}^{t+\tau} = \mathbf{H}^t + \tau \sigma_1(- \sigma_2(\mathbf{H}^t \mathbf{\Omega}_{as}) +  \mathbf{\bar{A}} \mathbf{H}^t_i\mathbf{W}_s)   
\end{equation}
Here, \( \tau \) is the integration step. Weight matrices are shared across layers, as specified in GRAFF. A-DGN performs similarly with or without weight sharing~\citep{ADGN}, but we adopt it for space efficiency when scaling to many layers (\( t \rightarrow \infty \)).\\ This design is supported by the following theorems.  
\begin{theorem}
\label{theo:stability}
Let us assume that the node features $\mathbf{H}^t$ evolve according to Equation \eqref{eq:sas-gnn_derivative}.
if $\bar{\mathbf{A}}$ does not contain self-loops, and the derivative of $\sigma_1$ is bounded, then the evolution of $\mathbf{H}^t$ is stable and non-dissipative. 
\end{theorem}
\textit{Proof sketch}. This result follows from a standard stability analysis of the ODE in Equation~\eqref{eq:sas-gnn_derivative}. The antisymmetric term $-\sigma_2(\mathbf{H}^t \boldsymbol{\Omega}_{\mathrm{as}})$ contributes Jacobian eigenvalues with purely imaginary components, while the symmetric term involving $\mathbf{W}_s$ does not increase their real parts. 

We can interpret the dynamics of SAS-GNN through an energy functional 
\begin{equation}
\label{eq:energy_functional}
    E_{\theta}(\mathbf{H}^t) = - \sum_{i,j} \frac{1}{ \sqrt{d_i \cdot d_j}} \langle \mathbf{h}_i^t, \mathbf{W}_{s} \mathbf{h}_j^t \rangle.
\end{equation}
The functional in Eq.~\eqref{eq:energy_functional} naturally encodes both attractive and repulsive forces between adjacent nodes through the symmetric weights 
$W_s$. Minimization therefore drives representations to respect structural patterns such as homophily (attraction) or heterophily (repulsion).
More formally the following theorem holds.
\begin{theorem}
\label{theo:energy}
Let us assume that the node features $\mathbf{H}^t$ evolve according to Equation \eqref{eq:sas-gnn_derivative}.
Assuming that $\sigma_1$, and $\sigma_2$ are defined s.t. $\forall x \in \mathbb{R}$, $\sigma_1(x), \sigma_2(x) \geq 0$. The evolution of $\mathbf{H}^t$ minimizes a parameterized functional $E_{\theta}(\mathbf{H}^t)$, inducing attraction or repulsion among adjacent nodes.
\end{theorem}
The proof can be found in Appendix \ref{sec:proofs}.
\begin{corollary}
\label{corr:sigma}
Let \( \sigma_1(x) = \text{ReLU}(\tanh(x)) \) and \( \sigma_2(x) = \text{ReLU}(x) \), both of which are non-negative functions. Given that the derivative of \( \sigma_1(x) \) is bounded, the evolution of \( \mathbf{H}^t \) is stable and non-dissipative. Furthermore, this evolution minimizes a parameterized energy functional \( E_{\theta}(\mathbf{H}^t) \), inducing attraction and repulsion among adjacent nodes.
\end{corollary}
\textit{Proof sketch.} The proof, follows directly from Theorems~\ref{theo:stability} and~\ref{theo:energy}, as ReLU+TanH and ReLU satisfy the required boundedness and non-negativity conditions. 

We adopt this activation pair in all experiments, and show in Appendices~\ref{sec:extended_nonlinearities} and~\ref{sec:extended_EE} that it affects the model performances differently w.r.t. other common activation functions.\\
Since edge features can also be included to improve graph learning~\cite{GINE}, we also propose a SAS-GNN version that encompasses their use $\mathbf{E} \in \mathbb{R}^{|\mathcal{E}| \times d}$ as follows.
\begin{equation}
\label{eq:edge_features}
    \dot{\mathbf{H}}^{t} = \sigma_1(- \sigma_2(\mathbf{H}^t \mathbf{\Omega}_{as}) + f_e(\mathbf{E}) +  \mathbf{\bar{A}} \mathbf{H}^t\mathbf{W}_s)
\end{equation}
\citet{GINE} implements $f_e(\mathbf{E}) = \mathbf{B}\mathbf{E}\mathbf{W}_e$, where $\mathbf{B} \in \mathbb{R}^{n \times |\mathcal{E}|}$ is the node-edges incidence matrix, s.t. $B_{i,(u,v)} = 1$ if $i = u \vee i = v$, otherwise  $B_{i,(u,v)} = 0$, and $\mathbf{W}_e \in \mathbb{R}^{d \times m'}$ is a learnable weight matrix. In this work, we propose $f_e(\mathbf{E})\equiv -ReLU(\mathbf{B}\mathbf{E}\mathbf{W}_e)$. Since $\mathbf{E}$ is not dependent on $t$ or the node features, we can state the following theorem.
\begin{theorem}
\label{theo:edge_features}
    Let us assume that the node features $\mathbf{H}^t$ evolve according to equation \eqref{eq:edge_features}. If $\sigma_1(x) = \text{ReLU}(\text{tanh}(x))$, $\sigma_2(x) = \text{ReLU}(x)$, $f_e(\mathbf{E})\equiv -ReLU(\mathbf{B}\mathbf{E}\mathbf{W}_e)$, and $\bar{\mathbf{A}}$ does not contain self-loops, then the evolution of $\mathbf{H}^t$ is stable and non-dissipative and minimizes a parameterized energy functional $E_{\theta}(\mathbf{H}^t)$, inducing attraction or repulsion among adjacent nodes. 
\end{theorem}
This theorem can be proved analogously to the previous ones, with full proofs provided in Appendix~\ref{sec:proofs}. Ablation studies of SAS-GNN with and without edge features are available in Appendix \ref{sec:extended_LRGB}. Empirical validation of SAS-GNN, using metrics for over-smoothing, over-squashing, and performance on long-range tasks and highly heterophilic datasets, is presented in Appendix~\ref{sec:theory_results}.
\begin{algorithm}[t]
\caption{Neural Adaptive-Step Early-Exit GNNs} 
\label{alg:eegnn}
\begin{algorithmic}[1]
\State Initialize $\mathbf{H}^{0}$, $L$, $\mathbf{\bar{A}}$, $f_e$, $f_c$, $f_{\nu}$, $\mathbf{W}_{s}$, $\mathbf{\Omega}_{as}$, $\sigma_1$, $\sigma_2$, $\nu_0$, $\mathbf{Z} = \mathbf{0}_{n \times d}$, $exit\_list = \{\}$

\For{$l=0$ \textbf{to} $L$}
    \State $\mathbf{C}^l \gets f_c(\mathbf{H}^l, \mathbf{\bar{A}})$; $\bm{\nu}^l \gets f_{\nu}(\mathbf{H}^l, \mathbf{\bar{A}}, \nu_0)$
    \State $\mathbf{c}^l \gets gumbel\_softmax(\mathbf{C}^l, \bm{\nu}^l)$
    
    \State $\bm{\tau}^l \gets \mathbf{c}^l(\texttt{0})$
    \State $\Delta \mathbf{H}^l \gets \sigma_1(- \sigma_2(\mathbf{H}^l \mathbf{\Omega}_{as})+ f_e(\mathbf{E})+\mathbf{\bar{A}} \mathbf{H}^l \mathbf{W}_s)$
    \State $\mathbf{H}^{l+1} \gets  \mathbf{H}^l + \bm{\tau}^l \Delta \mathbf{H}^l$
    \label{alg:message_passing_line}
    \For{$i=0$ \textbf{to} $n$}
        \If{$argmax\{\mathbf{c}^l\} = \texttt{1} \wedge i \notin exit\_list$}
            \State $\mathbf{Z}_i \gets \mathbf{h}_i^l$; $exit\_list.add(i)$
        \EndIf
    \EndFor
\EndFor
\For{$i=0$ \textbf{to} $n$}
    \If{$i \notin exit\_list$}
        \State $\mathbf{Z}_i \gets \mathbf{h}_i^L$;$exit\_list.add(i)$
    \EndIf
\EndFor
\State \textbf{return} $\mathbf{Z}$
\end{algorithmic}
\end{algorithm}
\\

\noindent \textbf{Gumbel Softmax Early-Exit Mechanism (O2)}.
Having addressed \textbf{O1}, we now turn to \textbf{O2}: implementing a contextual early-exit mechanism. We first focus on node classification; the extension to graph classification is presented in Appendix~\ref{sec:extended_algorithm}. 
Taking inspiration from \citet{coognns},  we employ the straight-through Gumbel-Softmax estimator~\citep{gumbel_softmax1, gumbel_softmax2}, which provides a differentiable relaxation of discrete sampling. 
Let $\Omega$ be the action space, $\Omega=\{\texttt{0},\texttt{1}\}$ that corresponds to \textit{continue} (\texttt{0}) or \textit{exit} (\texttt{1}). For each node, a confidence vector $\mathbf{C}\in\mathbb{R}^{|\Omega|}$ defines action probabilities (e.g., $\mathbf{C}(\texttt{1})$ is the exit probability).
The Gumbel-Softmax estimator approximates the categorical distribution \( \mathbf{C} \) using a Gumbel-distributed vector \( \mathbf{g} \in \mathbb{R}^{|\Omega|} \), where each component \( g(a) \sim \text{GUMBEL}(0, 1) \) for \( a \in \{\texttt{0}, \texttt{1}\} \). 
For a node $i$, given its confidence vector \( \mathbf{C}_i \) and a temperature parameter \( \nu_i \), the Gumbel-Softmax score is computed as:
\looseness=-1
\[
\mathbf{c}_i(\mathbf{C}_i; \nu_i) = \text{Softmax}\left( \frac{\log(\mathbf{C}_i) + \mathbf{g}_i}{\nu_i} \right)
\]
which approaches a one-hot encoding as $\nu_i \to 0$.

We compute $\mathbf{C}^t\in\mathbb{R}^{n\times|\Omega|}$ and $\bm{\nu}^t\in\mathbb{R}^{n\times 1}$ from hidden states $\mathbf{H}^t$ using two GNN modules, $f_c$ and $f_\nu$, with fixed depth $L_f$ and hidden size $m_f$. These are shared across layers for efficiency. To our knowledge, this is the first application of the Gumbel-Softmax reparametrization to early exit in GNNs. Importantly, $f_c$ and $f_\nu$ are trained end-to-end with only the task loss (e.g., cross-entropy, MSE), so exit decisions depend purely on task-driven context without auxiliary supervision.
Additional details on the Gumbel-Softmax distribution and implementation of \( f_c \) and \( f_{\nu} \) are provided in Appendix~\ref{sec:gumbel_softmax}.

\noindent \textbf{Early-Exit Graph Neural Networks}.
As shown, \textsc{SAS-GNN} satisfies Goal O1; combining it with the Gumbel-Softmax early-exit mechanism (O2) yields our full framework, \textsc{EEGNNs}, presented in Algorithm~\ref{alg:eegnn}. The algorithm is presented for node classification but extends naturally to inductive and graph-level tasks (see Appendix~\ref{sec:extended_algorithm}).
To integrate our Early-Exit mechanism into SAS-GNN, we need to include the Gumbel-Softmax scores in the message-passing update. 
Unlike \citet{coognns}, who modify the topology (and backpropagate through $\mathbf{\bar{A}}$), we adapt the integration constant $\tau$, making it node- and layer-dependent. We introduce so the \textit{Neural Adaptive-step}: we set $\bm{\tau}^l=\mathbf{c}^l(\texttt{0})$, the non-exit probability. When $\mathbf{c}^l(\texttt{0})\to 0$, the update reduces to $\mathbf{H}^{l+1}\leftarrow\mathbf{H}^l$, naturally encoding the exit bias. Nodes predicted to exit are stored at step $l$, and any remaining nodes are output at depth $L$.
\looseness=-1

Thus, each node has a personalized trajectory: its total ``time in the network” is $\sum_{l=0}^L \tau^l_u$, giving a continuous view of exit rather than discrete steps (visualized in Appendix~\ref{sec:node_distributions}). Practically, $L$ acts as the maximum number of exit points. Setting $L$ high removes the need for manual depth tuning and enables long-range reasoning and deep feature processing, as long as intermediate features remain informative; setting it within hardware limits guarantees inference never exceeds the cost of a non-exiting model. Weight sharing ensures memory remains constant regardless of $L$.

To summarize, using SAS-GNN in line \ref{alg:message_passing_line} of the algorithm offers two key benefits: (1) it provides a fallback that mitigates message-passing failures at any depth; and (2) the weight sharing is a byproduct of the Neural ODE design, which is an advantage for space complexity, and also avoids parameter waste from unused layers.
In Section \ref{sec:extended_graphclass}, we discuss the connection of our Neural Adaptive Step, with Adaptive Step in Runge-Kutta solvers \citep{DOPRISOLVER}, and also the mitigation of underreaching.


\textbf{Discussion on Complexity}. We analyze the space complexity in terms of parameter count, comparing MPNNs, GTs, and Co-GNNs with our models. Thanks to weight sharing, SAS-GNN maintains constant complexity w.r.t. \( L \) and quadratic complexity in the hidden dimension \( m' \) with symmetry and antisymmetry halving the effective number of parameters.
EEGNN adds parameters via \( f_c \) and \( f_{\nu} \), but this overhead depends only on \( L_f \), not \( L \), since these modules are shared across layers. In the long-range experiments, we implemented \( f_c \) and \( f_{\nu} \) as additional SAS-GNNs, so there the overall complexity becomes \( \mathcal{O}(m'^2 + m_f^2) \).
Table~\ref{tab:space_complexities} also includes Polynormer, a GT-based model that uses two modules: local attention (with \( L_l \) layers) and global attention (with \( L_g \) layers), each using 4 non-shared weight matrices of size \( m'^2 \). Assuming \( L_l = L_g = L \), the total complexity becomes \( \mathcal{O}(8Lm'^2) \) as shown in table.
Regarding time complexity, SAS-GNN offers no clear runtime advantage, but EEGNN may reduce computation by exiting early for many nodes. Although some nodes may still require full-depth processing, potentially becoming bottlenecks, this is not always the case, as explored in Section~\ref{sec:impact_earlyexit}.
\begin{table*}[t]
    \centering
    \caption{Space complexity comparison among models.}
    \resizebox{0.75\textwidth}{!}{
    \begin{tabular}{l c c c c c} \hline
        Models & GCN & SAS-GNN & Co-GNN & EEGNN & Polynormer\\ \hline
        $\mathcal{O}(\cdot)$ & $\mathcal{O}(Lm'^2)$ & $\mathcal{O}(m'^2)$ & $\mathcal{O}(Lm'^2 + 2L_fm'^2_f)$ & $\mathcal{O}(m'^2 + 2L_fm^2_f)$ & $\mathcal{O}(8Lm'^2)$ \\ \hline
    \end{tabular}}
    \label{tab:space_complexities}
\end{table*}

\section{Experimental Evaluation}
\label{sec:experimental_setup}

\subsection{Impact of the Early-Exit Components}
\label{sec:impact_earlyexit}

\textbf{Task-driven learning}.
Unlike approaches that rely on auxiliary budget-aware losses~\citep{AdaProp}, EEGNN updates the exit network's weights coming from ($f_c, f_{\nu}$), directly via the task loss. This theoretically ensures that decisions are strictly context-aware and eliminates the need to tune depth $L$ per dataset. We emprically validate the efficacy of this task-driven mechanism through two sets of experiments.

\begin{figure}
    \centering    \includegraphics[width=\linewidth]{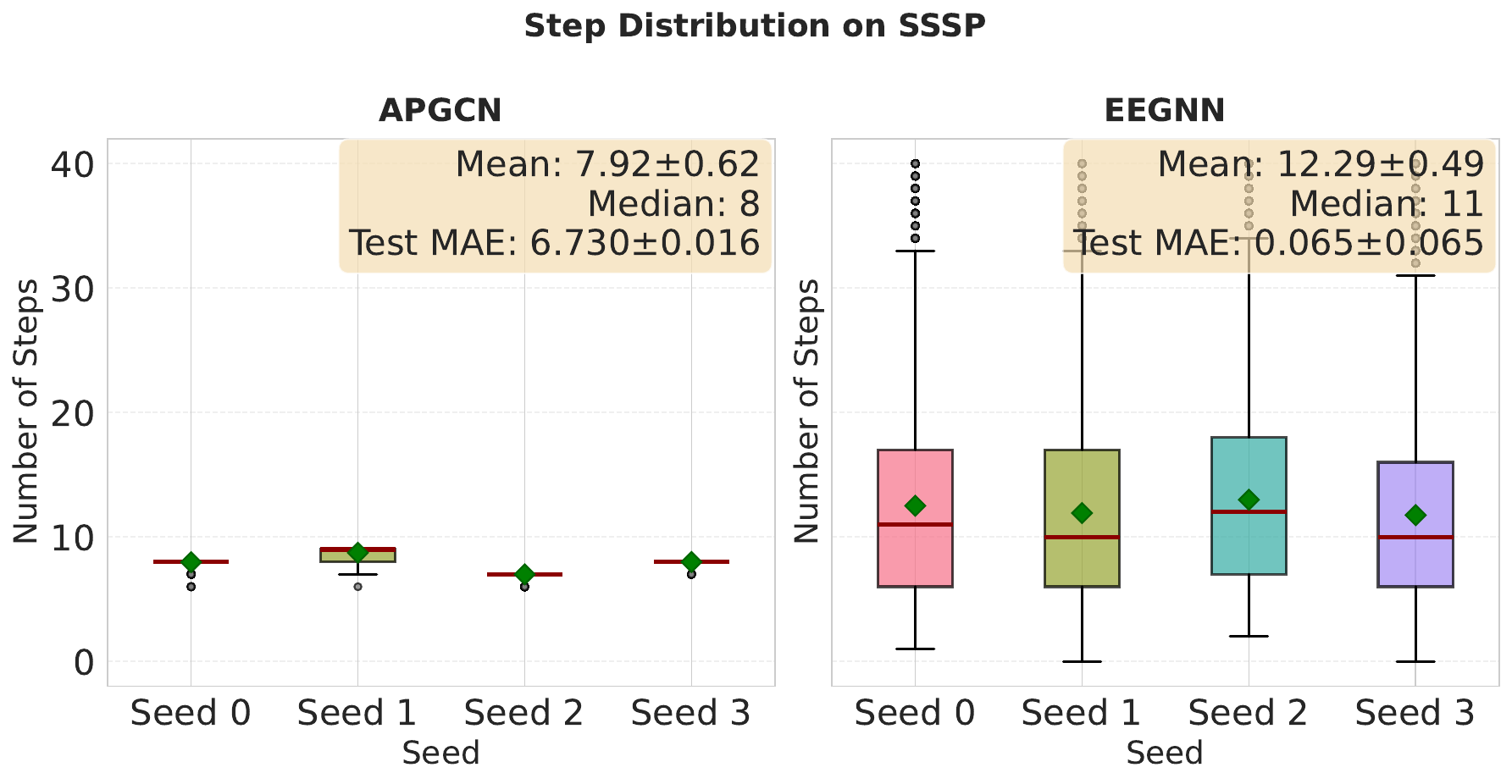}
    \caption{Node exit level statistics on \texttt{sssp}: APGCN and EEGNN.}
\label{fig:box_plot_sssp_exit}
\end{figure}

\textbf{Task awareness}. 
We first evaluate the ability to recognize when deep computation is strictly necessary using the \texttt{SSSP} dataset from the ECHO benchmark~\citep{echo2025benchmark}, a task requiring long-range propagation. We compare EEGNN against APGCN~\citep{AdaProp}, which also utilizes node-level exits. As shown in Table~\ref{tab:echo_results}, EEGNN achieves state-of-the-art performance (MAE 0.065) by correctly maintaining deep exits. Figure~\ref{fig:box_plot_sssp_exit} illustrates the exit statistics across 4 seeds: EEGNN consistently maintains exits in the deep regime, aligning more with the need for long-range reasoning. Conversely, APGCN exits prematurely due to its budget-aware bias, resulting in significantly higher error. Examples of task awareness in Appendix \ref{sec:apgcnvseegnn}.

\begin{figure}[t]
    \centering
    \includegraphics[width = \linewidth]{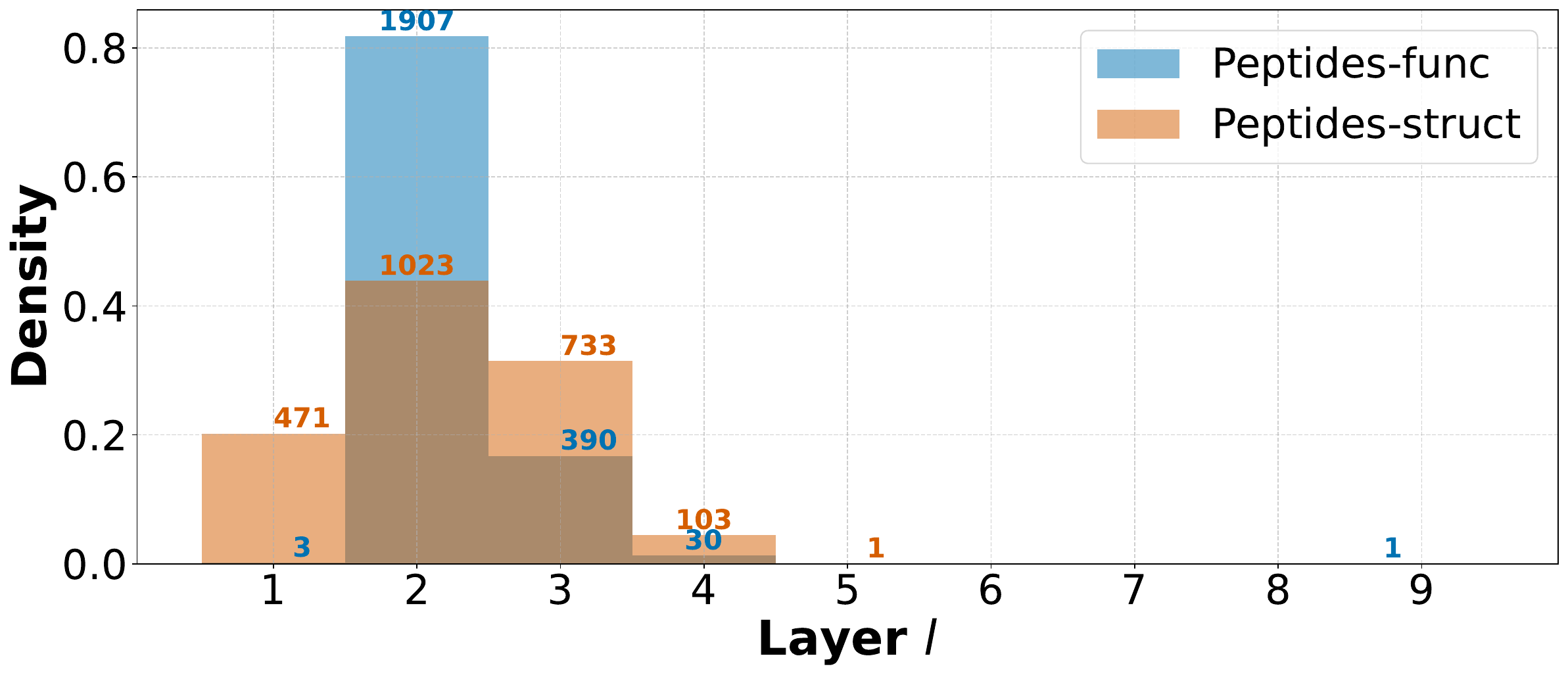}
      \caption{EEGNN's Graph exit layer distributions. Numbers on top of the bars correspond to the number of test graphs that decided to exit at that layer. The associated performance is in Table \ref{tab:result_lrgb_extended}.}
       \label{fig:funcvsstruct}
\end{figure}

\textbf{Task adaptivity}. 
Second, we assess how EEGNN adapts depth to task definitions using \texttt{Peptides-func} (classification) and \texttt{Peptides-struct} (regression) from LRGB~\citep{LRGB}. Despite sharing the same set of graphs, Figure~\ref{fig:funcvsstruct} shows that EEGNN learns distinct exit distributions for each. For \texttt{Peptides-func}, the model predominantly exits after just two layers, while for \texttt{Peptides-struct}, it shifts to a slightly deeper distribution. This aligns with recent findings that these tasks do not strictly require long-range propagation~\citep{measuringlongrange}. EEGNN's adaptive shallow processing achieves $>68\%$ AP, (see Table \ref{tab:result_lrgb_extended}), demonstrating its ability to tailor computational depth to specific task requirements, a key advantage over fixed-depth GNNs. This experiments also complement those  from ECHO, as here EEGNN learns to exit very early. Other robustness checks on the other datasets are provided in Appendix~\ref{sec:node_distributions}.

\begin{figure}[t]
    \centering    
\includegraphics[width=0.8\linewidth]{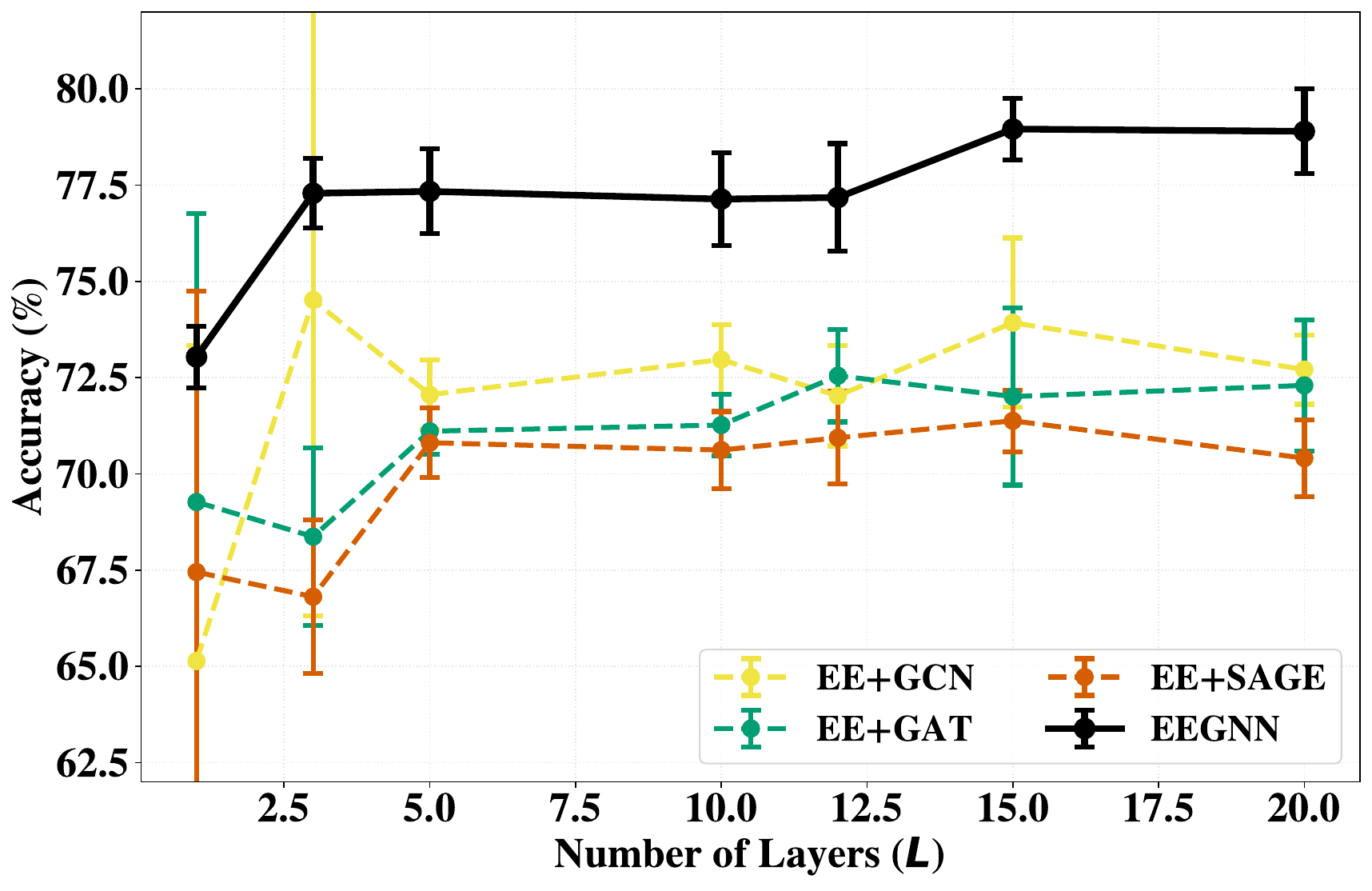}
     \caption{Early-Exit on classic backbones presents unstable results on the \texttt{Questions} dataset.}
         \label{fig:early-exit_unstable}
\end{figure}

\textbf{Modularity}.  
To illustrate modularity, we attach the exit module to standard MPNNs (Appendix~\ref{sec:extended_EE}). We observe that in deep regimes ($L=20$), baseline MPNNs degrade severely, and early exits offer only unstable recovery. In contrast, EEGNN remains stable as depth grows, confirming that the SAS-GNN backbone provides the necessary stability to preserve information across layers, enabling robust early exiting (see Figures~\ref{fig:depth_performance_degrade}-\ref{fig:early-exit_unstable}).

\textbf{Runtime and efficiency}.
EEGNN reduces inference costs by skipping redundant computation, treating $L$ as a maximum budget rather than a fixed requirement. Table~\ref{tab:runtime_questions} (averaged over 1,500 passes, $m' = 32$) confirms that while baselines like Polynormer scale linearly with depth, EEGNN achieves nearly constant inference time even as $L$ increases. Furthermore, thanks to weight sharing, EEGNN maintains a constant parameter count significantly lower than competitive baselines, highlighting both computational and memory efficiency. Later in this section we discuss about the accuracy-efficiency trade-off of our models. Additional runtime and parameters analyses are provided in Appendix~\ref{appendix:run_time_analysis}.
\looseness=-1


\begin{table}[t]
    \centering
        \caption{Runtime and parameter analysis in \texttt{Questions}. Times are in seconds. }
    \label{tab:runtime_questions}
     \adjustbox{max width=\columnwidth}{
    \begin{tabular}{l|c|c|c|c}
        \hline
        \multirow{2}{*}{\textbf{Model}} 
        & \multicolumn{2}{c|}{\textbf{Inference Time (s)}} 
        & \multicolumn{2}{c}{\textbf{Number of Parameters}} \\
        \cline{2-5}
        & \textbf{10 Layers} & \textbf{20 Layers} & \textbf{10 Layers} & \textbf{20 Layers} \\
        \hline
        \textbf{GCN}     
        & $0.0168 \pm 0.0012$ & $0.0251 \pm 0.0037$ & 20,288 & 30,848 \\
        \textbf{Co-GNN}  
        & $0.0352 \pm 0.0038$ & $0.0598 \pm 0.0087$ & 34,982 & 55,782 \\
        \textbf{Polynormer}  
        & $0.0278 \pm 0.0053$ & $0.0381 \pm 0.0078$ & 50,404 & 80,100 \\
        \textbf{SAS-GNN} 
        & $0.0279 \pm 0.0103$ & $0.0425 \pm 0.0133$ & 11,840 & 11,840 \\
        \textbf{EEGNN}   
        & $0.0216 \pm 0.0064$ & $0.0257 \pm 0.0078$ & 12,562 & 12,562 \\
        \hline
    \end{tabular}}
\end{table}

\begin{table}[t]
    \centering
    \caption{Node classification under heterophily. 
    The scores are marked in red for the \textcolor{red}{first}, blue for the \textcolor{blue}{second}, and green for the \textcolor{MyDarkGreen}{third}. Results are averaged across different graph splits.
    Scores are from \citet{coognns, classicgnnsstrongbaselines}.}
    \label{tab:result_heterophilic}
     \adjustbox{max width=\linewidth}{
        \begin{tabular}{l c c c }
        \hline
        \textbf{Model} & \texttt{Amazon Ratings $\uparrow$} & \texttt{Tolokers $\uparrow$} & \texttt{Questions $\uparrow$} \\
        \hline
        \textbf{GCN} & 51.37 ± 0.34 & 83.64 ± 0.67 & 76.58 ± 0.40 \\
        \textbf{SAGE} & 51.12 ± 0.66 & 82.43 ± 0.44 & 76.36 ± 1.50 \\
        \textbf{GAT} & \textbf{\textcolor{MyDarkGreen}{51.48 ± 0.28}} & 83.78 ± 0.43 & 77.95 ± 0.51 \\
        \textbf{GT} & 51.17 ± 0.66 & 83.23 ± 0.64 & 77.95 ± 0.68 \\
        \textbf{Polynormer} & \textbf{\textcolor{red}{54.81 ± 0.49}} & \textbf{\textcolor{red}{85.91 ± 0.74}} & \textbf{\textcolor{blue}{78.92 ± 0.89}} \\
        \textbf{Co-GNN} & 54.17 ± 0.37 & 84.45 ± 1.17 & 76.54 ± 0.95 \\
        \hline
        \textbf{SAS-GNN} & 51.47 ± 0.68 & \textbf{\textcolor{blue}{85.80 ± 0.79}} & \textbf{\textcolor{red}{79.60 ± 1.15}} \\
        \textbf{EEGNN}  & \textbf{\textcolor{blue}{51.54 ± 0.5}} & \textbf{\textcolor{MyDarkGreen}{85.26 ± 0.65}} & \textbf{\textcolor{MyDarkGreen}{78.90 ± 1.15}}\\ 
        \hline
    \end{tabular}}
\end{table}
\begin{table}[ht]
    \centering
    \caption{Performance comparison on ECHO benchmark.}
    \label{tab:echo_results}
    \resizebox{\columnwidth}{!}{%
    \begin{tabular}{l c c c}
        \toprule
        \textbf{Model} & \texttt{diam} $\downarrow$ & \texttt{ecc} $\downarrow$ & \texttt{sssp} $\downarrow$ \\
        \midrule
        DRew & $1.243 \pm 0.047$ & $\textcolor{red}{\mathbf{4.651 \pm 0.020}}$ & $1.279 \pm 0.011$ \\
        GraphCON & $2.969 \pm 0.189$ & $5.474 \pm 0.001$ & $5.734 \pm 0.011$ \\
        GCN & $3.832 \pm 0.262$ & $5.233 \pm 0.034$ & $2.102 \pm 0.094$ \\
        GCNII & $2.005 \pm 0.093$ & $5.241 \pm 0.030$ & $2.128 \pm 0.429$ \\
        GIN & $1.630 \pm 0.161$ & $4.869 \pm 0.092$ & $2.234 \pm 0.271$ \\
        GPS & $2.160 \pm 0.098$ & \textcolor{blue}{$4.758 \pm 0.021$} & $0.472 \pm 0.050$ \\
        GRIT & \textcolor{red}{$\mathbf{1.014 \pm 0.046}$} & $5.091 \pm 0.158$ & \textcolor{blue}{$0.121 \pm 0.013$} \\
        PH-DGN & $1.627 \pm 0.398$ & $5.068 \pm 0.126$ & $1.323 \pm 0.485$ \\
        SWAN & \textcolor{blue}{$1.121 \pm 0.070$} & $4.840 \pm 0.045$ & $0.896 \pm 0.232$ \\
        A-DGN & \textcolor{MyDarkGreen}{$1.151 \pm 0.038$} & $4.981 \pm 0.037$ & $1.176 \pm 0.140$ \\
        GRAFF & $2.731 \pm 0.052$ & $5.344 \pm 0.086$ & $3.363 \pm 1.356$ \\
        APGCN & - & $11.080 \pm 0.099$ & $6.729 \pm 0.018$ \\
        \midrule
        SAS-GNN & $1.914 \pm 0.397$ & \textcolor{MyDarkGreen}{$4.806 \pm 0.018$} & \textcolor{MyDarkGreen}{$0.244 \pm 0.243$} \\
        EEGNN & $1.805 \pm 0.068$ & $5.048 \pm 0.237$ & \textcolor{red}{$\mathbf{0.065 \pm 0.074}$} \\
        \bottomrule
    \end{tabular}%
    }
\end{table}
\subsection{Comparison with the Heterophilic and ECHO Benchmarks}
\label{sec:hetero_lrgb_results}
\textbf{Datasets}.  
(i) \textbf{Heterophilic Node Classification.} We use \texttt{Amazon Ratings} (\texttt{ACC}),
\texttt{Tolokers} (\texttt{AUROC}), and \texttt{Questions} (\texttt{AUROC})~\citep{criticallookatgnn} to assess backbone robustness in low-homophily settings, where connected nodes frequently differ in labels. Our goal is to verify if our parameter-efficient backbone can match state-of-the-art methods in these structurally complex scenarios. 

(ii) \textbf{The ECHO Benchmark.} To evaluate long-range capabilities, we employ the recently released ECHO benchmark \citep{echo2025benchmark}. This suite was designed to bridge the gap left by previous benchmarks, such as LRGB \citep{LRGB}, where tasks were often solvable with shallow receptive fields. We report performance on the synthetic node-level tasks \texttt{sssp} and \texttt{ecc}, and the synthetic graph-level task \texttt{diam}, which require several hop of message-passing in order to perform well.
This benchmark provides an ideal setting to validate the \textit{adaptivity} of our early-exit mechanism. Specifically, we aim to test how EEGNN ranks among long-range specialized architectures, and how it 1) correctly identifies the depth required by the task node/graph-wise, and 2) keeps its features reliable for inference, without suffering from depth-related degradation. Furthermore, we highlight that the ECHO benchmark encompasses a wide and heterogeneous set of topologies~\cite{echo2025benchmark}; these structures frequently induce \textit{topological bottlenecks}, which typically cause performance degradation via over-squashing \cite{demystifyingcommonbeliefs}, presenting another crucial challenge for EEGNN.
In the appendix, we provide additional results on the original LRGB datasets, transductive homophilic node classification, graph classification (TUDataset \citep{tudataset}), and large-scale OGB benchmarks \citep{ogb} in Appendix~\ref{sec:extended_homophily}--\ref{sec:ogb_results}. Full dataset descriptions and implementation details are provided in Appendix~\ref{sec:datasets}.

\textbf{Baselines}. For \textbf{heterophilic datasets} \citep{criticallookatgnn}, we compare with standard MPNNs (GCN, SAGE, GAT), asynchronous methods (Co-GNN), and Transformers (GT, Polynormer). Crucially, to ensure a fair comparison with our theoretically constrained backbone, we adopt MPNN variants without dropout or normalization, as they are also the main responsible for boosting performance \citep{classicgnnsstrongbaselines}. 
For the \textbf{ECHO benchmark}, we categorize baselines into four groups: (i) standard MPNNs (GCN, GIN \cite{WL1}, GCNII \cite{chen2020simpledeepgraphconvolutional}); (ii) rewiring methods as DRew \cite{Drew}; (iii) Graph Transformers (GPS \cite{graphgps}, GRIT \cite{ma2023graphinductivebiasestransformers}); and (iv) Graph NODEs (GraphCON \cite{graphcon}, A-DGN, SWAN \cite{swan}, PH-DGN \cite{PHGNN}), the same class of SAS-GNN and EEGNN.
Additionally, we include APGCN \cite{AdaProp}, to represent budget driven early-exit GNNs. We also implement GRAFF~\cite{GRAFF2} as a form of symmetric-only ablation, complementing the antisymmetric dynamics of A-DGN. Full details on hyperparameters are provided in Appendix~\ref{subsec:hyperparameters}.

\textbf{Results}. On the \textbf{heterophilic datasets} (Table~\ref{tab:result_heterophilic}), both SAS-GNN and EEGNN consistently match or surpass standard MPNN baselines and frequently outperform more complex GTs. For instance, in \texttt{Questions}, SAS-GNN achieves state-of-the-art performance, while EEGNN maintains robust accuracy across all tasks. 
It is important to note that for GCN, SAGE, and GAT, we report results without dropout. This ensures a fair comparison, as both SAS-GNN and EEGNN are trained without normalization layers, dropout, or bias terms. This design choice is imposed to satisfy the theoretical conditions of our stability theorems, and minimization of the energy functional. We further analyze the efficiency advantages resulting from this lightweight design later in this section.

The most significant capabilities of our proposed framework are evident in the \textbf{ECHO benchmark} (Table~\ref{tab:echo_results}).Here, EEGNN and SAS-GNN track closely with GraphNODEs explicitly specialized for long-range reasoning (e.g., SWAN, A-DGN, PH-DGN). 
Notably, in the \texttt{sssp} task, both models rank among the top-3, with EEGNN achieving the best overall performance (MAE 0.065), significantly outperforming the runner-up GRIT (0.121). In other metrics, our models generally offer a superior alternative to standard MPNNs (with the minor exception of GIN on \texttt{diam}).

Unlike standard MPNNs (e.g., GCN), EEGNN manages to outperform GTs on specific tasks. This result is particularly significant given that GTs are inherently resilient to topological over-squashing due to their global attention mechanisms. The superior performance of EEGNN suggests that the inductive biases of the SAS-GNN backbone effectively mitigate topological bottlenecks, allowing for stable deep propagation, while the exit module successfully identifies a meaningful halting point, as in Figure~\ref{fig:box_plot_sssp_exit}.
Finally, the comparison with APGCN is particularly instructive. As shown in Table~\ref{tab:echo_results}, APGCN exhibits always high test MAE. This failure stems primarily from its budget-aware halting mechanism, which forces nodes to exit prematurely. In particular, when long-range reasoning is required, harder samples are ignored rather than deeply processed. In contrast, EEGNN's success corroborates the importance of our \textbf{Neural Adaptive Step}: by learning exit decisions directly via task loss gradients, our method ensures that the chosen halting points align strictly with the computational needs of the task.


\subsection{Accuracy–Efficiency Trade-off}
\label{sec:accuracy_efficiency}
\begin{figure}[t]
    \centering
\includegraphics[width=0.8\linewidth]{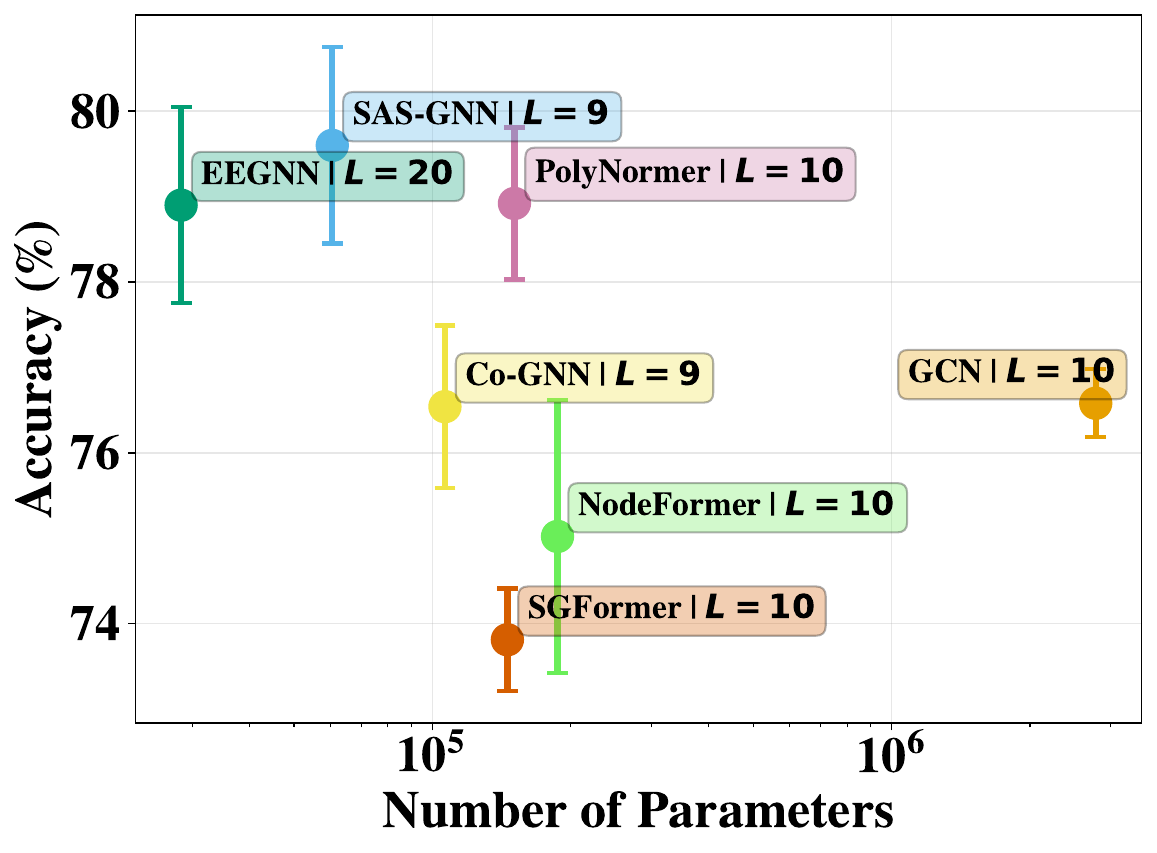}
    \caption{Accuracy-Efficiency trade-off on \texttt{Questions}.}
    \label{fig:tradeoff_questions}
\end{figure}

While our primary objective is not to surpass the raw accuracy of large models, EEGNN aims to occupy a distinct operational point: delivering competitive performance with superior parameter efficiency.
To assess this, we analyze the trade-off between accuracy and model size (trainable parameters) using the best-performing hyperparameters from Table~\ref{tab:result_heterophilic}. Figure~\ref{fig:tradeoff_questions} illustrates this relationship on the \texttt{Questions} dataset. Thanks to the layer-wise weight sharing inherited from the SAS-GNN backbone, EEGNN maintains a constant parameter count regardless of depth. This allows it to reach strong accuracy with an order of magnitude fewer parameters than heavier architectures such as Polynormer, SGFormer~\cite{sgformer} or NodeFormer~\cite{nodeformer}. 
Although EEGNN may slightly underperform on datasets where dropout or normalization are essential for generalization, it may offer a highly favorable balance for resource-constrained scenarios. By avoiding the parameter explosion typical of deep GNNs, EEGNN proves that encompassing adaptive exits can be integrated without inflating the computational budget. Additional analyses, including inference time comparisons, are provided in Appendix~\ref{sec:extended_tradeoff}.
\looseness=-1

\looseness=-1

\subsection{Practical Recommendations}

\textbf{When to use}. EEGNN is ideal when the required reasoning depth is unknown or variable across samples (e.g., ECHO tasks). Its success relies on a core design principle: \textit{early exits are only reliable when the backbone resists deep processing}. Without a stable backbone like SAS-GNN, premature halting often acts merely as a safeguard against OST or topological OSQ rather than a confident decision. EEGNN combines stability with adaptivity, ensuring that the model continues processing until it is truly confident, but without requiring full-depth processing.

\textbf{When not to use}. As a Graph NODE, EEGNN inherits certain limitations. It tends to lag behind heavily regularized architectures or with a large number of parameters on tasks where such mechanisms are critical for generalization. In these cases, where the strict theoretical constraints of ODEs (no dropout/bias) limit performance, or if the problem radius is self-contained, we recommend using other models.

\textbf{Hyperparameter strategy}. Unlike fixed-depth GNNs where depth ($L$) is a critical hyperparameter requiring extensive tuning, EEGNN treats $L$ as a budget. Since our results (Figure~\ref{fig:early-exit_unstable}) show that performance does not degrade with increased depth, we recommend setting $L$ to the maximum value allowed by hardware constraints. This simplifies the search space, allowing practitioners to focus tuning efforts on other hyperparameters, trusting the Neural Adaptive Step to identify the effective depth per sample.
\looseness=-1
\looseness=-1
\section{Conclusions}
\label{sec:conclusions}

We presented EEGNN, an end-to-end framework that removes the fixed-depth limitation of conventional GNNs by introducing a differentiable early-exit mechanism at both node and graph levels. This design enables the network to adapt its depth dynamically based on input complexity: simple inputs exit early to conserve computation, while complex inputs, requiring long-range propagation or richer feature extraction, traverse deeper layers. This flexibility allows users to tailor the model fixing the number of layers as a budget, optimizing for either strict resource budgets or maximum reasoning capacity.
Unlike previous early-exit approaches, which often ignore the inherent instability of deep MPNNs, EEGNN explicitly addresses this challenge through SAS-GNN. This novel, provably stable backbone ensures that intermediate representations remain informative, allowing the model to safely leverage deep propagation when necessary without suffering from classic depth-related message-passing flaws.
Empirical evaluation on heterophilic and long-range benchmarks confirms that EEGNN achieves competitive accuracy, while maintaining constant memory complexity and superior runtime efficiency, while adapting the exit choices to the task at hand.

\textbf{Limitations}.  Despite these advantages, EEGNN inherits some limitations of Graph Neural ODE–style architectures, potentially underperforming on tasks requiring extremely high expressive power. Furthermore, EEGNN does not inherently overcome the expressiveness limits of the 1-WL test, nor does it offer a formal solution to under-reaching issues.

\textbf{Future Work}. Future directions include: (i) integrating more expressive backbones that retain stability in deep configurations, and (ii) extending the early-exit mechanism to edge-level tasks (e.g., link prediction), where handling pairwise dependencies presents a non-trivial challenge.



\bibliographystyle{abbrvnat}
\bibliography{main}

\newpage
\appendix
\onecolumn
\section{Appendix Overview}
\begin{itemize}
    \item \textbf{Appendix \ref{sec:extended_preliminaries}: Additional Preliminaries.}\\
    We provide information related to GraphNODEs, GRAFF and A-DGN, which are the foundation of SAS-GNN, but are not discussed in the main paper. We discuss the seminal work on early-exit mechanisms for GNNs, which is APGCN, that we also use in this paper. We also provide some preliminaries on the Gumbel-softmax distribution.
    \item \textbf{Appendix \ref{sec:proofs}: Proofs of Theorems in Section 4.}\\
    Here, we provide the proofs of the theoretical results that support the design of SAS-GNN. We also show that including edge features into the SAS-GNN update rule, SAS-GNN preserves its stability, non-dissipativeness, and its capability of inducing attraction and repulsion. 
    \item \textbf{Appendix \ref{sec:extended_algorithm}: Additional details on Neural Adaptive-Step Early-Exit.}\\
    We report the extension of Algorithm \ref{alg:eegnn} for node classification to the task of classifying graphs. We comment on how EEGNN is related to the underreaching phenomenon, and we draw a connection of Neural Adaptive Step with Adaptive-Step Runge-Kutta solvers.
    \item \textbf{Appendix \ref{sec:datasets}: Additional Details on the Experimental Setup.}\\
    This section presents additional information on the datasets, training procedures, and the hardware used in the experiments.
    \item \textbf{Appendix \ref{sec:extended_results}: Additional Results.}\\
    This section presents extended results from our experiments with EEGNN. We bring additional evidence that SAS-GNN's design is a proxy to mitigate over-smoothing and over-squashing, and also to retain performance as depth increases. We provide additional baselines on the heterophilic benchmark and LRGB datasets. We provide results on short-range, small, and large-scale homophilic node classification and graph classification datasets. We test how classic GNNs perform when equipped with our early-exit module. We provide evidence that Early-Exit is a promising direction to significantly improve the GNN performance. We illustrate the nodes' exit distributions in the discrete and continuous cases that we obtain for the other datasets.
     \item \textbf{Appendix \ref{sec:extended_related_works}: Extended Related Work.}\\
     In the main paper, we focused our discussion on related work, mainly on GNN papers where the targets are over-smoothing and over-squashing, or papers discussing the adaptive depth choice. Here we extend the discussion by also including GNNs and Early-Exit Neural Networks in general.
\end{itemize}
\looseness=-1
\section{Additional Preliminaries}
\label{sec:extended_preliminaries}
In this section, we provide the necessary background on the specific architectures that inspire our framework, namely Graph Neural ODEs (GraphNODEs), A-DGN, and GRAFF. Furthermore, we detail the Adaptive Propagation GCN (APGCN), the seminal work on early exiting for graphs, and the Gumbel-Softmax distribution, which enables the differentiability of our discrete exit decisions.

\textbf{Graph Neural Ordinary Differential Equations (GraphNODEs).} 
GraphNODEs~\citep{GraphNODE} generalize discrete Graph Neural Networks (GNNs) by modeling the evolution of node features $\mathbf{H}(t)$ as a continuous-time dynamical system. Instead of a fixed sequence of layers $l=1, \dots, L$, the feature transformation is governed by an Ordinary Differential Equation (ODE):
\begin{equation}
    \frac{d\mathbf{H}(t)}{dt} = f(\mathbf{H}(t), \mathcal{G}, \theta_t), \quad \mathbf{H}(0) = \mathbf{X},
\end{equation}
where $\mathbf{X}$ represents the input features, $\mathcal{G}$ the graph structure, and $f$ is a neural network (e.g., a GNN layer) parametrized by $\theta$. The output features are obtained by integrating this system up to a time $T$:
\begin{equation}
\label{eq:integration_eq}
    \mathbf{H}(T) = \mathbf{H}(0) + \int_{0}^{T} f(\mathbf{H}(t), \mathcal{G}, \theta_t) dt.
\end{equation}

This framework allows the use of many ODE solvers (e.g., Euler, Runge-Kutta) for the forward pass and adjoint methods for memory-efficient backpropagation. If we consider the Euler integration (i.e. what we did in the main text) with step size $\tau$, Equation \eqref{eq:integration_eq} becomes the following iterative update for each step $k$:
\begin{equation}
\label{eq:euler_discretization}
    \mathbf{H}^{k+1} = \mathbf{H}^k + \tau f(\mathbf{H}^k, \mathcal{G}, \theta_k).
\end{equation}

Our proposed EEGNN follows this continuous-depth paradigm but introduces a mechanism to dynamically determine the integration endpoint $T_i$ for each node or graph, which relies on considering $\tau$ not as fixed, but as the output of a neural network. In the main text we refer to it as Neural Adaptive Step.

\textbf{Anti-Symmetric Deep Graph Networks. }Let a message-passing update for a node $i$ be
\begin{equation}
\label{eq: ODE}
    \dot{\mathbf{h}}_i^t = f(\mathbf{h}_i^t) = \sigma(\mathbf{\Omega}\mathbf{h}_i^t + \phi_{\mathbf{W}}(\mathbf{H}^t, \Gamma(i)) + b), 
\end{equation}
where $\mathbf{\Omega}, \mathbf{W} \in \mathbb{R}^{d \times d}$ are trainable matrices, $\sigma$ is a non-linearity, $\mathbf{h}_i^t$ is the feature of node $i$ at time $t$, $\Gamma(i)$ is the set of neighbors of $i$, $\mathbf{H}^t \in \mathbb{R}^{n \times d}$ is the matrix containing all the $d$ dimensional features for each node, $b \in \mathbb{R}^{d}$. 
The following definitions are from \cite{ADGN}.
\begin{definition}
A solution $\mathbf{h}_i^t$  of the ODE in Equation \ref{eq: ODE}, with initial condition $\mathbf{h}_i^0$, is stable if for any $ \omega > 0 $, there exists a \( \delta > 0 \) such that any other solution $\mathbf{\tilde{h}}_i^t$ of the ODE with initial condition $\mathbf{\tilde{h}}_i^0$ satisfying  $|\mathbf{h}_i^0 -  \mathbf{\tilde{h}}_i^0| \leq \delta $ also satisfies $|\mathbf{h}_i^t -  \mathbf{\tilde{h}}_i^t| \leq \omega $, for all $ t \geq 0 $.
\end{definition}
\begin{definition}
Let $ E \subseteq \mathbb{R}^d $ be a bounded set that contains any
initial condition $\mathbf{h}_i^0$ for the ODE in Equation 
\ref{eq: ODE}.
The system defined by the ODE in Equation \ref{eq: ODE} is dissipative if there is a bounded set $ B $ where, for any $ E $, there exists $ t^* \geq 0 $ such that $ \{ \mathbf{h}_i^t \mid \mathbf{h}_i^0 \in E \} \subseteq B $ for $t > t^* $.
\end{definition}
In \cite{ADGN}, the authors propose an instance of Equation \eqref{eq: ODE} that is both stable and non-dissipative, leading to an evolution of node features that retain all the information collected during the forward pass (e.g., from $0$ to an arbitrary $t$). Such an instantiation is 
\begin{equation}
\label{eq: ODE1.1}
 \dot{\mathbf{h}}_i^t = f(\mathbf{h}_i^t) = \text{tanh}((\mathbf{\Omega} - \mathbf{\Omega}^{\top})\mathbf{h}_i^t + \phi_{\mathbf{W}}(\mathbf{H}^t, \Gamma(i)) + b),    
\end{equation}
which can be discretized through the Euler method (i.e. This is because it follows the GraphNODE's paradigm) as
\begin{equation}
\frac{\mathbf{h}_i^{t+\tau} - \mathbf{h}_i^t}{\tau} = f(\mathbf{h}_i^t) = \text{tanh}((\mathbf{\Omega} - \mathbf{\Omega}^{\top})\mathbf{h}_i^t + \phi_{\mathbf{W}}(\mathbf{H}^t, \Gamma(i)) + b).
\end{equation}
Here, the author uses the $\tanh$ to keep bounded the Jacobian of the system, and $\phi_{\mathbf{W}}(\mathbf{H}^t, \Gamma(i))$, do not have any dependency on $\mathbf{h}_i^t$.\\
\textbf{Graph Neural Networks as Gradient Flows. }Another message-passing rule \citep{GRAFF2} exploits symmetric matrices $\mathbf{\Omega}_s$ and $\mathbf{W}_s$ to contrast over-smoothing in heterophilic graphs, which is a desired feature for message-passing neural networks. Using the symmetric bias, they are minimizing an energy functional that induces both attraction and repulsion to connected nodes via the eigenvalues of the weight matrices. This is the message-passing update rule:\\
\begin{equation}
\label{eq: graff_gradient_flow}
    \mathbf{H}^{t + \tau} =  \mathbf{H}^t + \tau \sigma(\mathbf{H}^t 
    \mathbf{\Omega}_s + \mathbf{A} 
    \mathbf{H}^t\mathbf{W}_s). 
\end{equation}
As long as $\sigma$ is a non-linearity s.t. $x \sigma(x) \geq 0$ (e.g., ReLU$(\cdot)$ or tanh$(\cdot)$), and the weight matrices are symmetric, this rule is proved in \cite{GRAFF2} to minimize the underlying functional:
\begin{align}
    \label{eq:param_dirich}
    E_{\theta}^{dir}(\mathbf{H}) &= \sum_i \langle \mathbf{h}_i, \mathbf{\Omega}_{s} \mathbf{h}_i \rangle - \sum_{i,j} a_{ij} \langle \mathbf{h}_i, \mathbf{W}_{s} \mathbf{h}_j \rangle \\
    &= \sum_i \langle\mathbf{h}_i, (\mathbf{\Omega}_{s} - \mathbf{W}_{s}) \mathbf{h}_i \rangle + \sum_i \langle\mathbf{h}_i, \mathbf{W}_{s} \mathbf{h}_i \rangle - \sum_{i,j} a_{ij} \langle \Theta_{+} \mathbf{h}_i, \Theta_{+} \mathbf{h}_j \rangle + \sum_{i,j} a_{ij} \langle \Theta_{-} \mathbf{h}_i, \Theta_{-} \mathbf{h}_j \rangle \\
     \label{eq: cholesky}
    &= \sum_i \langle\mathbf{h}_i, (\mathbf{\Omega}_{s} - \mathbf{W}_{s}) \mathbf{h}_i \rangle + \frac{1}{2} \sum_{i,j} \| \Theta_{+}( \nabla \mathbf{H})_{ij} \|^2 - \frac{1}{2} \sum_{i,j} \| \Theta_{-}( \nabla \mathbf{H})_{ij} \|^2,
\end{align}
We consider $a_{i,j} = \frac{1}{ \sqrt{d_i \cdot d_j}}$, which assigns to each edge a weight that depends on the degrees $d_i$, $d_j$ of the nodes $i$ and $j$. We leverage the symmetry of \( \mathbf{W}_s \in \mathbb{R}^{m' \times m'} \), which allows spectral decomposition as \( \mathbf{W}_s = \Psi \mathrm{diag}(\pmb{\mu}) \Psi^\top \). The eigenvalue vector \( \pmb{\mu} \) can be split into its positive and negative components, yielding the decomposition:
\[
\mathbf{W}_s = \Psi \mathrm{diag}(\pmb{\mu}_+) \Psi^\top + \Psi \mathrm{diag}(\pmb{\mu}_-) \Psi^\top = \mathbf{W}_+ - \mathbf{W}_-,
\]
where \( \mathbf{W}_+ \) and \( \mathbf{W}_- \) are real, symmetric, and positive semi-definite matrices.

We then apply the Cholesky decomposition to each term, expressing:
\[
\mathbf{W}_+ = \Theta_+^\top \Theta_+, \qquad \mathbf{W}_- = \Theta_-^\top \Theta_-,
\]
where \( \Theta_+, \Theta_- \in \mathbb{R}^{m' \times m'} \) are lower triangular matrices. 

We can see that a gradient flow for such energy, namely Equation \eqref{eq: graff_gradient_flow}, can minimize or maximize the edge gradients computed on the node features. This results in a model's behavior that allows for attraction and repulsion weighted by $\Theta_+$, and $\Theta_-$, specifically the positive semi-definite part \( \Theta_+ \) contributes to smoothing by encouraging alignment between neighboring node features, while \( \Theta_- \) induces repulsion, preserving sharp differences.

\textbf{Adaptive Propagation GCN (APGCN).}
APGCN~\citep{AdaProp} is the first approach to introduce adaptive depth in GNNs using a halting mechanism inspired by Adaptive Computation Time (ACT)~\cite{graves2017adaptivecomputationtimerecurrent} for RNNs. In APGCN, each node $i$ maintains a halting probability $p_i^k$ at each layer $k$. The halting unit is defined as a sigmoid function over the current node state: $h_i^k = \sigma(\text{MLP}(\mathbf{h}_i^k))$.
The process continues until the cumulative probability exceeds a threshold:
\begin{equation}
    N(i) = \min \{ k' : \sum_{k=1}^{k'} h_i^k \ge 1 - \epsilon \},
\end{equation}
where $1 - \epsilon$ is a confidence threshold (typically 0.99).
Crucially, APGCN requires an auxiliary \textit{ponder cost} added to the loss function, $\mathcal{L}_{budget} = \sum_i N(i) + R_i$, to penalize the number of steps and prevent the model from running indefinitely. This explicit penalty creates a "budget-aware" bias, where the model is incentivized to exit early to minimize the loss, potentially at the cost of task performance (as discussed in Section \ref{sec:impact_earlyexit}). The final loss for APGCN becomes: 
\begin{equation}
    \mathcal{L} = \mathcal{L}_{task} + \mu \cdot \mathcal{L}_{budget}
\end{equation}

Here $\mu$ is a coefficent needed to weight the need for budget in the early-exit setting. We remark that the halting unit cannot be trained end-to-end if this additional loss is missing.

\subsection{The Gumbel Distribution and the Gumbel-Softmax Temperature}
\label{sec:gumbel_softmax}
The following exposition largely follows \citet{coognns}.
The Gumbel distribution is widely used to model the maximum (or minimum) of a set of random variables. Its probability density function is asymmetric and has heavy tails, making it suitable for representing rare or extreme events. When applied to logits or scores corresponding to discrete choices, the Gumbel-Softmax estimator transforms these into a probability distribution over the available options.

\begin{figure}[t]
    \centering
    \includegraphics[width=0.5\textwidth]{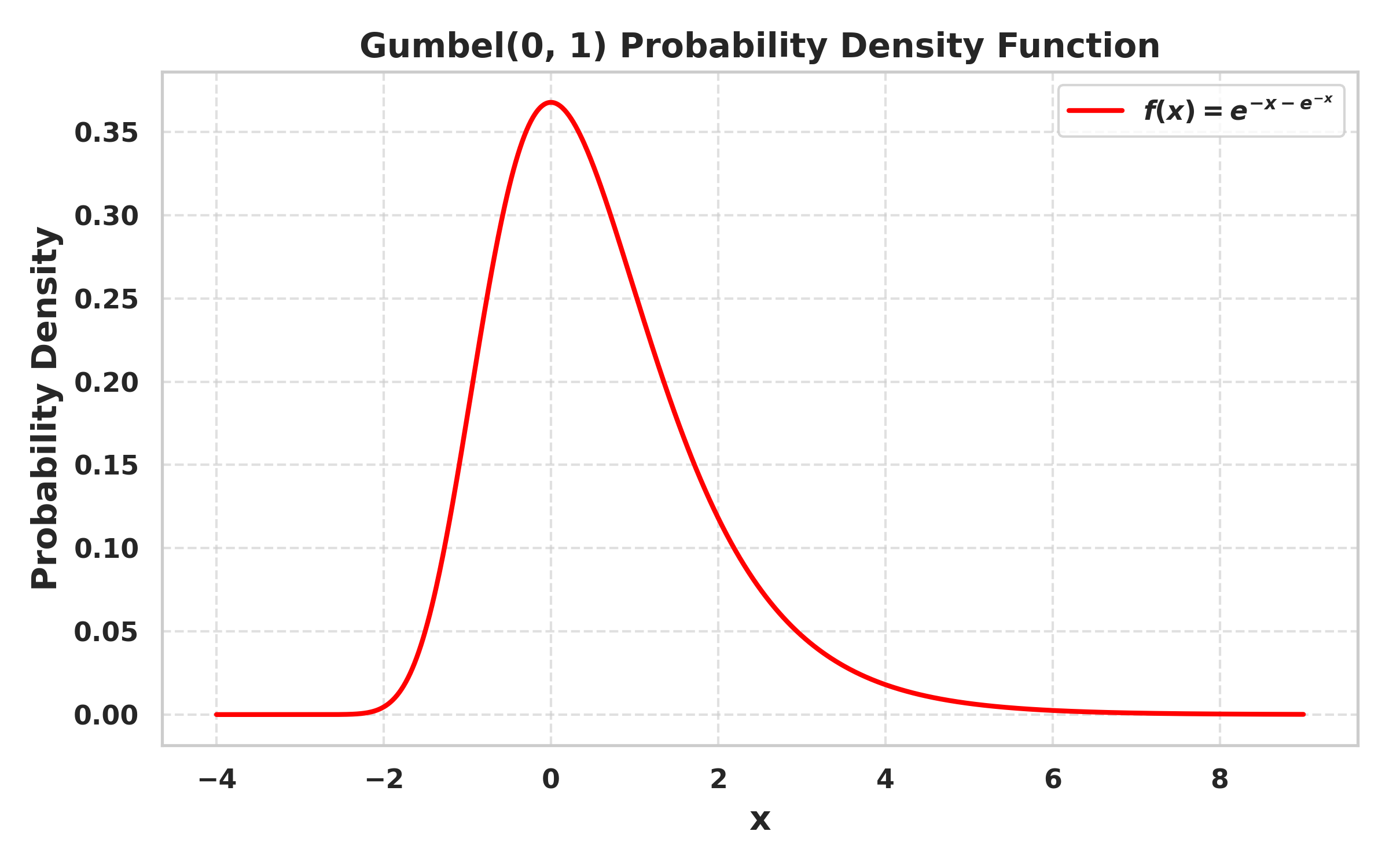}
    \caption{The pdf \( f(x) = e^{-x - e^{-x}} \) of Gumbel(0, 1).}
    \label{fig:gumbel_pdf}
\end{figure}

The probability density function of a variable \( X \sim \text{Gumbel}(0, 1) \) is defined as:
\begin{equation}
f(x) = e^{-x - e^{-x}},
\end{equation}
and is shown in Figure~\ref{fig:gumbel_pdf}.

The Straight-Through Gumbel-Softmax estimator typically benefits from learning an inverse temperature parameter before sampling actions, a strategy we adopt in our experiments. For a given graph \( \mathcal{G} = (\mathcal{V}, \mathcal{E}, \mathbf{X}) \), the inverse temperature for node \( v \in \mathcal{V} \) is computed by applying a bias-free linear layer \( f_{\nu} : \mathbb{R}^{m'} \rightarrow \mathbb{R} \) to the intermediate node representation \( \mathbf{H}^l \in \mathbb{R}^{n \times m'} \). To ensure that the temperature remains positive, we apply a smooth approximation of the ReLU function followed by a bias term \( \nu_0 \in \mathbb{R} \):

\begin{equation}
\label{eq:straight_through}
f_{\nu}(\mathbf{H}^l, \nu_0) = \frac{1}{\bm{\nu}^l} = \log \left(1 + \exp\left(\text{g}(\mathbf{H}^l)\right)\right) + \nu_0    
\end{equation}

Here, \( \nu_0 \) sets the minimum inverse temperature, thereby controlling the upper bound of the temperature. In our setup, we implement \( f_{\nu} \) as in Equation~\ref{eq:straight_through}, with \( \text{g} = \text{GNN}(\cdot) \) for node classification (see Algorithm~\ref{alg:eegnn}) and \( \text{g} = \text{MLP}(\cdot) \) for graph classification (see Algorithm~\ref{alg:eegnn_gc}). The confidence estimator \( f_c \) is defined as \( \text{g}(\mathbf{H}^l) \), choosing $\text{g}$ based on the task, similarly to \( f_{\nu} \). We implemented $f_{\nu}$, and $f_{c}$ using SAS-GNN in the ECHO benchmark experiments, while we used MeanGNN~\cite{coognns} otherwise.

\section{Proofs of Section \ref{sec:method}}
\label{sec:proofs}
We want to prove and understand the conditions under which our message-passing rule is stable, non-dissipative, and can induce attraction or repulsion among adjacent nodes. Our proposal is the following:
\begin{equation}
\label{eq:sas-gnn_derivative_appendix}
    \dot{\mathbf{H}}^{t} = \sigma_1(- \sigma_2(\mathbf{H}^t \mathbf{\Omega}_{as}) +  \mathbf{\bar{A}} \mathbf{H}^t\mathbf{W}_s),
\end{equation}
where $\mathbf{\Omega}_{as} = \mathbf{\Omega} - \mathbf{\Omega}^{\top}$.
We state two different Theorems to understand if we can guarantee all of these specifics. In the first proof, we leverage the node-wise Equation for $\mathbf{h}_i^t$ to demonstrate that each node evolution is stable and non-dissipative. \\The following exposition share similarities with \cite{AntisymmetricRNNs, ADGN, GRAFF2}.
\subsection{Proof of Theorem \ref{theo:stability}: stability and non-dissipativeness}
\textit{Proof of the Theorem. }We assume that we have a message-passing of the type:
\begin{equation}
\label{eq:node_wise_sas_GNN}
    \dot{\mathbf{h}}_i^t = \sigma_1(-\sigma_2((\mathbf{\Omega} - \mathbf{\Omega}^{\top}))\mathbf{h}_i^t + \phi_{\mathbf{W_s}}(\mathbf{H}^t, \Gamma(i))),
\end{equation}
where the derivative of $\sigma_1$ is a bounded and point-wise non-linear function, $\sigma_2$ is a non-linear activation function, $\phi_{\mathbf{W_s}}(\mathbf{H}^t, \Gamma(i))$ do not depend on $\mathbf{h}_i^t$, and $\mathbf{W}_s$ is a symmetric matrix.
We can  rewrite \eqref{eq: ODE} as:
\begin{equation}
\label{eq: ODE2}
\frac{\partial \mathbf{h}_i^t}{d t} = f(\mathbf{h}_i^t) 
\end{equation}
If we differentiate \eqref{eq: ODE2} w.r.t. the initial condition of the node $i$ we have:
\begin{equation}
\label{eq: ODE3}
\frac{d}{d \mathbf{h}_i^0}\frac{d \mathbf{h}_i^t}{d t} =\frac{d}{\partial \mathbf{h}_i^0} f(\mathbf{h}_i^t), 
\end{equation}
which can be adjusted through the chain rule as
\begin{equation}
\label{eq: ODE4}
\frac{d}{d t}\frac{\partial \mathbf{h}_i^t}{\partial \mathbf{h}_i^0} =\frac{d(f(\mathbf{h}_i^t))}{\partial \mathbf{h}_i^t} \frac{\partial \mathbf{h}_i^t}{\partial \mathbf{h}_i^0}, 
\end{equation}
where $\frac{d(f(\mathbf{h}_i^t))}{\partial \mathbf{h}_i^t}$ is the Jacobian of the system. So we rewrite Equation \eqref{eq: ODE4} in these terms.
\begin{equation}
\label{eq: ODE5}
\frac{d}{d t}\frac{\partial \mathbf{h}_i^t}{\partial \mathbf{h}_i^0} =J^t \frac{\partial \mathbf{h}_i^t}{\partial \mathbf{h}_i^0}, 
\end{equation}
Here, assuming that $J^t$ does not change with $t$, and $\frac{\partial \mathbf{h}_i^t}{\partial \mathbf{h}_i^0}$ as the system's state variables, we can write the solution of Equation \eqref{eq: ODE5} as
\[
\frac{\partial \mathbf{h}_i^t}{\partial \mathbf{h}_i^0} = e^{tJ} = \mathbf{T} e^{t\mathbf{\Lambda}} \mathbf{T}^{-1}
\]
where we apply the spectral decomposition of $J$, and get its eigenvectors $\mathbf{T}$, and eigenvalues $\mathbf{\Lambda} = diag(\lambda_i)$. As denoted in \cite{ADGN}, to guarantee the stability of the system, we require the $Re(\lambda_i) \leq 0,  \forall i = 1,\dots,d$. However, when the eigenvalues are less than zero, we can lose the information of the previous node representations, leading to a dissipative behavior. For this reason we want $Re(\lambda_i) = 0,  \forall i = 1,\dots,d$, having only imaginary eigenvalues. In the case of Equation \eqref{eq: ODE1.1}, the Jacobian has only imaginary eigenvalues and takes the form of
\begin{equation}
    J^t = \text{diag}[\text{tanh}'(( \mathbf{\Omega} - \mathbf{\Omega}^{\top}) \mathbf{h}_i^t + \phi_W(\mathbf{H}^t, \Gamma(i)) + b)] (\mathbf{\Omega} - \mathbf{\Omega}^{\top})
\end{equation}
Here, the choice of the antisymmetric multiplication by $\mathbf{\Omega} - \mathbf{\Omega}^{\top}$ guarantees the presence of imaginary eigenvalues, and left-multiplying it by a diagonal matrix does not change the nature of the overall spectrum. This was proved in \cite{ADGN}. Our Jacobian instead becomes:
\begin{equation}
    J^t = -\text{diag}[\sigma_1'(-\sigma_2(( \mathbf{\Omega} - \mathbf{\Omega}^{\top}) \mathbf{h}_i^t) + \phi_{W_s}(\mathbf{H}^t, \Gamma(i)))]\text{diag}[\sigma'_2( (\mathbf{\Omega} - \mathbf{\Omega}^{\top})\mathbf{h}_i^t)](\mathbf{\Omega} - \mathbf{\Omega}^{\top}).
\end{equation}
Since we assume that the derivative of $\sigma_1$ is bounded, the assumption that $J^t$ stays constant can be valid.
Hereafter, we show this proof, similarly to \cite{ADGN}. \\Defining \(\mathbf{A} =  -\text{diag}[\sigma_1'(-\sigma_2(( \mathbf{\Omega} - \mathbf{\Omega}^{\top}) \mathbf{h}_i^t) + \phi_{W_s}(\mathbf{H}^t, \Gamma(i)))]\text{diag}[\sigma'_2( (\mathbf{\Omega} - \mathbf{\Omega}^{\top})\mathbf{h}_i^t)] \) and \( \mathbf{B} = (\mathbf{\Omega} - \mathbf{\Omega}^{\top}) \), we have:

\[
J^t = \mathbf{AB}.
\]

We now want to prove that the eigenvalues of $\mathbf{AB}$ are purely imaginary, namely $Re(\lambda_i)=0$ for all $\lambda_i$ eigenvectors of $\mathbf{AB}$.
If all eigenvectors are zero, then the statement holds. 
Let us now consider an eigenpair of AB, where the eigenvector is denoted by $\mathbf{v}$ and the eigenvalue by $\lambda$. Then:
\begin{equation}
\mathbf{ABv} = \lambda \mathbf{v},
\label{eq:ABv}
\end{equation}

$\mathbf{A}$ is a diagonal matrix with non-negative entries, let us denote by $\mathcal{I} \in [d]$ the set of indices s.t. $\mathcal{I}  = \{ i \in [d]: \mathbf{A}_{ii} = 0 \} $. If $\mathbf{v}$ eigenvector of $\mathbf{AB}$ then $\mathbf{v}_j = 0 \; \forall j \in \mathcal{I}$.  
Indeed, we need  for each entry $k$, $\sum_{i}\mathbf{AB}_{ki}v_i = \lambda v_k$; for $j $ in $\mathcal{I} $ the row the $j-$th row of $\mathbf{AB}$ is 0. So we have $\lambda v_j = 0$,  since we assumed $\lambda \neq 0,$ this implies $v_j = 0$.

Let $\mathbf{D}$ be diagonal matrix with entries that are defined as $ \mathbf{D}_{ii} = \sqrt{\mathbf{A}_{ii}}$. We moreover denote by $\tilde{\mathbf{D}}^{-1}$ the matrix defined as 
\begin{equation}
\tilde{\mathbf{D}}^{-1}_{ii} = \begin{cases}
        \frac{1}{\mathbf{D}_{ii}} \; \; \text{if} \; i \notin \mathcal{I} \\
        0 \; \; \text{if} \; i \in \mathcal{I}.  
    \end{cases}
\end{equation}

and by $\tilde{\mathbf{I}}$ the diagonal matrix defined as 

\begin{equation}
\tilde{\mathbf{I}}_{ii} = \begin{cases}
        1 \; \; \text{if} \; i \notin \mathcal{I} \\
        0 \; \; \text{if} \; i \in \mathcal{I}.  
    \end{cases}
\end{equation}

Notice that $\tilde{\mathbf{D}}^{-1} {\mathbf{D}} = {\mathbf{D}} \tilde{\mathbf{D}}^{-1} = \tilde{\mathbf{I}}$, and that 
$ \tilde{\mathbf{I}} \mathbf{A} = \mathbf{A} $ and $\tilde{\mathbf{I}} \mathbf{D} = \mathbf{D}  $.

Let $\mathbf{M}$ be defined as  $\mathbf{M} =\tilde{\mathbf{D}}^{-1} \mathbf{AB} \mathbf{D}$. 

$\mathbf{M} $ is skew symmetric, indeed $\mathbf{M} = \tilde{\mathbf{D}}^{-1} \mathbf{AB} \mathbf{D} = \tilde{\mathbf{D}}^{-1} \mathbf{D D B} \mathbf{D} = \tilde{\mathbf{I}} \mathbf{ D B D}  =  \mathbf{ D B D} $.

It is straightforward to derive that $\mathbf{M}$ is so that the rows in the set $\mathcal{I}$ are null vectors.

We have that for every vector $\mathbf{w}$ s.t. $\mathbf{w}_j = 0 \; \forall j \in \mathcal{I}$, it holds that 
\begin{equation}
\label{eq:almost_similarity}
    \mathbf{ DM} \tilde{\mathbf{D}}^{-1} \mathbf{w} = \mathbf{ABw}.
\end{equation}

This property can be easily verified by noticing that if $\mathbf{w}$ has such a property, then $\tilde{\mathbf{I}}\mathbf{w} = \mathbf{w}$.
Doing all the calculations,
$\mathbf{ DM} \tilde{\mathbf{D}}^{-1} \mathbf{w}= \mathbf{ D \tilde{\mathbf{D}}^{-1}} \mathbf{ABD} \tilde{\mathbf{D}}^{-1} \mathbf{w}   =  \tilde{\mathbf{I}} \mathbf{AB} \tilde{\mathbf{I}}\mathbf{w} =  \mathbf{AB}\mathbf{w}. $

We can now prove that every eigenvector of $\mathbf{AB}$ is also an eigenvector of $\mathbf{M}$.

We have already observed that if $\mathbf{v}$ eigenvector of $\mathbf{AB}$, it has entries in $\mathcal{I}$ equal to $0$.
If $  \mathbf{AB}\mathbf{v } = \lambda \mathbf{v}$, using (\ref{eq:almost_similarity}),  

$\mathbf{ DM} \tilde{\mathbf{D}}^{-1} \mathbf{v} = \lambda \mathbf{v}$, from which using that all the rows in $\mathbf{M}$ with index in $\mathcal{I}$ are null vectors and that similarly the entries of $\mathbf{v}$ with index in $\mathcal{I}$ are null we obtain, $\mathbf{ M} \tilde{\mathbf{D}}^{-1} \mathbf{v} = \lambda \tilde{\mathbf{D}}^{-1}  \mathbf{v}$.

Since $\mathbf{M} $ is skew-symmetric,  it has only pure imaginary eigenvalues, so $\lambda$ is pure imaginary.








As a result, all eigenvalues of $\mathbf{J}^t$ are purely imaginary.













As we can see, this process remains the same regardless of the symmetric matrix $\mathbf{W}_{s}$. We must assure that $\phi_{\mathbf{W}}(\mathbf{H}^t, \Gamma(i))$ does not depend on $\mathbf{h}_i$, otherwise the Jacobian would have become
\begin{equation}
    J^t = -\text{diag}[\sigma_1'(-\sigma_2(( \mathbf{\Omega} - \mathbf{\Omega}^{\top}) \mathbf{h}_i^t) + \phi_{W_s}(\mathbf{H}^t, \Gamma(i)))]\text{diag}[\sigma'_2( (\mathbf{\Omega} - \mathbf{\Omega}^{\top})\mathbf{h}_i^t)](\mathbf{\Omega} - \mathbf{\Omega}^{\top}) + C),
\end{equation}
we can guarantee that there is no dependency, since we consider $\phi_{\mathbf{W}}(\mathbf{H}^t, \Gamma(i)) = \bar{\mathbf{A}}\mathbf{H}^t\mathbf{W}_s$, with no self-loops in $\bar{\mathbf{A}}$.
Thus, we can conclude that Equation \ref{eq:node_wise_sas_GNN} is stable and non-dissipative, even if $\mathbf{W}_s$ is symmetric, and no specific characteristics are associated with $\sigma_2$; it must only be a point-wise function. Moreover, the derivative of $\sigma_1$ must be bounded within an interval, in order to let the constant Jacobian assumption be valid. 
This concludes the proof of the Theorem.\\
\subsection{Proof of the Theorem \ref{theo:energy}: attraction and repulsion}
Now, what we aim to understand is whether Equation \ref{eq:sas-gnn_derivative_appendix} can also induce attraction and repulsion edge-wise as in \cite{GRAFF2}. Our assumptions are the same as the previous proof, but we also assume $\sigma_2$ to be a point-wise activation function that returns only positive values, which do not violate the assumption of the previous proof.\\
Taking into consideration Equation \eqref{eq:param_dirich}, we notice that $\mathbf{\Omega}$ is not the main responsible for the attraction and repulsion behavior, but is $\mathbf{W}_s$. Because of this, we want to understand whether substituting $\mathbf{\Omega_{s}}$ with $\mathbf{\Omega_{as}}$ could preserve stability, non-dissipativeness, edge-wise attraction and repulsion. By simply doing this substitution, we have:
\begin{align}
    \label{eq:param_dirich2}
    E_{\theta}^{dir}(\mathbf{H}) &= \sum_i \langle \mathbf{h}_i, \mathbf{\Omega}_{as} \mathbf{h}_i \rangle - \sum_{i,j} a_{ij} \langle \mathbf{h}_i, \mathbf{W}_{s} \mathbf{h}_j \rangle \\
    &= \sum_i \langle\mathbf{h}_i, (\mathbf{\Omega}_{as} - \mathbf{W}_{s}) \mathbf{h}_i \rangle + \sum_i \langle\mathbf{h}_i, \mathbf{W}_{s} \mathbf{h}_i \rangle - \sum_{i,j} a_{ij} \langle \Theta_{+} \mathbf{h}_i, \Theta_{+} \mathbf{h}_j \rangle + \sum_{i,j} a_{ij} \langle \Theta_{-} \mathbf{h}_i, \Theta_{-} \mathbf{h}_j \rangle \\
     \label{eq: cholesky2}
    &= \sum_i \langle\mathbf{h}_i, (\mathbf{\Omega}_{as} - \mathbf{W}_{s}) \mathbf{h}_i \rangle + \frac{1}{2} \sum_{i,j} \| \Theta_{+}( \nabla \mathbf{H})_{ij} \|^2 - \frac{1}{2} \sum_{i,j} \| \Theta_{-}( \nabla \mathbf{H})_{ij} \|^2. 
\end{align}
which seems to imply that the antisymmetry would not affect the attraction/repulsion framework. However, the gradient flow of this energy would become: 

\begin{equation}
\label{eq: graff_gradient_flow_fail}
    \dot{\mathbf{H}}^t =  (\mathbf{H}^t 
    \mathbf{\Omega}_{as} + \mathbf{H}^t 
    \mathbf{\Omega}_{as}^{\top}) + \mathbf{A} 
    \mathbf{H}^t\mathbf{W}_s 
\end{equation}
which leads to:
\begin{equation}
\label{eq: graff_gradient_flow_fail2}
    \dot{\mathbf{H}}^t =  \mathbf{A} 
    \mathbf{H}^t\mathbf{W}_s,
\end{equation}
due to the antisymmetric matrix properties of $\mathbf{\Omega}_{as} = -\mathbf{\Omega}_{as}^{\top}$.
Unfortunately, Equation \ref{eq: graff_gradient_flow_fail2} would not exploit the antisymmetric properties associated with the Jacobian, and the results of the previous proof would not be valid.\\
So, we cannot directly use antisymmetric weights to minimize the energy in Equation~\eqref{eq:param_dirich}, since antisymmetric matrices do not yield a positive semi-definite quadratic form. However, we can still investigate whether the message-passing dynamics defined in Equation~\eqref{eq:sas-gnn_derivative_appendix} induce edge-wise attraction and repulsion. To do so, we define an alternative energy functional whose minimization would have equivalent edge-level effects.

We consider the following energy:
\begin{align}
    \label{eq:reduced_param_dirich2}
    E_{\theta}(\mathbf{H}) &= - \sum_{i,j} a_{ij} \langle \mathbf{h}_i, \mathbf{W}_s \mathbf{h}_j \rangle \\
    &= -\sum_i \langle \mathbf{h}_i, \mathbf{W}_s \mathbf{h}_i \rangle + \sum_i \langle \mathbf{h}_i, \mathbf{W}_s \mathbf{h}_i \rangle - \sum_{i,j} a_{ij} \langle \Theta_+ \mathbf{h}_i, \Theta_+ \mathbf{h}_j \rangle + \sum_{i,j} a_{ij} \langle \Theta_- \mathbf{h}_i, \Theta_- \mathbf{h}_j \rangle \\
    \label{eq:cholesky_energy}
    &= -\sum_i \langle \mathbf{h}_i, \mathbf{W}_s \mathbf{h}_i \rangle + \frac{1}{2} \sum_{i,j} \| \Theta_+ (\nabla \mathbf{H})_{ij} \|^2 - \frac{1}{2} \sum_{i,j} \| \Theta_- (\nabla \mathbf{H})_{ij} \|^2,
\end{align}
where \( (\nabla \mathbf{H})_{ij} = \frac{\mathbf{h}_i}{\sqrt{d_i}} - \frac{\mathbf{h}_j}{\sqrt{d_j}} \) is the discrete node-wise gradient.

To arrive at this form, we leverage the symmetry of \( \mathbf{W}_s \in \mathbb{R}^{m' \times m'} \), which allows spectral decomposition as \( \mathbf{W}_s = \Psi \mathrm{diag}(\pmb{\mu}) \Psi^\top \). The eigenvalue vector \( \pmb{\mu} \) can be split into its positive and negative components, yielding the decomposition:
\[
\mathbf{W}_s = \Psi \mathrm{diag}(\pmb{\mu}_+) \Psi^\top + \Psi \mathrm{diag}(\pmb{\mu}_-) \Psi^\top = \mathbf{W}_+ - \mathbf{W}_-,
\]
where \( \mathbf{W}_+ \) and \( \mathbf{W}_- \) are real, symmetric, and positive semi-definite matrices.

We then apply the Cholesky decomposition to each term, expressing:
\[
\mathbf{W}_+ = \Theta_+^\top \Theta_+, \qquad \mathbf{W}_- = \Theta_-^\top \Theta_-,
\]
where \( \Theta_+, \Theta_- \in \mathbb{R}^{m' \times m'} \) are lower triangular matrices. Substituting these into the energy functional results in Equation~\eqref{eq:cholesky_energy}, which clearly separates the smoothing (attraction) and sharpening (repulsion) effects. The positive semi-definite part \( \Theta_+ \) contributes to smoothing by encouraging alignment between neighboring node features, while \( \Theta_- \) induces repulsion, preserving sharp differences.

Although this energy is no longer a direct generalization of the classic Dirichlet energy, the spectral properties of \( \mathbf{W}_s \) still allow edge-level attraction and repulsion to emerge via its eigenstructure. In the following, we analyze the time derivative of this energy using the dynamics from Equation~\eqref{eq:sas-gnn_derivative_appendix}, to verify whether the system implicitly minimizes it through evolution.

\begin{equation}
\label{eq:gradient_energy}
    \nabla_{\mathbf{H}} E_{\theta}(\mathbf{H}^t) =  -\mathbf{A}\mathbf{H}^t(\frac{\mathbf{W}_s + \mathbf{W}_s}{2}) = -\mathbf{A}\mathbf{H}^t\mathbf{W}_s,
\end{equation}
Without loss of generality, we use a unique symmetric matrix $\mathbf{W}_s$.
If we want to represent $\nabla_{\mathbf{H}} E_{\theta}(\mathbf{H}^t)$ in a vectorized form, where we stack all the columns together, $ \nabla_{\mathbf{H}} E_{\theta}(\mathbf{H}^t) \in \mathbb{R}^{n \times m'}$ becomes $\text{vec}( \nabla_{\mathbf{H}} E_{\theta}(\mathbf{H}^t)) \in \mathbb{R}^{nm' \times 1}$. According to this we derive Equation \eqref{eq:gradient_energy} through the kronecker product $\otimes$ formalism as
\begin{equation}
\label{eq:gradient_energy_vec}
   \text{vec}(\nabla_{\mathbf{H}} E_{\theta}(\mathbf{H}^t)) =  -(\mathbf{W}_s^\top \otimes \mathbf{A})\text{vec}(\mathbf{H}^t) 
\end{equation}
And we can also derive Equation \eqref{eq:sas-gnn_derivative_appendix} as 
\begin{equation}
    \text{vec}(\dot{\mathbf{H}}^t) = \sigma_1((\mathbf{W}_s^\top \otimes \mathbf{A})\text{vec}(\mathbf{H}^t) - \sigma_2((\mathbf{\Omega}_{as}^\top \otimes \mathbf{I}_n)\text{vec}(\mathbf{H}^t)))
\end{equation}

The time derivative through this formalism can be written as: 
\begin{align}
\label{eq:time_derivative}
    \frac{dE_{\theta}(\mathbf{H}^t)}{dt} &= \text{vec}(\nabla_{\mathbf{H}} E_{\theta}(\mathbf{H}^t))^{\top} \text{vec}(\dot{\mathbf{H}}^t) \\ 
    &= -(\mathbf{W}_s^\top \otimes \mathbf{A})\text{vec}(\mathbf{H}^t) \sigma_1((\mathbf{W}_s^\top \otimes \mathbf{A})\text{vec}(\mathbf{H}^t) - \sigma_2((\mathbf{\Omega}_{as}^\top \otimes \mathbf{I}_n)\text{vec}(\mathbf{H}^t)))\\
    &= -\mathbf{Z}^t \sigma_1(\mathbf{Z}^t + \mathbf{Y}^t)
\end{align}
We consider $\mathbf{Z}^t = (\mathbf{W}_s^\top \otimes \mathbf{A})\text{vec}(\mathbf{H}^t)$ and $\mathbf{Y}^t = - \sigma_2((\mathbf{\Omega}_{as}^\top \otimes \mathbf{I}_n)\text{vec}(\mathbf{H}^t))$, which are $nm'$-dimensional vectors. We refer to $\mathbf{Z}^t_i$ and $\mathbf{Y}^t_i$ as their $i$-th component. Every choice of $\sigma$ s.t. the following condition is verified 
\begin{equation}
\label{eq:conditions_for_monotonic}
\mathbf{Z}^t_i \cdot \sigma_1(\mathbf{Z}^t_i + \mathbf{Y}^t_i) \geq 0 \hspace{0.5cm} \forall i = 1, ... nm',    
\end{equation}
will make Equation \eqref{eq:time_derivative} monotonically decreasing as $t \rightarrow \infty$, and the associated evolution equation would be expressive enough to induce attraction and repulsion edge-wise. What would be such a choice of non-linearity? \\ 
ReLU, or $tanh$, does not satisfy this condition, thus, we elaborate on this step by imposing the following constraints. 
\begin{equation}
    \frac{dE_{\theta}(\mathbf{H}^t)}{dt} = -\mathbf{Z}^t \sigma_1(\mathbf{Z}^t - \text{ReLU}(\mathbf{Y}^t))
\end{equation}

\begin{equation}
    \frac{dE_{\theta}(\mathbf{H}^t)}{dt} = -\mathbf{Z}^t \text{ReLU}(\text{tanh}((\mathbf{Z}^t - \text{ReLU}(\mathbf{Y}^t))
\end{equation}
The conditions in Equation \eqref{eq:conditions_for_monotonic} are always satisfied, since when $\mathbf{Z}^t_i < 0$, we have that everything is $0$. Otherwise, we consider two cases, namely when $\mathbf{Z}^t_i > \text{ReLU}(\mathbf{Y}^t_i)$ and $\mathbf{Z}^t_i < \text{ReLU}(\mathbf{Y}^t_i)$. In the former, we have that $\text{ReLU}(\text{tanh}(\cdot))$ is always positive, and also $\mathbf{Z}^t_i$. In the latter case, we have a negative argument, and the Equation goes to $0$. Generally, we could satisfy these conditions for any choice of $\sigma_1(x) \geq 0, \forall x \in \mathbb{R}$, and also for $\sigma_2 \geq 0, \forall x \in \mathbb{R}$. We conclude that through our assumptions, Equation \eqref{eq:sas-gnn_derivative_appendix} can induce attraction and repulsion edge-wise.
\subsection{Proof of Corollary \ref{corr:sigma}}
Assuming that $\sigma_1 = \text{ReLU}(\text{tanh}(\cdot))$, $\sigma_2 = \text{ReLU}(\cdot)$, and considering $\bar{\mathbf{A}}$, with no self-loops, we can enclose all the assumptions made in our Theorems' proofs. Such an instantiation of $\sigma_1$ would have a bounded derivative that guarantees the Jacobian being approximately constant as follows:
\begin{equation}
    \sigma'_1(x) = (\text{ReLU}(\text{tanh}(x)))'
\begin{cases} 
1 - \tanh^2(x) & \text{if } x > 0 \\
0 & \text{if } x \leq 0
\end{cases},
\end{equation}
then the time derivative of the energy functional will be monotonically decreasing, and the Jacobian of each node feature will be constant and with purely imaginary eigenvalues. These steps are easy to verify once the proofs of the Theorems have been understood. According to these assumptions, we conclude that Equation \eqref{eq:sas-gnn_derivative_appendix}, with $\sigma_1 = \text{ReLU}(\text{tanh}(\cdot))$, $\sigma_2 = \text{ReLU}(\cdot)$, is stable and non-dissipative, and induces attraction and repulsion edge-wise.

\subsection{Proof of Theorem \ref{theo:edge_features}}
\textit{Proof of the theorem}. Theorem \ref{theo:edge_features}, is simple to demonstrate, since $- \sigma_2(\mathbf{B}\mathbf{E}\mathbf{W}_e)$ is not dependent on the node features and can be encompassed in the $\phi_{W_s}(\mathbf{H}^t, \Gamma(i))$ term of Equation \eqref{eq:node_wise_sas_GNN}, and repeat the same proof. According to this, Equation \eqref{eq:edge_features} is stable and non-dissipative. To prove that it induces attraction and repulsion edge-wise, we can take Equation \eqref{eq:time_derivative}, and now specify that 
$\mathbf{Y}^t = - \sigma_2((\mathbf{\Omega}_{as}^\top \otimes \mathbf{I}_n)\text{vec}(\mathbf{H}^t)) - \sigma_2((\mathbf{E}^\top \otimes \mathbf{I}_n)\text{vec}(\mathbf{B}))$. We will derive the same result, since each element of $\mathbf{Y}^t$ is less than or equal to 0.\\
This concludes the proof.
\section{Early-Exit Graph Neural Networks Algorithm for Graph Classification}
\label{sec:extended_algorithm}
\textbf{Neural Adaptive Step for Graph Classification}. In Algorithm~\ref{alg:eegnn}, we describe how the Straight-Through Gumbel-Softmax estimator is integrated with SAS-GNN to allow each node to decide, at every layer, whether to exit or continue processing. However, since we also address graph classification, we refine this procedure to enable early-exit decisions at the graph level. Specifically, instead of applying the Gumbel-Softmax to individual nodes as in Algorithm~\ref{alg:eegnn}, we apply it to the entire graph, so that the model determines whether a graph’s representation is sufficiently complete to make a prediction.
This modification ensures that exit decisions align with the prediction target: nodes for node classification, graphs for graph classification. To maintain consistency, we define the "agent"—the entity responsible for triggering the exit—as the unit on which the prediction task is performed: nodes for node-level tasks and graphs for graph-level tasks.
While it is theoretically possible to perform graph classification with node-level exits (e.g., by aggregating node predictions), we leave such extensions for future work. The refined algorithm follows.

\begin{algorithm}[t]
\caption{Neural Adaptive-Step Early-Exit GNNs for Graph Classification} 
\label{alg:eegnn_gc}
\begin{algorithmic}[1]
\State Initialize $\mathcal{G} = (\mathbf{H}^{0}, \mathbf{\bar{A}}), \mathbf{E} \in \mathbb{R}^{|\mathcal{E}| \times d}, L, f_c, f_{\nu}, f_e, \text{Pool}, \mathbf{W}_{s}, \mathbf{\Omega}_{as}, \sigma_1, \sigma_2, \nu_0$
\For{$l = 0$ \textbf{to} $L$}
    \State $C^l \gets f_c(\text{Pool}(\mathbf{H}^l))$
    \State $\nu^l \gets f_{\nu}(\text{Pool}(\mathbf{H}^l, \nu_0))$
    \State $\mathbf{c}^l \gets gumbel\_softmax(C^l, \nu^l)$
    \State $\tau^l \gets \mathbf{c}^l(\texttt{0})$
    \State $\mathbf{H}^{l+1} \gets \mathbf{H}^l + \tau^l 
    \sigma_1(-\sigma_2(\mathbf{H}^l \mathbf{\Omega}_{as}) + f_e(\mathbf{E}) + \mathbf{\bar{A}} \mathbf{H}^l \mathbf{W}_s)$
    \If{$\arg\max\{\mathbf{c}^l\} = \texttt{1}$}
        \State $\mathbf{Z} \gets pool(\mathbf{H}^l)$
        \State \textbf{return} $\mathbf{Z}$
    \EndIf
\EndFor
\State $\mathbf{Z} \gets pool(\mathbf{H}^L)$
\State \textbf{return} $\mathbf{Z}$

\end{algorithmic}
\end{algorithm}
Here, we consider the case of a single graph, but it can be iterated all over the dataset $\mathcal{D}$ as well.
As we can see, $C^l$ and $\nu^l$ are scalars, conversely from node classification, since we have pooled the node representation. As a consequence, $\tau^l$ is a scalar, and is not defined node-wise anymore. Despite this, we retain the same meaning, indeed $\tau^l = 0$ leads the algorithm to stop updating the node features ($\mathbf{H}^{l+1} \gets \mathbf{H}^l$). We understand that now the personalization will be at the graph level. Indeed for graphs $i$ and $j$, we will have $\tau_i^l \neq \tau_j^l$, since these depends on $\mathbf{H}^l_i$ and $\mathbf{H}^l_j$ respectively. In the same fashion as Algorithm \ref{alg:eegnn}, we have a varying integration constant, namely $\tau^l$. This value is continuous, but we denote $l$ as the $l$-th exit point. Also, here $L$ can be seen as the number of candidate exit points. In the time domain, the total time that a graph undergoes into the GNN is computed as $\sum_{l = 0}^L \tau^l$, where each $\tau^l$ is predicted by $f_c$ and $f_{\nu}$ and it corresponds with the \textit{non-exiting} probability.

\textbf{Relation to Adaptive-Step in Runge-Kutta solvers. }
Our framework is conceptually related to adaptive-step ODE solvers such as DOPRI~\citep{DORMAND198019}. However, our approach differs in that a neural network determines the stopping condition, whereas DOPRI solvers rely on integration error tolerance. Although integrating these methods may be beneficial, it is outside the scope of this work. 

\textbf{Mitigation of Underreaching.}
EEGNN is inherently robust to under-reaching by treating the total depth $L$ as a computational budget rather than a fixed architectural constraint. As demonstrated in Figure~\ref{fig:early-exit_unstable}, increasing $L$ does not degrade performance, as the model naturally exits early when deep processing is unnecessary.
Crucially, the Neural Adaptive Step, which is driven exclusively by the task loss, tend to align with the effective depth with the problem radius. For tasks requiring long-range propagation, such as \texttt{sssp} (Figure~\ref{fig:box_plot_sssp_exit}) and \texttt{ecc} (Figure~\ref{fig:ecc_step_distributions}), the model learns to delay exits, utilizing the full depth to capture global dependencies. Conversely, on tasks with local signals like \texttt{Peptides-func} and \texttt{Peptides-struct} (Figure~\ref{fig:peptides_exit_analysis}), the distribution shifts towards shallow layers.
Unlike prior works that learn a single fixed depth for the entire dataset~\citep{AdaptiveMP}, our approach adapts depth dynamically per sample. It is true that our method is not tailored for mitigating underreaching, but \cite{AdaptiveMP} is complementary to our effort, as we could use methodologies learn the budget $L$ without changing our model. 

\section{Additional Details on the Experimental Set-Up}
\label{sec:datasets}
\subsection{Datasets}
Table \ref{tab:datasets} lists some statistics of the heterophilic benchmark that we use. They are computed after the graphs were made undirected, which is a standard procedure with GNNs. 
The datasets selected belong to a recent collection \citep{criticallookatgnn}, which was proposed to enrich the current dataset availability for the GNN experimental setting under heterophily. $\xi_{edge}$ (edge homophily) and $\xi_{adj}$ (adjusted homophily) are metrics that estimate the heterophily level of a graph; the lower they are, the more the graph is considered as heterophilic. For a deeper understanding of these metrics, we suggest the interested reader refer to the original paper \citep{criticallookatgnn}.
\begin{table}[ht]
    \centering 
    \caption{Dataset Information}
    \begin{tabular}{lcccccc}
        \hline
        \textbf{Datasets} & $N$ & $|\mathcal{E}|$ & $d$ & $|C|$ & $\xi_{edge}$ & $\xi_{adj}$ \\
        \hline
        \texttt{Amazon Ratings}    & 24492 &186100 & 300 & 5& 0.38& 0.14\\
        \texttt{Roman Empire}      & 22662& 65854 & 300 & 18 & 0.05 & -0.05\\
        \texttt{Minesweepers}       & 10000 &78804  &7 & 2& 0.68 & 0.01\\
        \texttt{Questions}         & 48921 & 307080 & 301 & 2& 0.84& 0.02\\
        \texttt{Tolokers}          & 11758 &1038000 & 10 & 2& 0.59& 0.09\\
        \hline
    \end{tabular}
    
    \label{tab:datasets}
\end{table}
\begin{table}[ht]
\centering

\caption{ECHO benchmarks statistics.}
\label{tab:echo_data_statistics}
\resizebox{\textwidth}{!}{
\begin{tabular}{l c c c c c c c c}
\toprule
\textbf{Dataset} & \textbf{\# Graphs} & \textbf{Avg Nodes} & \textbf{Avg Deg.} & \textbf{Avg Edges} & \textbf{Avg Diam} & \textbf{\# Node Feat} & \textbf{\# Edge Feat} & \textbf{\# Tasks} \\
\midrule
Graph Topologies & 10,080 & $83.69 \pm 66.24$ & $2.53 \pm 1.19$ & $211.63 \pm 209.39$ & $28.50 \pm 6.92$ & 2 & None & 3 \\
\midrule
\texttt{line} & 1,680 & $75.60 \pm 27.32$ & $2.37 \pm 0.10$ & $90.10 \pm 33.89$ & $28.50 \pm 6.92$ & 2 & None & 3 \\
\texttt{ladder} & 1,680 & $56.52 \pm 13.82$ & $2.92 \pm 0.02$ & $82.54 \pm 20.72$ & $28.50 \pm 6.92$ & 2 & None & 3 \\
\texttt{grid} & 1,680 & $193.10 \pm 93.10$ & $2.95 \pm 0.12$ & $288.32 \pm 145.29$ & $28.50 \pm 6.92$ & 2 & None & 3 \\
\texttt{tree} & 1,680 & $60.42 \pm 17.17$ & $1.96 \pm 0.01$ & $59.42 \pm 17.17$ & $28.50 \pm 6.92$ & 2 & None & 3 \\
\texttt{caterpillar} & 1,680 & $34.71 \pm 7.96$ & $1.94 \pm 0.02$ & $33.71 \pm 7.96$ & $28.50 \pm 6.92$ & 2 & None & 3 \\
\texttt{lobster} & 1,680 & $81.79 \pm 25.46$ & $1.97 \pm 0.01$ & $80.79 \pm 25.46$ & $28.50 \pm 6.92$ & 2 & None & 3 \\
\bottomrule
\end{tabular}%
}
\end{table}
In Table \ref{tab:echo_data_statistics}, we report the statistics associated with the ECHO benchmark. In particular we describe the input node and edge features, the graph structural properties such as average degree, average edges, and average diameter, and we report the associated graph topology. As remarked in the main paper the topologies are structured in a way that the tasks can be solved only through a correct and effective long-range reasoning. This is due to the nature of the task but also their structure. These graphs contains topological bottlenecks because of the type of generated topology, and this is a good way of measuring how models behave in scenario where topological over-squashing, and also computational over-squashing \cite{demystifyingcommonbeliefs}. combined may arise. The tasks that are considered are Single-Source-Shortest Path (\texttt{sssp}), which predicts the shortest path across all the nodes, then we have Eccentricity (\texttt{ECC}), builds on this by computing the longest shortest path from each node. Finally Diameter (\texttt{diam}), is a task where we look for
the longest shortest path between any two nodes. All of these tasks require to span information all over the graph, implying global understanding, and long-range reasoning form the modle. This benchmark is ad-hoc for long-range interaction w.r.t. the existing and widely adopted LRGB \cite{LRGB}.

The dataset released by \citep{LRGB} instead has the statistics reported in Tables \ref{tab:dataset_tasks} and \ref{tab:lrgb_statistics}.
\begin{table}[t]
    \centering
    \caption{Dataset details, tasks, and performance metrics.}
    \label{tab:dataset_tasks}
    \resizebox{\textwidth}{!}{%
    \begin{tabular}{l c c c c c}
        \hline
        \textbf{Dataset} & \textbf{Domain} & \textbf{Task} & \textbf{Node Features (dim)} & \textbf{Edge Features (dim)} & \textbf{Performance Metric} \\
        \hline
        \texttt{PascalVOC-SP}    & Computer Vision & Node Classification & Pixel + Coord (14) & Edge Weight (1 or 2) & macro F1 \\
       \texttt{Peptides-func}   & Chemistry & Graph Classification & Atom Encoder (9) & Bond Encoder (3) & AP \\
        \texttt{Peptides-struct} & Chemistry & Graph Regression & Atom Encoder (9) & Bond Encoder (3) & MAE \\
        \hline
    \end{tabular}%
    }
\end{table}

\begin{table}[t]
    \centering
    \caption{Statistics of the proposed LRGB datasets.}
    \label{tab:lrgb_statistics}
    \resizebox{\textwidth}{!}{%
    \begin{tabular}{l c c c c c c c c}
        \hline
        \textbf{Dataset} & \textbf{Total Graphs} & \textbf{Total Nodes} & \textbf{Avg Nodes} & \textbf{Mean Deg.} & \textbf{Total Edges} & \textbf{Avg Edges} & \textbf{Avg Shortest Path} & \textbf{Avg Diameter} \\
        \hline
        \texttt{PascalVOC-SP}    & 11,355  & 5,443,545  & 479.40  & 5.65  & 30,777,444  & 2,710.48  & $10.74 \pm 0.51$ & $27.62 \pm 2.13$ \\
        \texttt{Peptides-func}   & 15,535  & 2,344,859  & 150.94  & 2.04  & 4,773,974   & 307.30    & $20.89 \pm 9.79$ & $56.99 \pm 28.72$ \\
        \texttt{Peptides-struct} & 15,535  & 2,344,859  & 150.94  & 2.04  & 4,773,974   & 307.30    & $20.89 \pm 9.79$ & $56.99 \pm 28.72$ \\
        \hline
    \end{tabular}%
    }
\end{table}
\subsection{Baselines}
\noindent \textbf{Baselines for heterophilic graphs}.
We consider the set of heterophilic graphs introduced in \cite{criticallookatgnn}. We test our models in heterophilic settings, since these are scenarios that typically cause over-smoothing, and we want to gauge in practice how they compare against existing models. We briefly describe the baselines used for comparison against SAS-GNN and EEGNN in node classification. From the benchmark introduced in \citep{criticallookatgnn}, we include the reported performances of GCN, GraphSAGE \citep{GCN, GraphSAGE}, GAT \citep{GAT}, and GT \citep{GT}. Additionally, we consider GAT-sep and GT-sep, two variants of these models specifically designed for heterophily. Even though these baselines seem outdated, it is known that the experiments in \citep{criticallookatgnn} revealed that classic models outperform methods specialized for heterophily. More recently, \citet{coognns} introduced the Cooperative Graph Neural Networks (Co-GNNs) model, which we also include in our evaluation. This family of models is also more expressive than classic message-passing neural networks \citep{WL1}, thanks to an asynchronous message-passing.\\

\noindent \textbf{Baselines for ECHO Benchmark.}
We categorize baselines into four groups: (i) standard MPNNs (GCN, GIN \cite{WL1}, GCNII \cite{chen2020simpledeepgraphconvolutional}), here we have the message-passing paradigm where each layer dictates how far information propagates; (ii) rewiring methods as DRew \cite{Drew}; (iii) Graph Transformers (GPS \cite{graphgps}, GRIT \cite{ma2023graphinductivebiasestransformers}) leveraging global attention; and (iv) related Graph NODEs (GraphCON \cite{graphcon}, A-DGN, SWAN \cite{swan}, PH-DGN \cite{PHGNN}), the class to which SAS-GNN and EEGNN belong.
Additionally, we include APGCN \cite{AdaProp}, a representative early-exit GNN, to contrast our task-driven mechanism against its budget-aware halting. We also employ GRAFF~\cite{GRAFF2} as a symmetric-only ablation, complementing the antisymmetric dynamics of A-DGN.

\noindent \textbf{Baselines for Long Range Graph Benchmark.}
We compare SAS-GNN and EEGNN on 3 datasets from LRGB \cite{LRGB}. These datasets were collected to assess how much GNNs can preserve information exchanged among distant nodes. We take advantage of these datasets to test how much our model performs compared to other methods.
Among the baselines, we compare \textit{Classic MPNNs}: GCN \cite{GCN}, GIN \cite{WL1}, GINE \cite{GINE}, GatedGCN \cite{gatedgcn}, which is the category that encompasses SAS-GNN and EEGNN. These methods need to process $l$ message-passing steps to allow nodes distant $l$  to communicate. \textit{Graph Transformers}: \cite{GT}, SAN \cite{SAN}, and GraphGPS \cite{graphgps}, this category uses self-attention to all the nodes, so it allows nodes to reach every part of the graph, but it has a quadratic complexity w.r.t $|\mathcal{V}|$. \textit{Rewiring methods}: DIGL \cite{DIGL}, MixHop \cite{mixhop}, and DRew \cite{Drew}, these approaches modify the graph topology during the message-passing, in this way messages can be delivered through paths that did not exist before, and so long-range interactions are fostered. \textit{Asynchronous Message-Passing methods}: Co-GNN \cite{coognns}, AMP \cite{AdaptiveMP}. These can be seen similarly to graph rewiring, since the computational graph is different from the input graph, as well as rewiring methods. However, here there exists the interpretation of nodes that decides what messages to filter, rather than rewiring the graph.

We test our approach only on 3 out of 5 datasets, because of computational constraints. These are \texttt{Peptides-func}, \texttt{Peptides-struct}, and \texttt{Pascal-VOC}. Results are averaged across 4 different weight initializations. 
\subsection{Implementation Details}
\label{subsec:implementation}
The metrics that we use are the \textit{Area Under the Curve} (AUC) for the binary classification experiments, accuracy for \texttt{Roman Empire} and \texttt{Amazon Ratings}, average precision (AP) for \texttt{Peptides-func}, mean-absolute error (MAE) for \texttt{Peptides-struct}, and macro F1-Score (F1) for \texttt{Pascal-VOC}. All the experiments concerning EEGNN and SAS-GNN implied that we removed self-loops from the graphs, in order to satisfy the assumptions made in Theorems \ref{theo:stability}, \ref{theo:energy}. We always apply $f(\mathbf{X})$ implemented via a one-layer perception with ReLU. And in the LRGB, we also use a final decoder MLP to refine the representations of $\mathbf{Z}$, details on the hyperparameter are in the code.
In the heterophilic graphs, each training lasts 3000 epochs, and the optimizer that we use is Adam \citep{adam}. Concerning the quantitative results, we report the mean and standard deviation, which are computed from the metric averaged across 10 runs. Each run corresponds to a different split of the nodes used for training, validation, and testing.
These experiments were run on a single Nvidia GeForce RTX 3090 Ti 24 GB and also a single NVIDIA A100 80GB PCIe. 

The experiments on ECHO, share same optimizers, metrics, and task objectives (except for APGCN which uses an additional loss term). There we run an hyperparameter optimization via optuna \cite{optuna} with 32 trials per model, and 1000 epochs per trial. The results we present are averaged across 4 weights initialization.  

As concerns the LRGB experiments, we run each configuration for a maximum of 1000 epochs across 4 different weights initializations, using the AdamW optimizer. We picked the best configuration based on the validation metric. These experiments were run on a single NVIDIA A100 80GB PCIe. 

As concerns with the experiments on the TUDatasets for short-range graph classification (see Table \ref{tab:results_tudataset}) and homophilic node classification (see Section \ref{sec:extended_homophily}), we also used a single NVIDIA A100 80GB PCIe. All the data was taken mostly from the PyTorch Geometric library \citep{pytorchgeometric}, and other open-source sources. We always refer to their provenance in the code when using them.

\smallskip
\subsection{Hyperparameters}
\label{subsec:hyperparameters}
For a fair comparison, the hyperparameter tuning process was done according to the hyperparameter set used by \citet{coognns}. In the SAS-GNN experiments, we choose $\tau$ among $\{0.1, 0.5, 1\}$. While in the experiments with EEGNNs, we do not need to tune $\tau$ because it is adaptive (see Algorithm \ref{alg:eegnn}), and we do not need to tune the number of layers parameter since we use a large and upper bound depth $L = 20$. In the case of EEGNN, we choose $f_c(\cdot)$ as a Mean-GNN \citep{GRL_book}. Similarly to \cite{coognns}, we test a contained number of layers $\{1, 2 ,3\}$, and then as hidden dimension we select $4$, $8$, $16$, $32$ or $64$. Then we also tune the value of the temperature $\nu_0$ between $0$ and 0.1.  $f_{\nu}(\cdot)$, the network that learns the adaptive temperature $\nu$, adopts the design proposed in \citep{coognns}.
In the ECHO benchmark, we performed the hyperparameter tuning based on the following hyperparameter ranges in Tables \ref{tab:eegnn_hyperparams}, \ref{tab:sasgnn_hyperparams}, \ref{tab:graff_hyperparams}, \ref{tab:apgcn_hyperparams}. There we also consider the best values per-dataset. 

\begin{table}[h!]
\centering
\caption{Hyperparameter search space and best configurations for \textbf{EEGNN (Adastep)} on ECHO tasks. $\tau$ and $L$ are fixed as they depends on node features at each layer.}
\label{tab:eegnn_hyperparams}
\resizebox{0.6\columnwidth}{!}{%
\begin{tabular}{lcccc}
\toprule
\textbf{Hyperparameter} & \textbf{Search Space} & \textbf{SSSP} & \textbf{Eccentricy} & \textbf{Diameter} \\
\midrule
$L$ & 40 & 40 & 40 & 40 \\
$m'$ & $[32, 256]$ & 69 & 57 & 254 \\
$m'_f$ & $[32, 128]$ & 87 & 55 & 113 \\
$L_f$ & $[1, 40]$ & 1 & 14 & 24 \\
Learning Rate & $[10^{-5}, 10^{-2}]$ & 0.001907 & 0.006095 & 0.003307 \\
Weight Decay & $[10^{-8}, 10^{-3}]$ & 0.000919 & 0.000133& 0.000744 \\
$\tau$ & 1.0 & 1.0 & 1.0 & 1.0 \\
$L_{\nu}$ & 2 & 2 & 2 & 2 \\
$\nu_0$ & 1.0 & 1.0 & 1.0 & 1.0 \\
\bottomrule
\end{tabular}%
}
\end{table}

\begin{table}[h!]
\centering
\caption{Hyperparameter search space and best configurations for \textbf{SAS-GNN} on ECHO tasks.}
\label{tab:sasgnn_hyperparams}
\resizebox{0.6\columnwidth}{!}{%
\begin{tabular}{lcccc}
\toprule

\textbf{Hyperparameter} & \textbf{Search Space} & \textbf{SSSP} & \textbf{Eccentricity} & \textbf{Diameter} \\
\midrule
$L$ & $[1, 40]$ & 32 & 27 & 19 \\
$m'$ & $[32, 256]$ & 50 & 196 & 66 \\
Learning Rate & $[10^{-5}, 10^{-2}]$ & 0.001907 & 0.000556 & 0.006141 \\
Weight Decay & $[10^{-8}, 10^{-3}]$ & 0.000270 & 0.000246 & 0.000165 \\
$\tau$ & $[0.01, 1.0]$ & 0.000165 & 0.09946355197568157 & 0.4037 \\
\bottomrule
\end{tabular}%
}
\end{table}

\begin{table}[h!]
\centering
\caption{Hyperparameter search space and best configurations for \textbf{GRAFF} on ECHO tasks.}
\label{tab:graff_hyperparams}
\resizebox{0.6\columnwidth}{!}{%
\begin{tabular}{lcccc}
\toprule
\textbf{Hyperparameter} & \textbf{Search Space} & \textbf{SSSP} & \textbf{Eccentricity} & \textbf{Diameter} \\
\midrule
$L$ & $[1, 40]$ & 21 & 15 & 9 \\
$m'$ & $[32, 256]$ & 37 & 32 & 123 \\
Learning Rate & $[10^{-5}, 10^{-2}]$ & 0.007387 & 0.006607 & 0.003168 \\
Weight Decay & $[10^{-8}, 10^{-3}]$ & 0.000259 & 0.000134 & 0.000077 \\
$\tau$ & $[0.01, 1.0]$ & 0.3253 & 0.8447 & 0.8447 \\
Self Loops & False & False & False & False \\
\bottomrule
\end{tabular}%
}
\end{table}

\begin{table}[h!]
\centering
\caption{Hyperparameter search space and best configurations for \textbf{APGCN} on ECHO tasks. Here $L$ is fixed, as we adapt the value at each layer, based on the node features. $L_{enc}$ stands for number of layers that the encoder of input node features has. $\mu$ is the multiplying coefficient of the budget-aware loss.}
\label{tab:apgcn_hyperparams}
\resizebox{0.6\columnwidth}{!}{%
\begin{tabular}{lcccc}
\toprule
\textbf{Hyperparameter} & \textbf{Search Space} & \textbf{SSSP} & \textbf{Eccentricity} \\
\midrule
$L$ & 40 & 40 & 40 \\
$m'$ & $[32, 256]$ & 183 & 183  \\
Learning Rate & $[10^{-5}, 10^{-2}]$ & 0.002801 & 0.003853  \\
Weight Decay & $[10^{-8}, 10^{-3}]$ & 0.000672 & 0.000997  \\
$L_{enc}$ & 5 & 5 & 5 \\
$\mu$ & $[10^{-4}, 10^{-2}]$ & 0.0054 & 0.0069 \\
\bottomrule
\end{tabular}%
}
\end{table}

In this case the exit and temperature networks, namely $f_c$ and $f_{\nu}$, were implemented using another SAS-GNN, so the overall parameter complexity can be simplified to $\mathcal{O}\!\left(m'^2 + m_f'^2\right)
$.
In the experiments concerning the LRGB datasets, $f_c$ and $f_{\nu}$ keep the same hyperparameter set as the heterophilic experiments. The main hyperparameter set is the same used in \cite{MPNN_good_lrgb}. In any experiment with SAS-GNN and EEGNN, we do not use any dropout or normalization layer within the message-passing updates, to follow the ODE interpretation and directly minimize the energy functional in Equation \eqref{eq:energy_functional}. We only use them in the feature encoder and in the final decoder. 

In the experiments with \texttt{Peptides-func} and \texttt{Peptides-struct}, we use two positional encoding techniques, which are typically approached in these settings. For \texttt{Peptides-func} we used RWSE \citep{RWSE}, while instead in \texttt{Peptides-struct}, we used LapPE \citep{LAP}. We used these according to the best hyperparameters used for graph convolutional methods in \cite{MPNN_good_lrgb}. We do not use positional encodings in \texttt{Pascal-VOC}.
\section{Additional Results}
\label{sec:extended_results}
In this section of the appendix we are going to include additional results that are incremental w.r.t. the claims made in the main paper. We show empirical validation of our theoretical properties, and also prove that both in synthetic and real-world experiments representative of long-range tasks and highly heterophilic graphs, SAS-GNN showcases its inductive bias, indicating that are good proxy to mitigate over-squashing and over-smoothing. We also provide complementary experiments that further support our claims or prove additional aspects, such as the scalability of our model.\\ 
We then move to provide a broader view on the real-world experiments presented in Tables \ref{tab:extended_result_heterophilic} and \ref{tab:result_lrgb_extended}. Then we also provide additional experiments on homophilic node classification, short-range graph classification on both medium-size and large scale datasets. Other content of this section are ablation and sensitivity studies on the early-exit components. 
\subsection{Empirical Validation of the SAS-GNN Theoretical Properties}
\label{sec:theory_results}
\begin{figure}[t]
    \centering
     \begin{subfigure}[t]{0.35\textwidth}
        \centering
        \includegraphics[width=\linewidth]{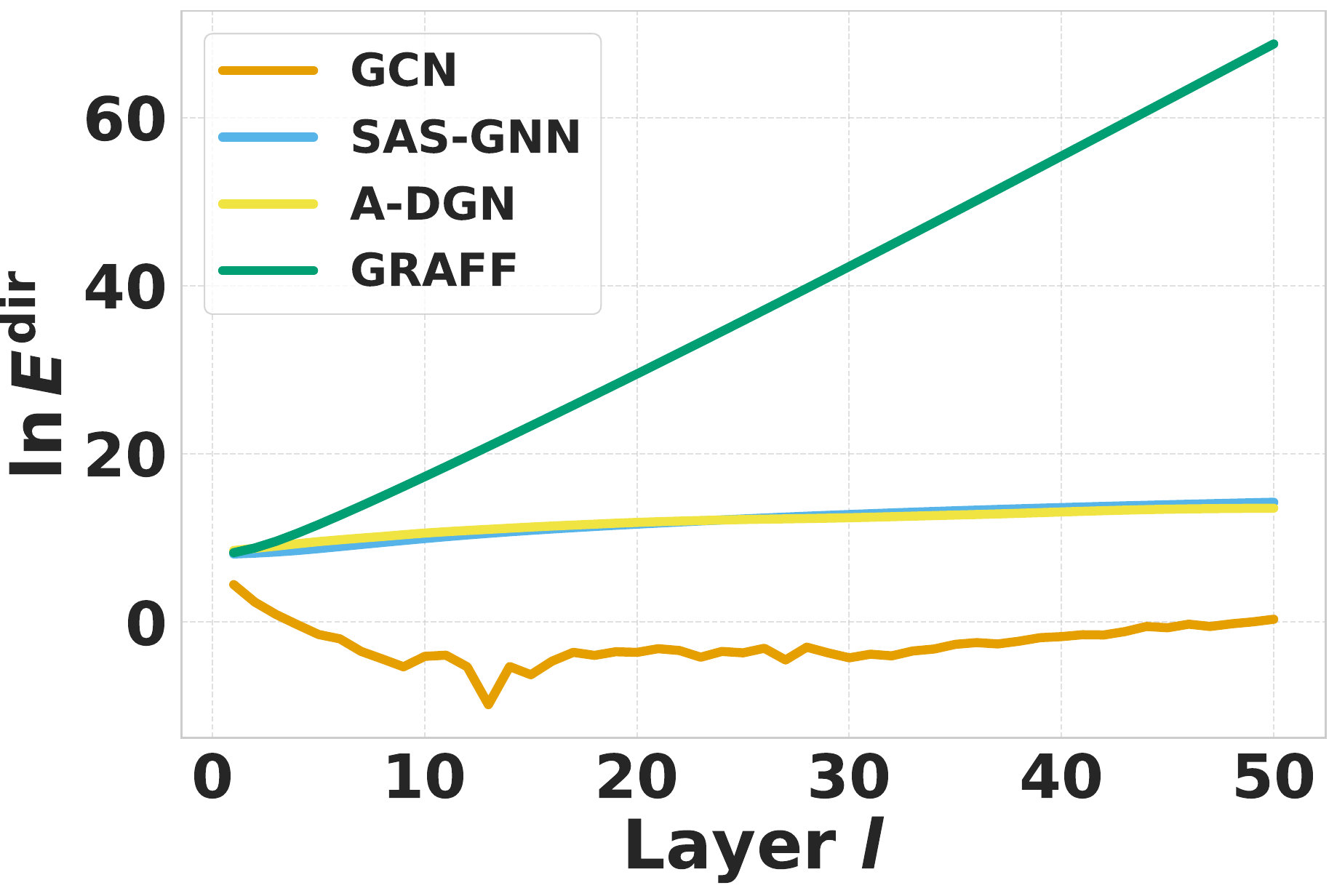}
        \caption{$E^{dir}$ on \texttt{Minesweeper}.}
        \label{fig:dirichlet_performance}
    \end{subfigure}
    \quad
    \begin{subfigure}[t]{0.35\textwidth}
        \centering
        \includegraphics[width=\linewidth]{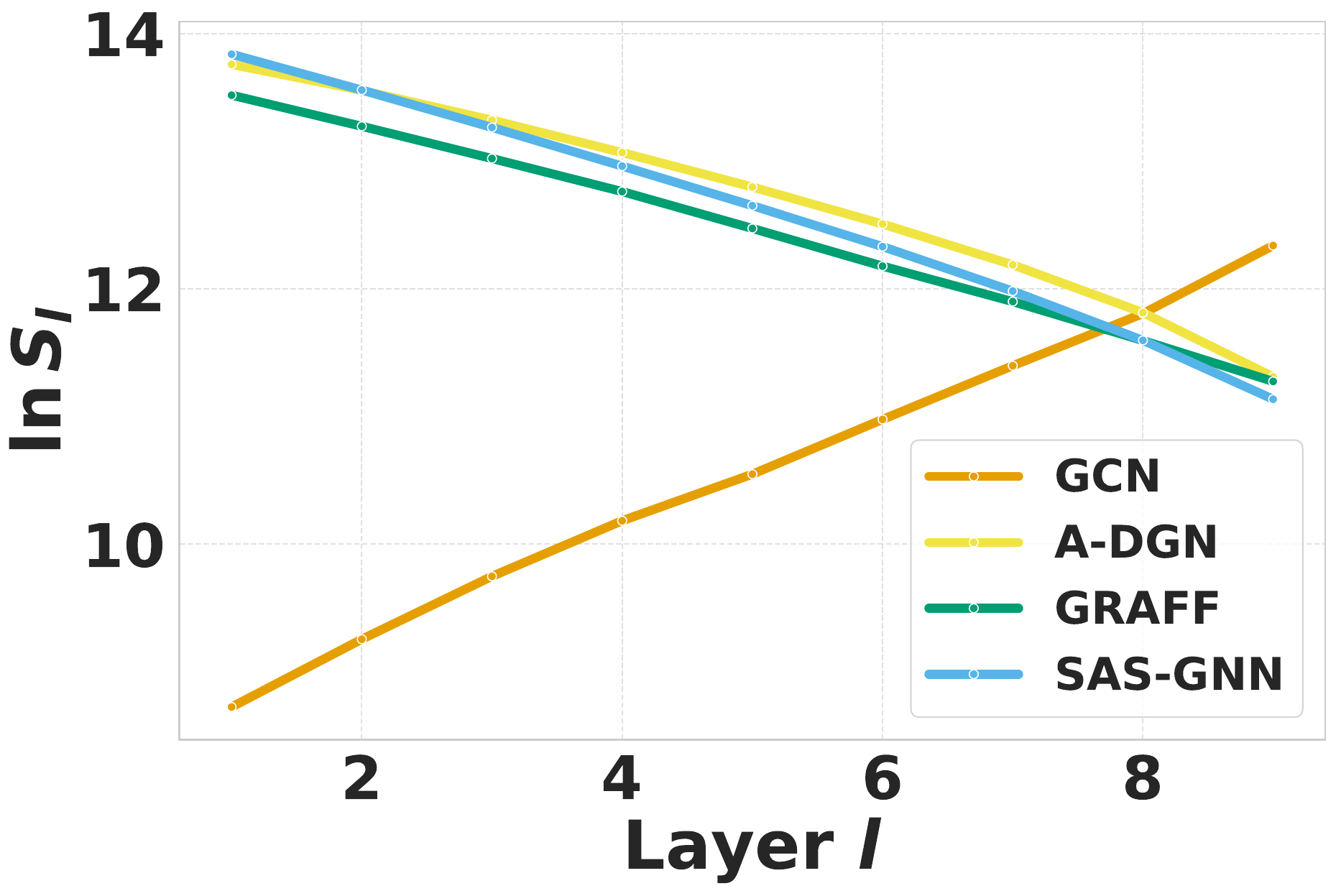}
        \caption{ln$S_{l}$ on \texttt{Peptides-func}.}
        \label{fig:func_sensitivity}
    \end{subfigure}
\end{figure}

Having introduced the new SAS-GNN backbone, and proved the theorem behind its architectural choices, we now turn to validating the theoretical claims underlying the SAS-GNN block. We already showed in the main text that SAS-GNN is capable of deep processing, and long-range reasoning. However, we want to understand whether this also translates in possible OST and OSQ mitigation, namely, do the SAS-GNN inductive biases also a good proxy to mitigate over-smoothing and over-squashing? We believe this might be the case according to the theoretical properties expressed in Section \ref{sec:method}.\\
\textbf{Over-smoothing}. To assess OST, we monitor the Dirichlet Energy $E^{dir}(\mathbf{H})$, which quantifies feature smoothness over a graph:  
\begin{equation}
    E^{dir}(\mathbf{H}) = \sum_{(i,j) \in \mathcal{E}} \| (\nabla \mathbf{H})_{ij} \|^2, \quad (\nabla \mathbf{H})_{ij} := \frac{\mathbf{h}_j}{\sqrt{d_{j}+1}} -  \frac{\mathbf{h}_i}{\sqrt{d_{i}+1}}.
\end{equation}
Lower values of $E^{dir}$ indicate smoothing; values near zero signal OST~\citep{oversmoothingdirichlet}. In Figure~\ref{fig:dirichlet_performance}, we report $E^{dir}(\mathbf{H}^t)$ at each layer on \texttt{Minesweeper}, a heterophilic dataset. GCN rapidly collapses, while GRAFF oversharpens, as described in \citet{GRAFF2}. SAS-GNN and A-DGN, in contrast, maintain stable energy profiles, suggesting SAS-GNN inherits its stabilizing behavior primarily from A-DGN. \\
\textbf{Over-squashing}. We evaluate OSQ using layer-wise sensitivity $S_l$, defined as $S_l = \sum_{(v, u) \in \mathcal{E}} 
    \left\lVert 
    \frac{\partial \mathbf{h}_v^L}{\partial \mathbf{h}_u^l} 
    \right\rVert_1$,
which measures intermediate embeddings affects final representations. Due to its computational cost, we compute $S_l$ on one test graph. Results on \texttt{Peptides-func} (Figure~\ref{fig:func_sensitivity}) show that GCN outputs are more affected by the latest layers, indicating limited early-layer influence. In contrast, GRAFF, A-DGN, and SAS-GNN maintain higher sensitivity in early layers, suggesting a more balanced propagation of information.\\
These trends affirm that SAS-GNN is effective as a proxy to contrast OST and OSQ. We must notice that the sensitivity metric do not directly distinguish among the types of OSQ, so we just show this as a way of highlighting the general advantage of SAS-GNN w.r.t. existing methods.

\begin{figure}[t]
    \centering
    \begin{subfigure}[b]{0.45\textwidth}
        \centering
        \includegraphics[width=\linewidth]{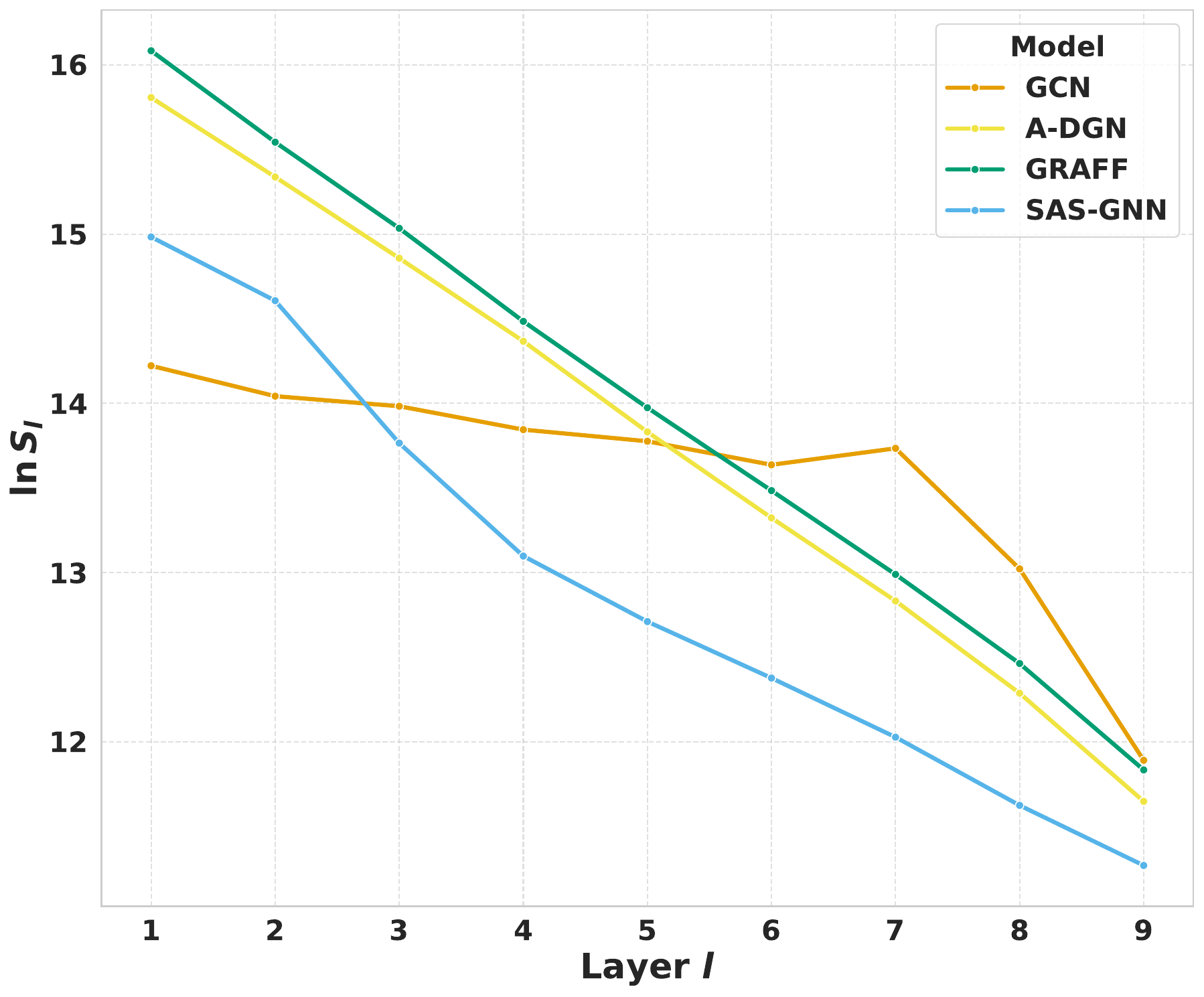}
        \caption{\texttt{Pascal-VOC}: Sensitivity analysis in node classification on LRGB dataset.}
        \label{fig:pascal_sensitivity}
    \end{subfigure}
    \hfill
    \begin{subfigure}[b]{0.45\textwidth}
        \centering
        \includegraphics[width=\linewidth]{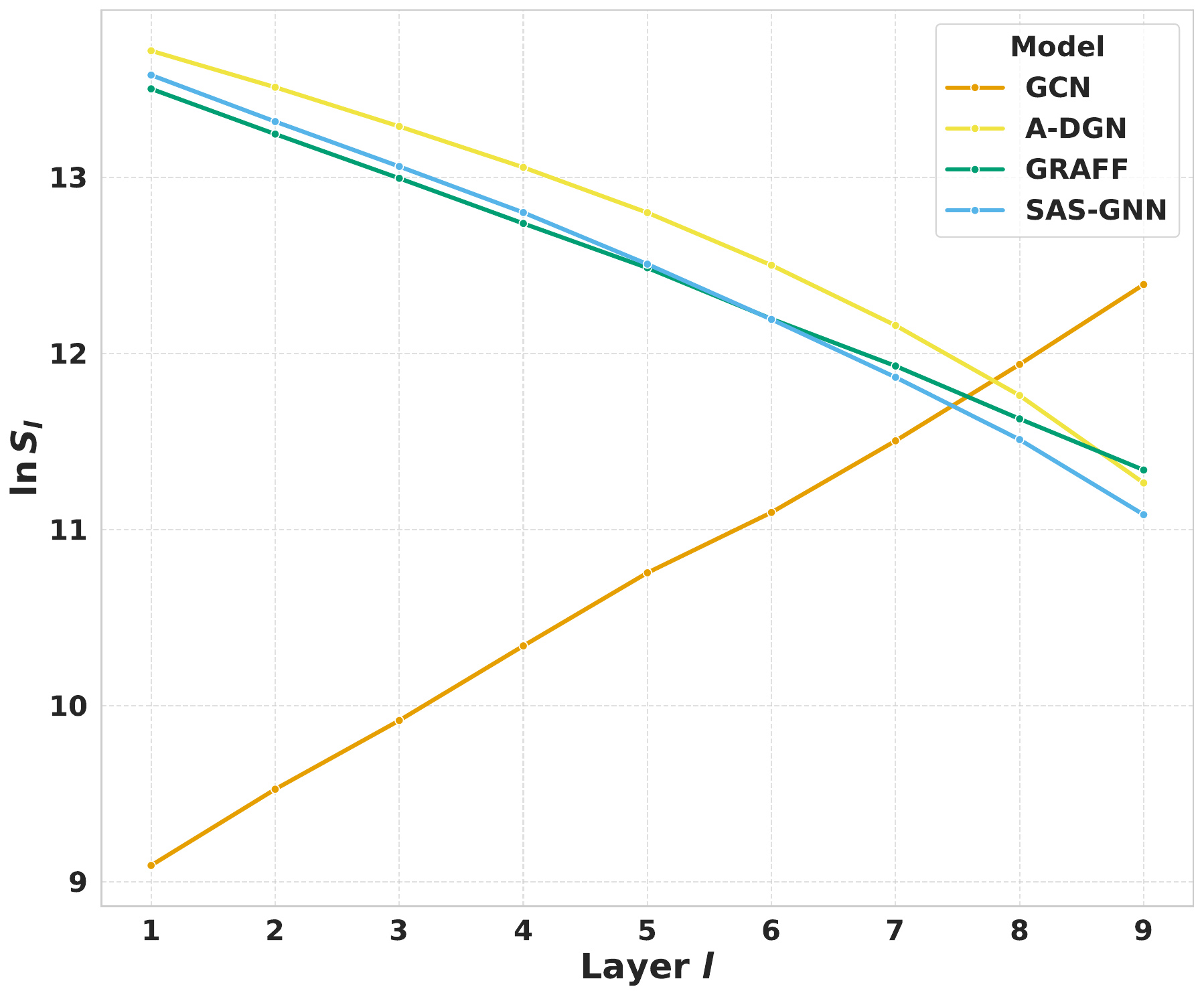}
        \caption{\texttt{Peptides-struct}: Sensitivity analysis in node classification on LRGB dataset.}
        \label{fig:peptides_struct_sensitivity}
    \end{subfigure}
    \caption{Comparison of models' performance on different datasets with increasing depth.}
    \label{fig:appendix_sensitivity}
\end{figure}

\begin{figure}[ht]
    \centering
    \begin{subfigure}[b]{0.45\textwidth}
        \centering
        \includegraphics[width=\textwidth]{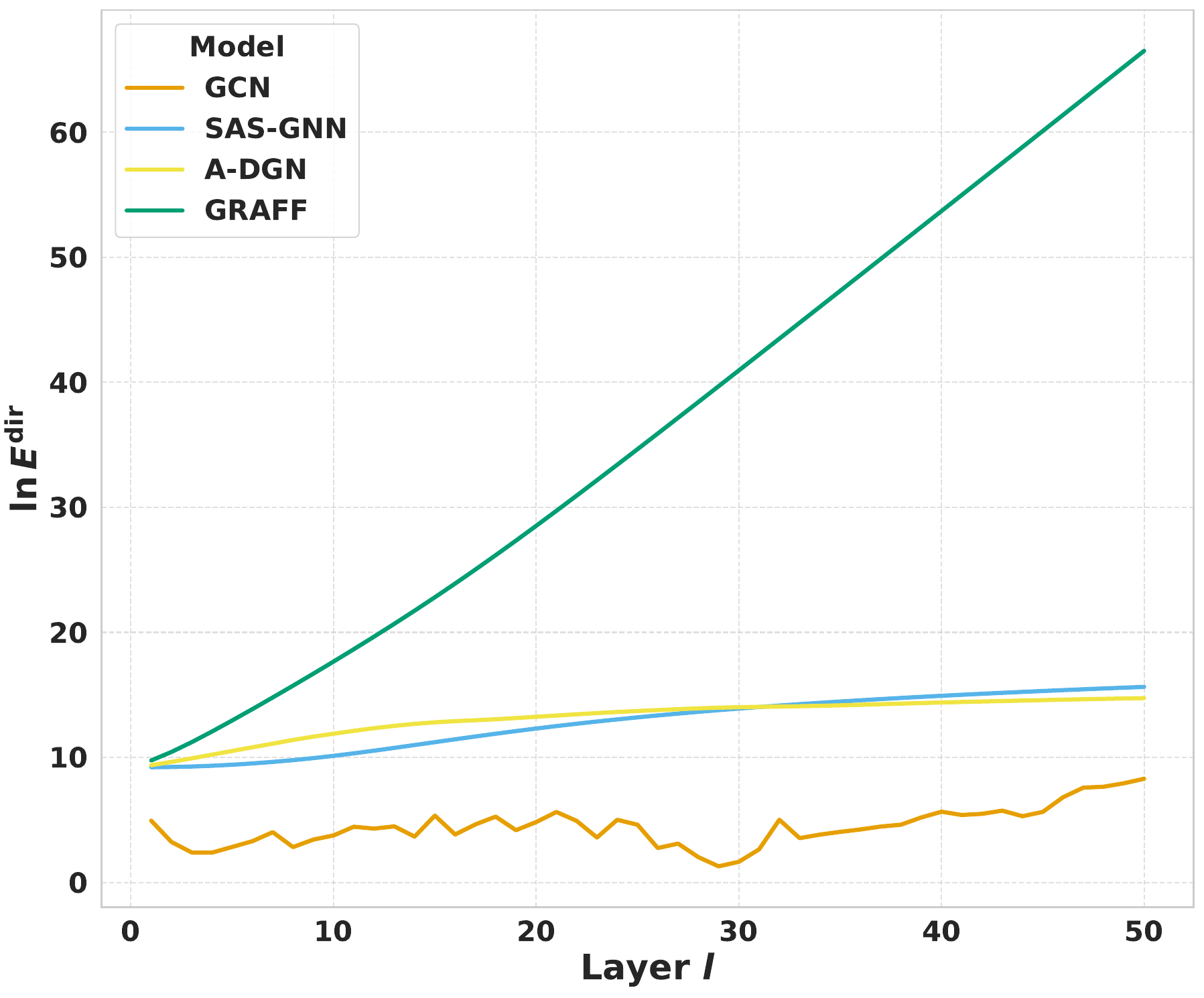}
        \caption{$E^{dir}$ trend for \texttt{Tolokers}.}
        \label{fig:tolokers_dirichlet}
    \end{subfigure}
    \hfill
    \begin{subfigure}[b]{0.45\textwidth}
        \centering
        \includegraphics[width=\textwidth]{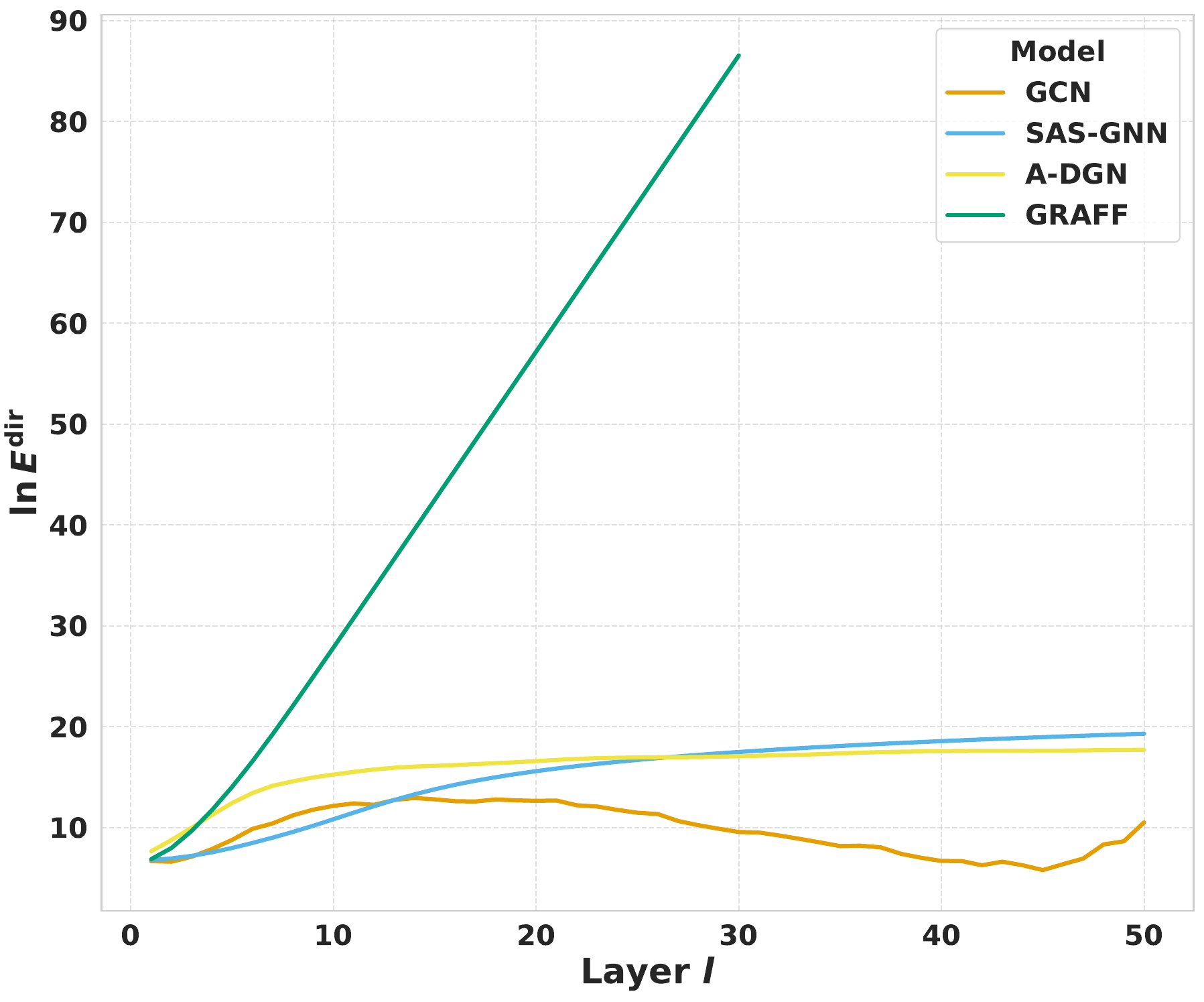}
        \caption{$E^{dir}$ trend for \texttt{Questions}.}
        \label{fig:questions_dirichlet}
    \end{subfigure}

    \vskip\baselineskip  

    \begin{subfigure}[b]{0.45\textwidth}
        \centering
        \includegraphics[width=\textwidth]{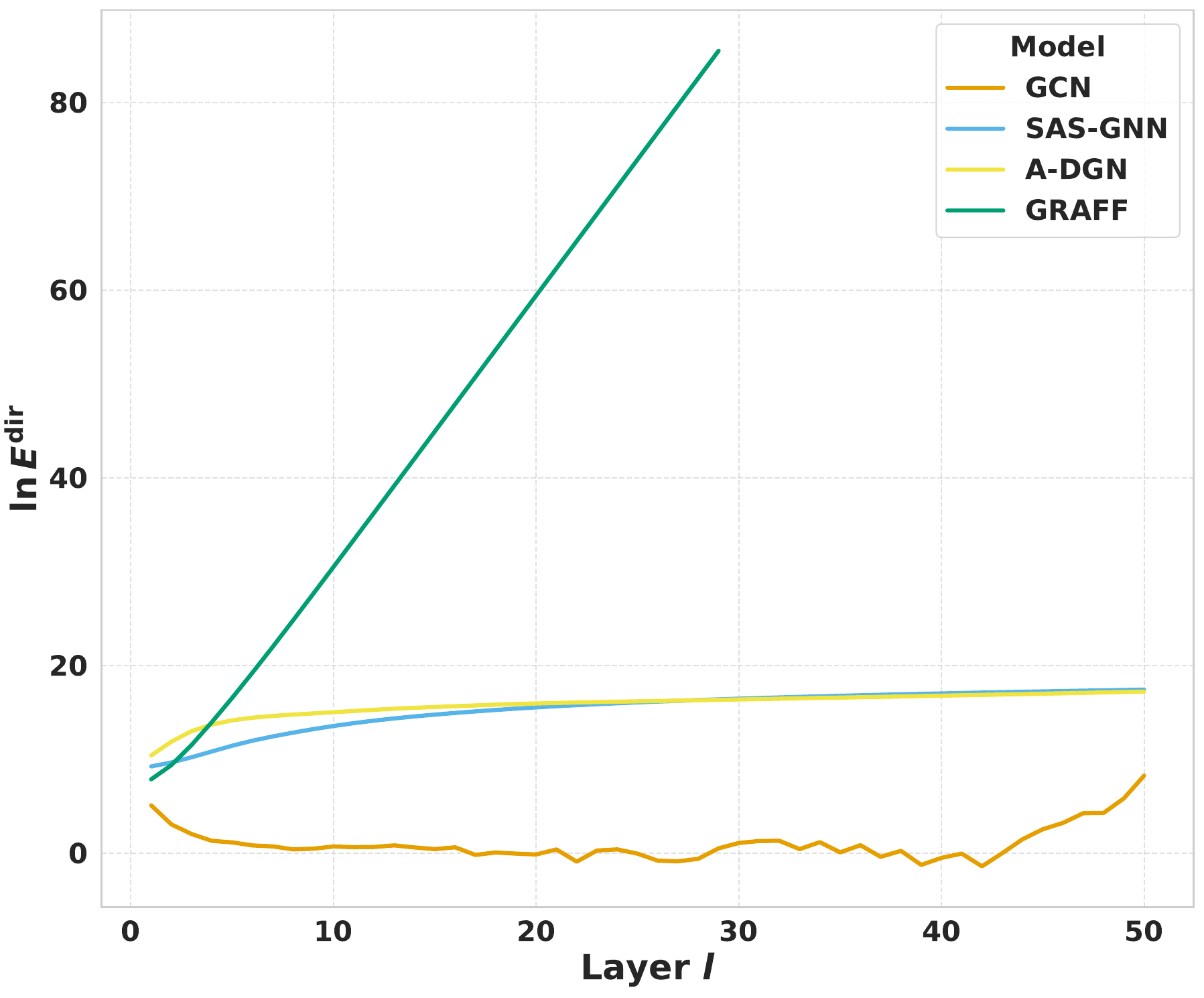}
        \caption{$E^{dir}$ trend for \texttt{Roman Empire}.}
        \label{fig:roman_empire_dirichlet}
    \end{subfigure}
    \hfill
    \begin{subfigure}[b]{0.45\textwidth}
        \centering
        \includegraphics[width=\textwidth]{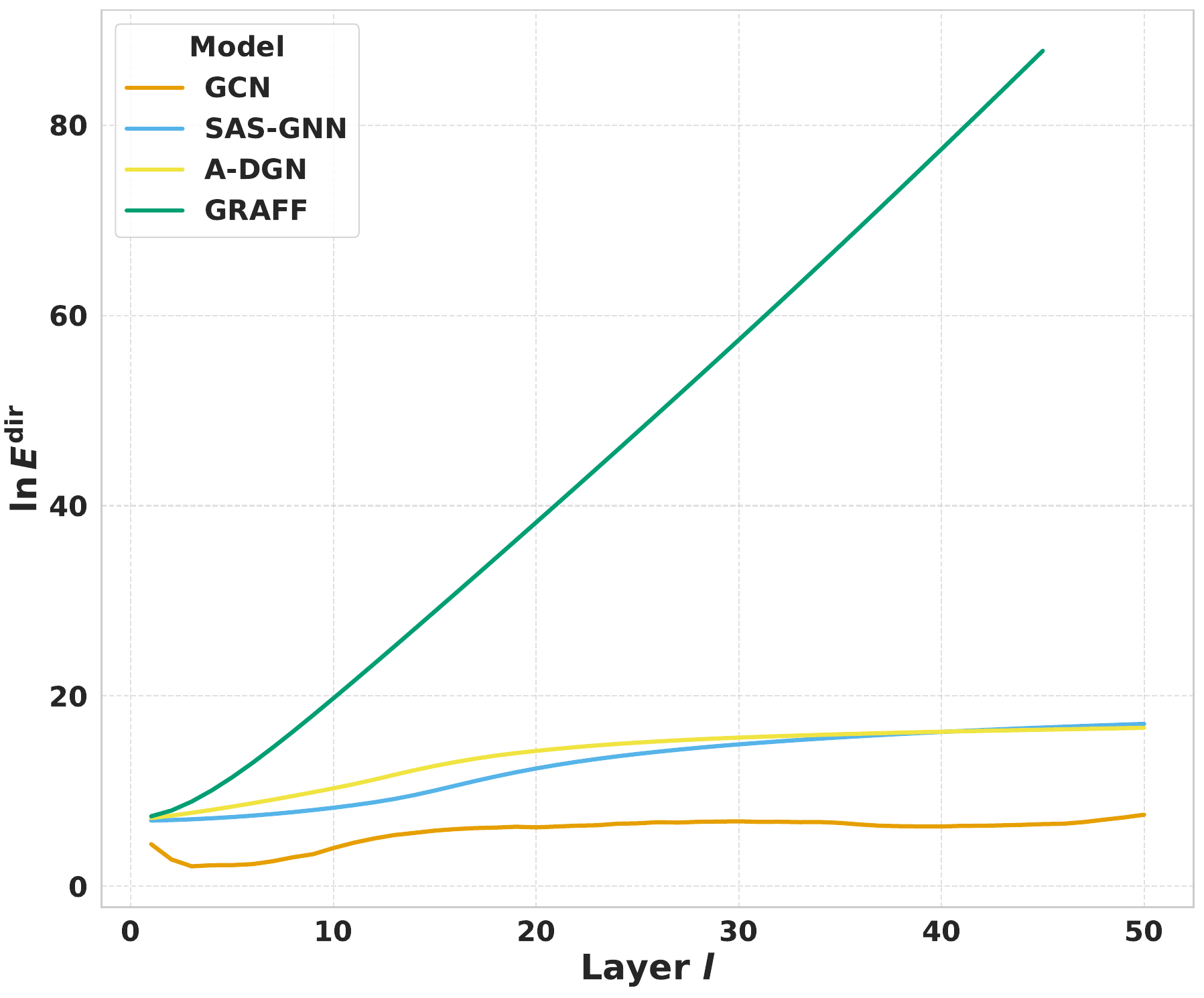}
        \caption{$E^{dir}$ trend for \texttt{Amazon Ratings}.}
        \label{fig:amazon_ratings_dirichlet}
    \end{subfigure}

    \caption{$E^{dir}$ trends for heterophilic graphs, with associated accuracies.}
    \label{fig:appendix_dirichlet}
\end{figure}

We provide additional plots of Dirichlet Energy and layer-wise sensitivity to witness the model's resilience to over-smoothing and over-squashing.
Here we perform one training per model, sharing for each model the hyperparameters in Table \ref{tab:hps_sensitivity_dirichlet}. GCN is implemented such that OST is prone to occur, so we did not use residual connections and used ReLU as activation function \citep{residualconnectionsnormalizationprovably, vanishinggradientsoversmoothingoversquashing}. 
\begin{table}[t]
\centering
\caption{Hypeparameters used for the Dirichlet Energy and Sensitivity analyses.}
\label{tab:hps_sensitivity_dirichlet}
\resizebox{\textwidth}{!}{
\begin{tabular}{lcccccccc}
\hline
 & \texttt{Amazon Ratings} & \texttt{Roman empire} & \texttt{Minesweeper} & \texttt{Questions} & \texttt{Tolokers} & \texttt{Peptides-func} & \texttt{Peptides-struct} & \texttt{Pascal-VOC} \\
\hline
$m'$ & 256 & 128 & 32 & 64 & 32 & 300 & 300 & 128 \\
$L$  & 50  & 50  & 50 & 50 & 50 & 10  & 10  & 10 \\
\hline
\end{tabular}}
\end{table}
We report the sensitivity $S$ for \texttt{Pascal-VOC} and \texttt{Peptides-struct} in Figure \ref{fig:appendix_sensitivity}. We notice that, similarly to the example in the main paper, we do not record any difficulties by the models to keep high sensitivity. Also, we notice that in \texttt{Peptides-struct}, we have a similar trend to \texttt{Peptides-func}, which reflects the same dataset domain. In Figure \ref{fig:pascal_sensitivity}, we record a decreasing trend in sensitivity from all the models, but in general, they all have $S_l \neq 0$, also in this case. 
In Figure \ref{fig:appendix_dirichlet}, we analyse the $E^{dir}$ trends for GCN, GRAFF, A-DGN, and SAS-GNN, to understand whether SAS-GNN effectively mitigates over-smoothing. We observe that also in this case, as in the \texttt{Minesweeper} example of Figure \ref{fig:dirichlet_performance}, SAS-GNN follows a similar trend of A-DGN, in all the other datasets, meaning that its ability to mitigate over-smoothing is more inherited from this model, rather than GRAFF. In these other experiments, we notice that $E^{dir}$ for GRAFF grows exponentially. As observed also in \texttt{Minesweeper}, it is clear that these plots describe the over-sharpening phenomenon discussed by \cite{GRAFF2}. For instance, GRAFF is a more expressive version of GCN, which is provably capable of both asymptotically experiencing over-smoothing, as well as the opposite effect, namely over-sharpening. The desired behavior is a trade-off of the two, which is what we observe when GRAFF has a contained number of layers. Here, GRAFF underperforms due to an asymptotic behavior that lets adjacent node features diverge; indeed, $E^{dir}$ increases. As concerns GCN, as expected, it experiences over-smoothing also in the remaining heterophilic datasets. 

\subsection{SAS-GNN is able to retain performance as depth increases}
\label{sec:appendix_depth}
\begin{figure}[t]
    \centering
    
    \begin{subfigure}[b]{0.43\textwidth}
        \centering
        \includegraphics[width=\linewidth]{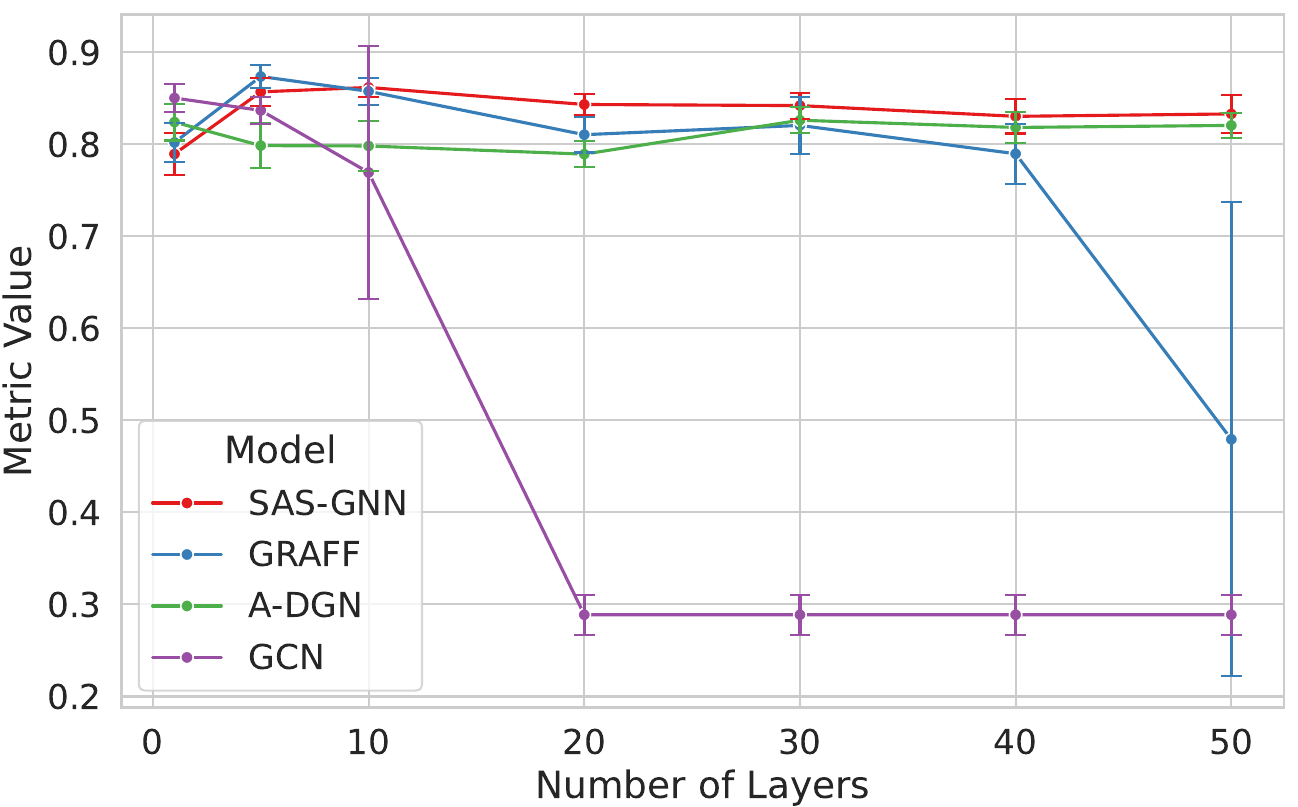}
        \caption{\texttt{Cora}: Comparison of models' performance w.r.t increasing depth, in a homophilic graph.}
        \label{fig:depth_performance}
    \end{subfigure}
    \hfill
    \begin{subfigure}[b]{0.47\textwidth}
        \centering
        \includegraphics[width=\linewidth]{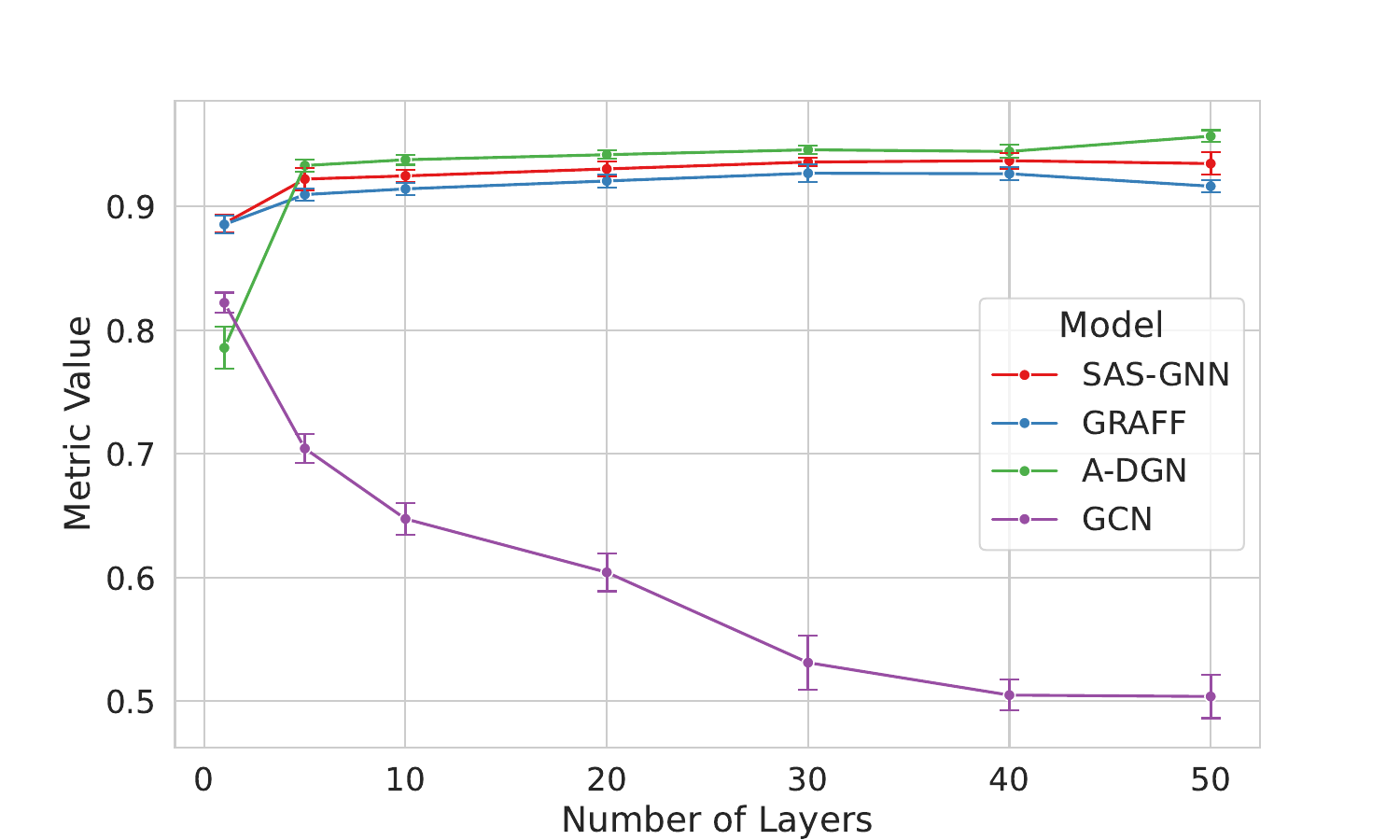}
        \caption{\texttt{Minesweeper}: Comparison of models' performance w.r.t increasing depth under heterophily.}
        \label{fig:mines_performance}
    \end{subfigure}
    \caption{Comparison of models' performance on different datasets with increasing depth.}
    \label{fig:combined_performance}
\end{figure}

Having proposed SAS-GNN, we have demonstrated that it is always stable and non-dissipative, and also able to allow each node to induce attraction or repulsion edge-wise. We study the behavior of SAS-GNN when the depth increases up to 50 layers, which is an overestimate of what we need in practice. In our experiments, we design SAS-GNN, as specified in Corollary \ref{corr:sigma}. All the data points corresponds to a different training procedure, where we display the best accuracy. We present the results in Figure \ref{fig:depth_performance}, where we show an experiment with \texttt{Cora}, a homophilic dataset. In Figure \ref{fig:mines_performance}, we assess the resilience to depth via \texttt{Minesweeper}, an example of a heterophilic graph. 
In \texttt{Cora}, we notice that both A-DGN and SAS-GNN can retain their performance. Also, GRAFF can maintain it up to the 30th layer, then it starts decreasing. This behavior is expected since GRAFF is only designed to contrast over-smoothing, which does not occur, but it is not specifically designed to be robust to the number of layers. It is interesting to notice that SAS-GNN, even though part of its architecture is inherited by GRAFF, can retain its performance, behaving similarly to A-DGN. As far as GCN is concerned, we see that it has started losing its performance in the earlier layers. Here, we consider a GCN design that lacks the residual connection and does not use weight sharing. As concerns \texttt{Minesweeper}, we notice that GRAFF is the model that maximizes the most $E^{dir}$, while SAS-GNN and A-DGN still have a similar trend. None of these models seems to experience over-smoothing, except for GCN, where $E^{dir}$ approaches 0. Also, in this case, all the models retain their performances, except for GCN, which starts decreasing from the beginning. Since GRAFF is the only model that perceive the performance degradation, and Dirichlet Energy does not record over-smoothing, we conclude that it probably suffer from other phenomena, that do not affect A-DGN and SAS-GNN. 
\begin{table}[t]
\centering
\caption{Performance across different model sizes and layer depths}
\label{tab:100_scaling_layer}
\begin{tabular}{c c c c c}
\hline
\textbf{Model} & \textbf{10 layers} & \textbf{50 layers} & \textbf{100 layers} & \textbf{\# of Parameters} \\
\hline
$m' = 32$ & 92.95 $\pm$ 0.5 & 93.05 $\pm$ 0.3 & 93.64 $\pm$ 0.2 & 2,496 \\
$m' = 64$ & 91.95 $\pm$ 0.1 & 92.54 $\pm$ 0.1 & 92.85 $\pm$ 0.3 & 9,088 \\
$m' = 128$ & 91.20 $\pm$ 0.2 & 92.87 $\pm$ 0.1 & 93.24 $\pm$ 0.1 & 34,560 \\
$m' = 256$ & 90.73 $\pm$ 0.1 & 92.67 $\pm$ 0.2 & 92.90 $\pm$ 0.3 & 134,656 \\
\hline
\end{tabular}

\end{table}

We also provide further experiments showing that SAS-GNN preserves its predictive power even when the number of layers increases. This property is crucial for two reasons. First, deeper models enable richer feature extraction, which can improve generalization, something that standard MPNNs often fail to achieve due to their limitations. Second, in the specific case of MPNNs, increased depth allows for long-range information propagation, which shallower networks cannot capture, but could be critical in some settings.

To assess these two aspects, we examine two different scenarios:

1. \textbf{Highly heterophilic graphs}.  
   In this setting, the challenge lies not in long-range propagation but in the ability to correctly separate adjacent nodes that frequently belong to different classes. We focus on the WebKB datasets introduced by \citet{webkb2}, namely \texttt{Texas}, \texttt{Wisconsin}, and \texttt{Cornell}. Their statistics are reported in Table~\ref{tab:datasets_small_heterophilic}.

   \begin{table}[ht]
       \centering
       \caption{Highly-Heterophilic Graphs.}
       \label{tab:datasets_small_heterophilic}
       \scalebox{0.95}{
       \begin{tabular}{lcccccc}
           \hline
           \textbf{Datasets} & $N$ & $|\mathcal{E}|$ & $d$ & $|C|$ & $\xi_{edge}$ & $\xi_{adj}$ \\
           \hline
           \textbf{Texas}             & 183 & 574 & 1703& 5& 0.09& -0.26\\
           \textbf{Wisconsin}         & 251& 916& 1703& 5& 0.20& -0.15\\
           \textbf{Cornell}           & 183& 557& 1703& 5& 0.13& -0.21\\ \hline
       \end{tabular}}
   \end{table}

Although these datasets are no longer widely used for benchmarking \citep{criticallookatgnn}, since models typically exhibit high variance across folds, they are valuable examples of extreme heterophily. Both $\xi_{edge}$ and $\xi_{adj}$ take very low values, meaning that neighbors share different classes very frequently, even regardless of class distribution (i.e. $\xi_{edge}$ does not consider number of classes of class balance). Compared to the benchmarks introduced by \citet{criticallookatgnn}, the WebKB graphs display significantly stronger heterophily.  

\begin{figure}[ht]
    \centering
    \begin{minipage}{0.48\linewidth}
        \centering
        \includegraphics[width=\linewidth]{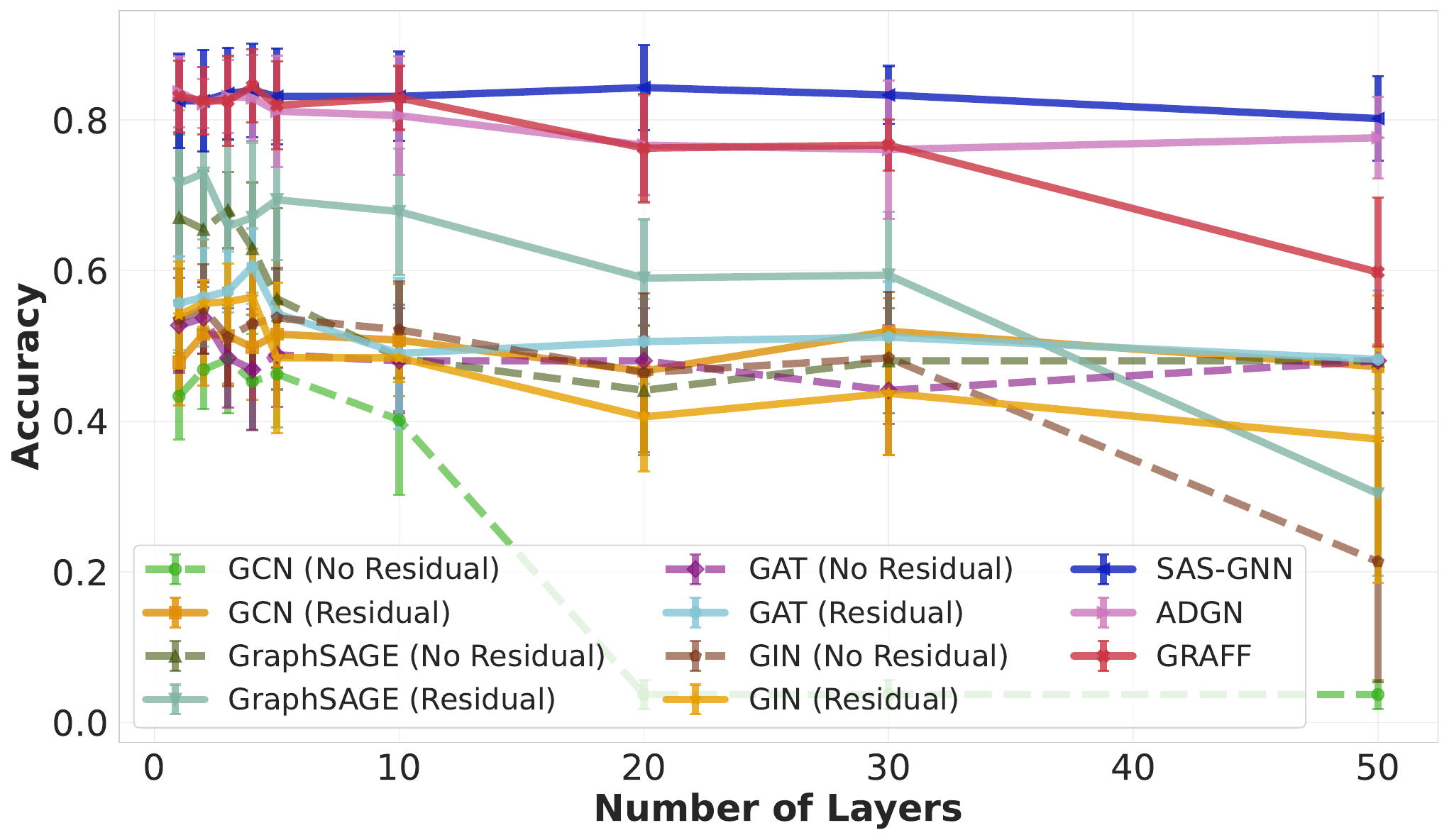}
        \subcaption{\texttt{Wisconsin}}
    \end{minipage}
    \hfill
    \begin{minipage}{0.48\linewidth}
        \centering
        \includegraphics[width=\linewidth]{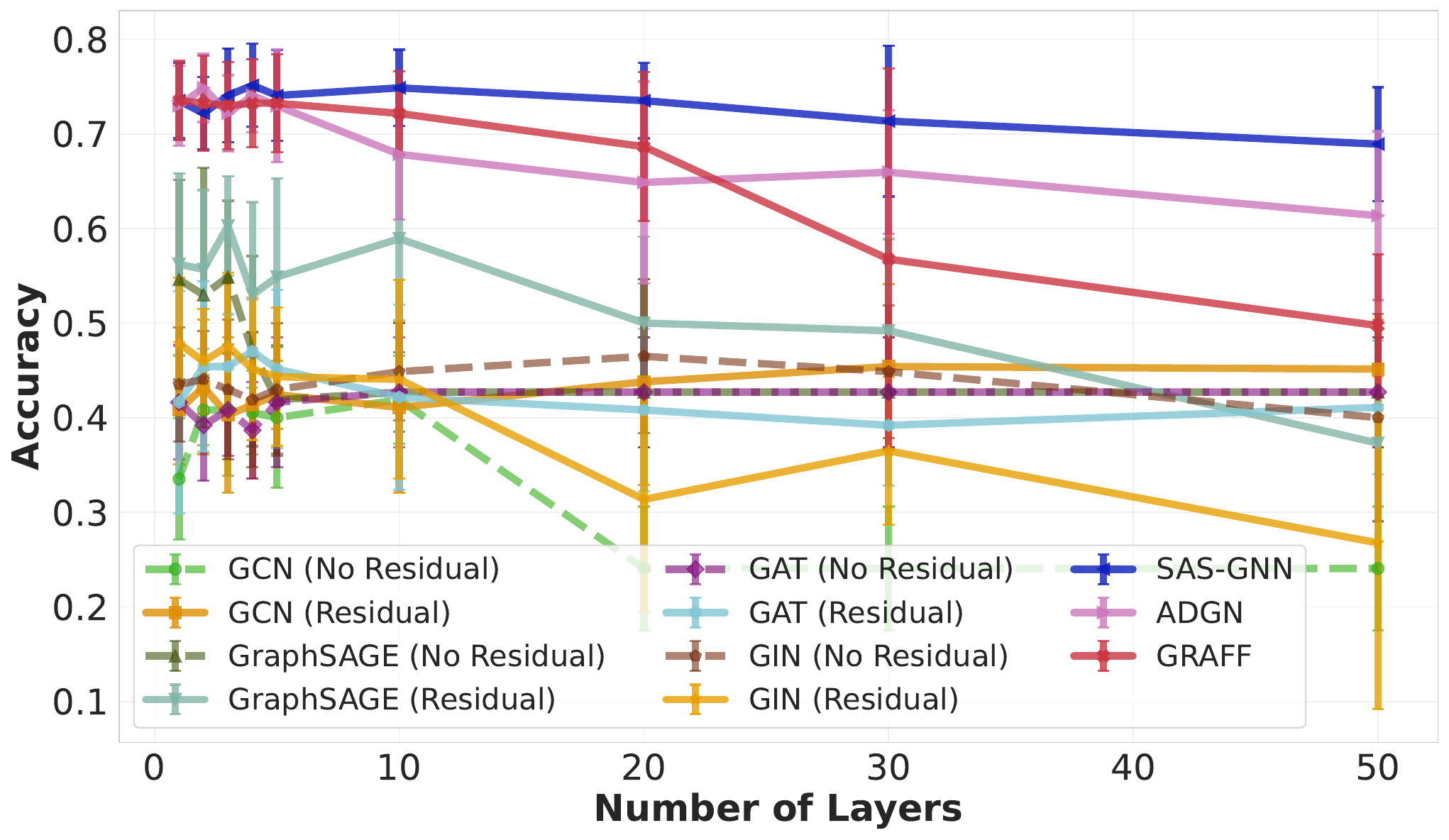}
        \subcaption{\texttt{Cornell}}
    \end{minipage}
    \caption{Accuracy comparison of all baselines on \texttt{Wisconsin} and \texttt{Cornell}.}
    \label{fig:wisconsin_and_cornell}
\end{figure}

For this scenario, depth is not used to capture long-range dependencies but rather as a stress test: we want to see whether the model can maintain performance across layers while still separating adjacent nodes correctly. SAS-GNN is a hybrid of GRAFF, which addresses over-smoothing and heterophily, and A-DGN, which addresses over-squashing and long-range propagation. Thus, it is particularly interesting to evaluate its behavior here. In the main paper, we already showed that SAS-GNN is actually effective in this scenario by maintaining its performance on \texttt{Texas} even as depth increases (see Figure~\ref{fig:depth_performance_degrade}), whereas classic MPNNs, even with residual connections, which safeguard from over-smoothing \citep{residualconnectionsnormalizationprovably}, still suffer performance drops at larger depths. In Figure \ref{fig:wisconsin_and_cornell}, we report additional results on \texttt{Cornell} and \texttt{Wisconsin}, confirming the same trend. The common aspects among these comparison are that GRAFF is not able to retain its performance across layers, which is expected, since it was not designed for deep configurations. Specifically, in the original paper of GRAFF, the best configurations were typically using very shallow configurations (i.e. 2-3 layers \citep{GRAFF2}).  The other aspect is that SAS-GNN is always retaining accuracy, showing also that regardless of the hostility of the class distribution, it still preserve information, similarly to A-DGN. A-DGN seems to lag behind SAS-GNN in these scenarios, probably because A-DGN was not designed to deal with high heterophily. Nonetheless, we did not tune extensively on of these models, these experiments are mostly a proof of concept that illustrates the capabilities of our model and how its inductive biases reflect on real-world graphs. The details to reproduce these experiments are in the codebase.

2. \textbf{Synthetic datasets for long-range propagation.}  
In contrast, our second scenario is designed to explicitly test whether depth enables the model to exploit long-range information. For this purpose, we use synthetic graphs where the predictive task depends on aggregating information across distant nodes. In this setting, the ability of SAS-GNN to propagate signals over many layers is the main factor under evaluation, rather than heterophily and retaining performance after many layers. 

Following the experimental setup of \citet{swan}, we consider the \emph{Graph Transfer} task. This resilience of a model to over-squashing when transmitting information across multiple hops.

\paragraph{Graph Transfer.}  
The Graph Transfer task focuses directly on the problem of topological and computational over-squashing. The setup consists of a source–target pair of nodes separated by $k$ hops ($k \in \{3, 5, 10, 50\}$). The goal is to transfer a binary label from the source to the target across different graph topologies (line, ring, and crossed-ring). A better definition of the task is described by \citet{swan}. This task can only be solved by preserving source information over long distances, where topology may be responsible for compressing excessively such information.  

\begin{figure}[ht]
    \centering
    \includegraphics[width=\linewidth]{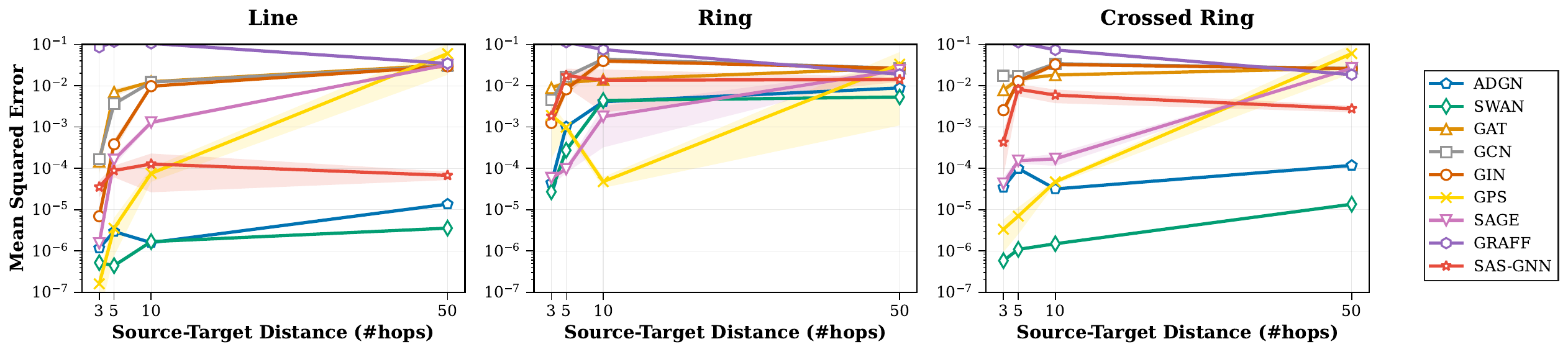}
    \caption{Performance on the Graph Transfer task for line, ring, and crossed-ring graphs at increasing distances. SAS-GNN remains stable across depths, though it lags behind A-DGN.}
    \label{fig:graph_transfer}
\end{figure}

In Figure~\ref{fig:graph_transfer}, we report the results of our experiments. SAS-GNN maintains stable performance across depths, confirming that it inherits the inductive bias of A-DGN, which was explicitly designed for long-range propagation. However, SAS-GNN performs slightly below A-DGN and SWAN in this benchmark. Notably, GRAFF alone performs poorly in this setting, and this suggests that the symmetric component inherited from it does not provide benefits here and likely reduces predictive accuracy. This likely explains why SAS-GNN lags behind the other two models despite retaining stability across layers.

\paragraph{Summary.}  
Taken together, these results show that SAS-GNN is effective under multiple challenging conditions: extreme heterophily, very deep architectures, and long-range propagation. By combining the strengths of GRAFF (robustness to heterophily) and A-DGN (resilience to over-squashing and long-range propagation), SAS-GNN emerges as a reliable architecture for both deep processing and information preservation. This makes it a strong candidate not only for tackling over-smoothing and over-squashing simultaneously, but also for practical applications such as the design of early-exit strategies.

\subsection{Analysis on the use of ReLU+TanH as activation}
\label{sec:extended_nonlinearities}
Here, we want to understand how our instantiation of the non-linearities $\sigma_1(x) = \text{ReLU}(\text{TanH}(x))$ and $\sigma_2 = \text{ReLU}(x)$ results in effective w.r.t. more common choices in the design of neural networks, such as ReLU or TanH. In order to assess these aspects, we compare SAS-GNN against A-DGN and GRAFF, using all the possible combinations of ReLU, TanH, and ReLU+TanH. Our objective is to understand whether we have any advantage in terms of performance, rather than a result that is limited to the theory. We present such a comparison in Table \ref{tab:activations_comparison}. From the Table, we see that our principled approach ranks first in 2 out of 5 datasets, namely \texttt{Tolokers} and \texttt{Questions}. While A-DGN performs significantly better than GRAFF and SAS-GNN in \texttt{Minesweeper} and \texttt{Roman Empire}. In the other datasets, the performance remains comparable for all the activation types. Surprisingly, GRAFF tends to perform worse than the other models, even though it was designed specifically to handle heterophilic graphs. However, when it was proposed by \cite{GRAFF2}, this specific benchmark was not introduced yet, thus, we attribute this gap to this new collection. We present additional evidence on the impact of ReLU+TanH, when used within other GNNs, such as GCN, GraphSAGE, GAT, and GIN, in Section \ref{sec:extended_EE}.
\begin{table}[t]
    \centering
    \caption{Comparison of the activation types based on the models. The scores are marked in red for the \textcolor{red}{first}, blue for the \textcolor{blue}{second}. These results are taken from \citet{criticallookatgnn, coognns, classicgnnsstrongbaselines}.}
    \label{tab:activations_comparison}
    \resizebox{0.85\textwidth}{!}{
    \begin{tabular}{c c c c c c c}
        \hline
        \textbf{Model} & \textbf{Activation} & \texttt{Amazon Ratings} & \texttt{Minesweeper} & \texttt{Roman Empire} & \texttt{Tolokers} & \texttt{Questions} \\
        \hline
        \multirow{3}{*}{A-DGN} 
        & TanH   & 49.14 $\pm$ 0.7  & 94.81 $\pm$ 0.4 & 82.53 $\pm$ 0.7  & 84.90 $\pm$ 1.1  & 76.51 $\pm$ 1.0 \\
        & ReLU   & 51.41 $\pm$ 0.4  & \textcolor{red}{95.08 $\pm$ 0.5} & \textcolor{red}{84.83 $\pm$ 0.6}  & 85.32 $\pm$ 1.0 & 78.24 $\pm$ 1.8\\
        & ReLU+TanH& \textcolor{red}{51.57 $\pm$ 1.3}  & \textcolor{blue}{94.86 $\pm$ 0.4} & \textcolor{blue}{84.66 $\pm$ 0.6}  & 85.35 $\pm$ 1.7  & 79.01 $\pm$ 1.1 \\
        \hline
        \multirow{3}{*}{GRAFF}
        & TanH   & 50.02 $\pm$ 1.5  & 92.44 $\pm$ 0.7 & 78.76 $\pm$ 0.8  & 85.03 $\pm$ 1.8  & 76.77 $\pm$ 1.5 \\
        & ReLU   & 51.41 $\pm$ 0.7  & 93.23 $\pm$ 0.5 & 78.98 $\pm$ 0.4  & 85.67 $\pm$ 0.7  & 76.27 $\pm$ 1.2 \\
        & ReLU+TanH& 50.77 $\pm$ 0.6  & 92.59 $\pm$ 0.7 & 78.59 $\pm$ 0.5  & 85.29 $\pm$ 0.9  & 79.04 $\pm$ 1.1 \\
        \hline
        \multirow{3}{*}{SAS-GNN}
        & TanH   & 48.91 $\pm$ 0.8  & 93.80 $\pm$ 0.5 & 82.07 $\pm$ 0.5  & 84.56 $\pm$ 0.7  & 77.93 $\pm$ 1.3 \\
        & ReLU   & 51.39 $\pm$ 0.6  & 93.38 $\pm$ 0.5 & 83.59 $\pm$ 0.5  & \textcolor{blue}{85.72 $\pm$ 1.8}  & \textcolor{blue}{79.13 $\pm$ 0.4}\\
        & ReLU+TanH& \textcolor{blue}{51.47 $\pm$ 0.7}  & 93.29 $\pm$ 0.6 & 83.46 ± 0.6  & \textcolor{red}{85.80 $\pm$ 0.8}  & \textcolor{red}{79.60 $\pm$ 1.1} \\
        \hline
    \end{tabular}}
\end{table}

\subsection{Full Results on the Benchmark for Heterophily}
\label{sec:extended_heterophily}

In this section, we present the full leaderboard results from the heterophily benchmark introduced by~\citet{criticallookatgnn}. While Table~\ref{tab:result_heterophilic} already provided a fair comparison with our approach, here we aim to give a more complete picture. 

The benchmark of~\citet{criticallookatgnn} evaluates GNNs on diverse medium-to-large graphs spanning different domains and exhibiting varying degrees of heterophily, density, and size. Their study revealed that many models explicitly designed for heterophily often underperform compared to classical GNNs such as GCN, GraphSAGE, GAT, or even GTs when tested under this more realistic setting. More recently, \citet{classicgnnsstrongbaselines} showed that the performance of these classic MPNNs can be pushed even further by incorporating simple architectural components, most notably, dropout layers interleaved by message-passing layers. Table~\ref{tab:extended_result_heterophilic} reports the updated benchmark. Results are taken by \citet{criticallookatgnn, coognns, classicgnnsstrongbaselines}.

From this broader context, two main observations emerge with respect to our approach. First, our models consistently outperform heterophily-specialized architectures, even without extensive tuning, demonstrating that our design generalizes well to heterophilic settings. Second, performance gains in these benchmarks are strongly correlated with components that are not currently supported by our theoretical framework, such as dropout or normalization layers. Indeed, ablating dropout leads to significant performance drops across baselines. In some cases (e.g., \texttt{Minesweeper} and \texttt{Roman Empire}) models without dropout still surpass ours, while in others (e.g., \texttt{Questions} and \texttt{Tolokers}) dropout is less critical and our models perform comparably or even better. These cases highlight situations where our principled design is particularly effective.

At the same time, these results underscore an important limitation of principled architectures: in practical scenarios, accuracy improvements often require the inclusion of black-box components that fall outside current theoretical guarantees. This gap between theoretical soundness and empirical performance remains a key challenge for future work. We included in the table also some non-principled version of SAS-GNN, with no hyperparameter tuning, where we remove weight-sharing, or include dropout, or include normalization, and all the possible combination of these components. Here we see, that with no hyperparameter tuning the test performance already improves significantly. An example of this is \texttt{Roman Empire} ($83.46 \rightarrow 87.32$). We believe that our model could benefit from an extensive hyperparameter tuning, however this questions its capabilities for safe early exits.

\textbf{Neural Adaptive Step.} 
We further analyze the role of our \textit{Neural Adaptive Step} mechanism in node classification, an aspect not covered in the main paper. Specifically, we compare the default formulation, where each node is assigned its own adaptive step vector, against a simplified variant where the step is reduced to a scalar shared across all nodes. To ensure differentiability, the scalar is defined as the mean of the vector entries. Concretely, in Algorithm~\ref{alg:eegnn}, the update rule becomes:
\[
\mathbf{H}^{l+1} \gets \mathbf{H}^l + \mathrm{mean}(\bm{\tau}^l)\, \Delta \mathbf{H}^l,
\]
so that every node is updated with the same constant factor, namely the average prediction of the confidence network.

This experiment highlights the impact of personalized integration constants. While the scalar variant enforces a uniform update across all nodes, the full neural adaptive step allows each node to adapt its integration constant based on its own confidence estimate. As shown in Table~\ref{tab:extended_result_heterophilic}, moving from the vector to the scalar consistently degrades performance across datasets, with particularly pronounced drops on \texttt{Questions}. We attribute this to the fact that averaging removes much of the information conveyed by the confidence network, weakening its guidance to the message-passing process.

We do not extend this analysis to graph-level tasks, as in those cases, we already use a scalar integration constant by design, since there the confidence prediction already operates at the graph level. Nevertheless, these results provide strong evidence that node-wise adaptive steps are essential for capturing heterogeneity in node-level tasks, and that personalized integration constants substantially contribute to the effectiveness of our approach.

\begin{table*}[t]
    \centering
    \caption{Node classification under heterophily. The scores are marked in red for the \textcolor{red}{first}, blue for the \textcolor{blue}{second}, and green for the \textcolor{MyDarkGreen}{third}. The models marked with the asterisks are those presented in Table \ref{tab:result_heterophilic}, reported by \citet{classicgnnsstrongbaselines}. Results are taken by \citet{criticallookatgnn, polynormer, coognns, classicgnnsstrongbaselines}.}
    \label{tab:extended_result_heterophilic}
    \resizebox{\textwidth}{!}{
        \begin{tabular}{l c c c c c }
        \hline
        \textbf{Model} & \texttt{Amazon Ratings} & \texttt{Minesweeper} & \texttt{Roman Empire} & \texttt{Tolokers} & \texttt{Questions} \\
        \hline
        \textbf{Classic GNNs} \cite{criticallookatgnn} & & & & & \\ \hline
        GCN & 48.70 ± 0.63 & 89.75 ± 0.52 & 73.69 ± 0.74 & 83.64 ± 0.67 & 76.09 ± 1.27 \\
        SAGE & 53.63 ± 0.39 & 93.51 ± 0.57 & 85.74 ± 0.67 & 82.43 ± 0.44 & 76.44 ± 0.62 \\
        GAT & 49.09 ± 0.63 & 92.01 ± 0.68 & 80.87 ± 0.30 & 83.70 ± 0.47 & 77.43 ± 1.20 \\
        GAT-sep & 52.70 ± 0.62 & 93.91 ± 0.35 & 88.75 ± 0.41 & 83.78 ± 0.43 & 76.79 ± 0.71 \\
        GT & 51.17 ± 0.66 & 91.85 ± 0.76 & 86.51 ± 0.73 & 83.23 ± 0.64 & 77.95 ± 0.68 \\
        GT-sep & 52.18 ± 0.80 & 92.29 ± 0.47 & 87.32 ± 0.39 & 82.52 ± 0.92 & 78.05 ± 0.93 \\
        \hline
        \textbf{Heterophily-Specialized Models} \cite{criticallookatgnn} & & & & & \\ \hline
            H$_2$GCN & 36.47 ± 0.23 & 89.71 ± 0.31 & 60.11 ± 0.52 & 73.35 ± 1.01 & 63.59 ± 1.46 \\
            CPGNN & 39.79 ± 0.77 & 52.03 ± 5.46 & 63.96 ± 0.62 & 73.36 ± 1.01 & 65.96 ± 1.95 \\
            GPR-GNN & 44.88 ± 0.34 & 86.24 ± 0.61 & 64.85 ± 0.27 & 72.94 ± 0.97 & 55.48 ± 0.91 \\
            FSGNN & 52.74 ± 0.83 & 90.08 ± 0.70 & 79.92 ± 0.56 & 82.76 ± 0.61 & 78.86 ± 0.92 \\
            GloGNN & 36.89 ± 0.14 & 51.08 ± 1.23 & 59.63 ± 0.69 & 73.39 ± 1.17 & 65.74 ± 1.19 \\
            FAGCN & 44.12 ± 0.30 & 88.17 ± 0.73 & 65.22 ± 0.56 & 77.75 ± 1.05 & 77.24 ± 1.26 \\
            GBK-GNN & 45.98 ± 0.71 & 90.85 ± 0.58 & 74.57 ± 0.47 & 81.01 ± 0.67 & 74.47 ± 0.86 \\
            JacobiConv & 43.55 ± 0.48 & 89.66 ± 0.40 & 71.14 ± 0.42 & 68.66 ± 0.65 & 73.88 ± 1.16 \\
        \midrule
        \textbf{Reassessment + Current SOTA} \cite{classicgnnsstrongbaselines} & & & & & \\ \hline
        \midrule
        GCN* & 53.80 ± 0.60 & \textbf{\textcolor{red}{97.86 ± 0.52}} & \textbf{\textcolor{MyDarkGreen}{91.27 ± 0.20}} & 83.64 ± 0.67 & \textbf{\textcolor{blue}{79.02 ± 1.27}} \\
        \hspace{0.5em} -- Dropout & 51.37 ± 0.34 & 94.28 ± 2.29 & 85.10 ± 0.61 & - & 76.58 ± 0.48 \\ 
        SAGE* & \textbf{\textcolor{blue}{55.40 ± 0.21}} & \textbf{\textcolor{blue}{97.77 ± 0.62}} & 91.06 ± 0.67 & 82.43 ± 0.44 & 77.21 ± 1.28 \\
        \hspace{0.5em} -- Dropout & 51.12 ± 0.66 & 93.83 ± 0.38 & 84.49 ± 0.35 & - & 76.36 ± 1.58 \\
        GAT* &  \textbf{\textcolor{red}{55.54 ± 0.51}} & \textbf{\textcolor{MyDarkGreen}{97.73 ± 0.73}} &  90.63 ± 0.14 & 83.78 ± 0.43 & 77.95 ± 0.51 \\
        \hspace{0.5em} -- Dropout & 51.48 ± 0.28 & 92.26 ± 4.63 & 82.47 ± 0.70 & - & 76.19 ± 0.88 \\

        GT & 51.17 ± 0.66 & 91.85 ± 0.76 & 86.51 ± 0.73 & 83.23 ± 0.64 & 77.95 ± 0.68 \\
        NodeFormer* & 43.79 ± 0.57 & 87.71 ± 0.69 & 74.83 ± 0.81&  78.10 ± 1.03 & 75.02 ± 1.61 \\
        SGFormer & 54.14 ± 0.62 & 91.42 ± 0.41 & 80.01 ± 0.44 & -& 73.81 ± 0.59\\
        Polynormer* & \textbf{\textcolor{MyDarkGreen}{54.81 ± 0.49}} & 97.46 ± 0.36 & \textbf{\textcolor{red}{92.55 ± 0.37}}& \textbf{\textcolor{red}{85.91 ± 0.74}} & \textbf{\textcolor{MyDarkGreen}{78.92 ± 0.89}} \\ \hline
        CO-GNN \citep{coognns} & 54.17 ± 0.37 & 97.31 ± 0.41 & \textbf{\textcolor{blue}{91.37 ± 0.35}} & 84.45 ± 1.17 & 76.54 ± 0.95 \\
        \hline
        \multicolumn{6}{l}{\textbf{Ours}} \\ 
        SAS-GNN & $51.47 \pm 0.68$ & $93.29 \pm 0.61$ & $83.46 \pm 0.61$ & \textbf{\textcolor{blue}{$85.80 \pm 0.79$}} & \textbf{\textcolor{red}{$79.60 \pm 1.15$}} \\ \hline
        SAS-GNN$_{-ws}$ & $51.51 \pm 0.41$ & $92.62 \pm 0.45$ & $84.45 \pm 0.44$ & $85.46 \pm 0.63$ & $78.79 \pm 0.97$ \\
        SAS-GNN$_{+drop+norm}$ & $51.19 \pm 0.53$ & \textbf{$93.40 \pm 0.48$} & $86.00 \pm 0.53$ & \textbf{$85.98 \pm 0.55$} & $76.64 \pm 1.40$ \\
        SAS-GNN$_{+drop+norm-ws}$ & \textbf{$52.16 \pm 0.50$} & $93.21 \pm 0.40$ & \textbf{$87.32 \pm 0.00$} & $85.87 \pm 0.70$ & $77.57 \pm 1.10$ \\ \hline
        EEGNN  & $51.47 \pm 0.51$ & $93.18 \pm 1.37$ & $80.36 \pm 0.43$ & \textbf{\textcolor{teal}{$85.26 \pm 0.65$}} & $78.90 \pm 1.15$\\ 
        EEGNN$_{\text{wo/NeuralAdaStep}}$  & $50.73 \pm 0.60$ & $91.92 \pm 1.20$ & $78.63 \pm 1.40$ & $84.89 \pm 0.70$ & $72.33 \pm 1.15$\\
        \bottomrule
    \end{tabular}}
\end{table*}
\subsection{Full Results on the Long Range Graph Benchmark}
\label{sec:extended_LRGB}

In the main paper, we compared SAS-GNN and EEGNN against classical MPNNs, Graph Transformers (GTs), asynchronous message-passing models such as Co-GNN, AMP, MTGCN \citep{multitrackgcn}, and graph neural ODEs, which are the class of models most closely related to ours. Among these we considered a diffusion-inspired message-passing, namely GRAND \cite{GRAND}, a physics-inspired scheme GraphCON \cite{graphcon}, and two examples of stable and non-dissipative methods, namely A-DGN and SWAN \citep{swan}. SWAN is an example of global as well as local stable and non-dissipative message-passing, making it a better alternative w.r.t. A-DGN. Our models, SAS-GNN and EEGNN are limited by local non-dissipativeness.

Here, we provide a broader view of the Long Range Graph Benchmark (LRGB).  

As discussed in the main text, peptide datasets are no longer considered representative of long-range dependencies~\citep{measuringlongrange}, a point we also validated in Figure~\ref{fig:funcvsstruct}. Beyond this, we also include the revised results of~\citet{classicgnnsbaselinesgraphlevel}, who revisited inductive tasks and introduced GCN$^+$, a variant that view classical GCN layers as a combination of message-passing with dropout, normalization, residual connections, and feedforward layers, inspired by Transformer architectures. In practice, dropout and normalization were the most influential components. We further include \textit{rewiring methods} specifically designed to address topological bottlenecks~\citep{demystifyingcommonbeliefs}, such as DIGL~\citep{DIGL}, MixHop~\citep{mixhop}, and DRew~\citep{Drew}. Results are taken from \citet{coognns, swan, classicgnnsbaselinesgraphlevel, MPNN_good_lrgb}, and are summarized in Table~\ref{tab:result_lrgb_extended}.

We report three SAS-GNN variants: SAS-GNN$_{noedge}$ ($f_e(\mathbf{E}) = 0$), SAS-GNN$_{edge}$ ($f_e(\mathbf{E}) = \mathbf{BEW}_e$), and SAS-GNN/EEGNN$_{ours}$ ($f_e(\mathbf{E}) = -\text{ReLU}(\mathbf{BEW}_e)$). While we do not outperform all baselines, we surpass classical MPNNs on \texttt{Peptides-func} and \texttt{Peptides-struct} with SAS-GNN$_{noedge}$. Interestingly, edge features offer little benefit on peptides, though our edge-based formulation still slightly improves over SAS-GNN$_{edge}$. On \texttt{Pascal-VOC}, however, classic MPNNs remain stronger overall, but edge features boost our F1 scores, showing their usefulness in this domain.  

Against rewiring methods, our models remain competitive: in \texttt{Peptides-func}, we outperform most approaches except DRew-GCN and its LapPE variant; in \texttt{Peptides-struct}, SAS-GNN$_{edge}$ underperforms, but the other variants remain strong. While DRew-GCN excels on peptide datasets, we surpass it on \texttt{Pascal-VOC}. Compared to DIGL, our methods are stronger on peptides but weaker on \texttt{Pascal-VOC}.  

When comparing to GTs, we generally perform better in peptide tasks, except for inductive node classification. Relative to asynchronous MPNNs, we fall slightly behind Co-GNN on \texttt{Peptides-func} and remain comparable to AMP on \texttt{Peptides-struct}, despite their higher expressivity beyond 1-WL. Results for \texttt{Pascal-VOC} are not available in this category.  

Finally, when comparing with GCN$^+$, we observe that its performance boost in peptides primarily comes from dropout, whose effect is much smaller in our framework. Yet, as already noted, peptides are not long-range benchmarks, and no class of models, including Transformers, rewiring methods, or asynchronous approaches—shows a consistent advantage there. In this setting, black-box components appear to dominate. By contrast, \texttt{Pascal-VOC} constitutes a true long-range task, as evidenced by the performance gap between Transformers and MPNNs and validated in~\citet{measuringlongrange}. Interestingly, even methods explicitly designed for long-range reasoning, such as A-DGN and SWAN, fall behind on this dataset, even trailing GCN$^+$. Since ablations reveal that GCN$^+$ benefits primarily from normalization layers, also present in Transformers, we attribute this gap not to an inability to capture long-range dependencies, but rather to the increased complexity required in message passing.  

Overall, these findings suggest that while Graph Neural ODEs provide transparency and theoretical grounding, they may underperform in practical long-range scenarios where architectural heuristics (e.g., dropout, normalization) play a decisive role.
\begin{table}[t]
    \centering
      \caption{Performance comparison of models across LRGB datasets. The scores are marked in red for the \textcolor{red}{first}, blue for the \textcolor{blue}{second}, and green for the \textcolor{MyDarkGreen}{third}. Results are taken from \citet{coognns, classicgnnsbaselinesgraphlevel, swan, AdaptiveMP, MPNN_good_lrgb}.}
    \label{tab:result_lrgb_extended}
    \resizebox{0.7\textwidth}{!}{
    \begin{tabular}{@{}lccc@{}}
        \hline
        \textbf{Model} & \texttt{Peptides-func} & \texttt{Peptides-struct} & \texttt{Pascal VOC-SP} \\ 
        & \texttt{AP $\uparrow$} & \texttt{MAE $\downarrow$} & \texttt{F1 $\uparrow$}\\
        \hline
        \multicolumn{4}{l}{\textbf{Classic MPNNs \citep{MPNN_good_lrgb}}} \\ 
        GCN & 68.60±0.50 & 0.2460±0.0013 & 20.78±0.31 \\
        GatedGCN & 67.65±0.47 & 0.2477±0.0009 & \textbf{\textcolor{blue}{38.80±0.40}} \\
        \hline
        \multicolumn{4}{l}{\textbf{Reassessment of MPNNs \citep{classicgnnsbaselinesgraphlevel}}} \\ 
        GCN$^+$ & \textbf{\textcolor{red}{72.61 ± 0.67}} & \textbf{\textcolor{red}{0.2421±0.0016}} & 33.57 ± 0.87 \\
        \hspace{0.5em} -- Dropout & 67.48 ± 0.55 & 0.2549 ± 0.0025 & 30.72 ± 0.69 \\
        \hspace{0.5em} -- Norm& 71.07 ± 0.27 & 0.2509 ± 0.0025 & 18.02 ± 1.11 \\
        GatedGCN$^+$ & 70.06 ± 0.33 & \textbf{\textcolor{blue}{0.2431 ± 0.0020}} & \textbf{\textcolor{red}{42.63 ± 0.57}} \\
        
        \hspace{0.5em} -- Dropout & 66.95 ± 1.01 & 0.2508 ± 0.0014 & 33.89 ± 0.66 \\
        
        \hspace{0.5em} -- Norm& 67.33 ± 0.26 & 0.2474±0.0015 & 36.28 ± 0.43 \\ \hline
        
        \multicolumn{4}{l}{\textbf{Rewiring Methods}} \\       
        DIGL+MPNN & 64.69±0.19 & 0.3173±0.0007 & 28.24±0.39 \\
        \hspace{0.5em} +LapPE & 68.30±0.26 & 0.2616±0.0018 & 29.21±0.38 \\
        MixHop-GCN & 65.92±0.36 & 0.2921±0.0023 & 25.06±1.33 \\
        \hspace{0.5em} +LapPE & 68.43±0.49 & 0.2614±0.0023 & 22.18±1.74 \\
        DRew-GCN & 69.96±0.76 & 0.2781±0.0028 & 18.48±1.07 \\
        \hspace{0.5em} +LapPE & \textbf{\textcolor{MyDarkGreen}{71.50±0.44}} & 0.2536±0.0015 & 18.51±0.92 
        \\\hline
        \multicolumn{4}{l}{\textbf{GTs}} \\ 
        GT+LapPE & 63.26±1.26 & 0.2529±0.0016 & 26.94±0.98 \\
        SAN+LapPE & 63.84±1.21 & 0.2683±0.0043 & 32.30±0.39 \\
        GraphGPS+LapPE & 65.35±0.41 & 0.2500±0.0005 & \textbf{\textcolor{MyDarkGreen}{37.48±1.09}} \\
        \hline
        \multicolumn{4}{l}{\textbf{Methods for OSQ and OST}} \\ 
        CO-GNNs & 69.90 ± 0.93 & 0.2503 ± 0.0025 & -\\
        AMP & \textbf{\textcolor{blue}{71.63 ± 0.58}} & \textbf{\textcolor{blue}{0.2431 ± 0.0004}} & -  \\
        MTGCN & 69.36 ± 0.89 &  0.2461 ± 0.0019 & - \\
        \hline 
        \multicolumn{4}{l}{\textbf{Graph Neural ODEs}} \\ 
        GRAND & 57.89 ± 0.62 & 0.3418 ± 0.0015  & 19.18 ± 0.97 \\
        GraphCON & 60.22 ± 0.68 & 0.2778 ± 0.0018  & 21.08 ± 0.91 \\
        A-DGN & 59.75 ± 0.44 & 0.2874 ± 0.0021  & 23.49 ± 0.54 \\
        SWAN & 63.13 ± 0.46 & 0.2571 ± 0.0018  & 27.96 ± 0.48 \\ \hline
        \multicolumn{4}{l}{\textbf{Ours}} \\ 
        SAS-GNN$_{noedge}$ & 69.71 ± 0.62 & \textbf{\textcolor{MyDarkGreen}{0.2449 ± 0.0013}}  & 22.65 ± 0.27\\
        SAS-GNN$_{edge}$ & 69.27 ± 0.58  & 0.2547 ± 0.0163  & 23.31 ± 0.49\\
        SAS-GNN$_{ours}$ & 69.44 ± 0.63 & 0.2528 ± 0.0130 & 25.64 ± 0.63\\
        EEGNN$_{ours}$ & 68.23 ± 0.37& 0.2532 ± 0.0050 & 24.10 ± 0.73 \\
        \hline
    \end{tabular}}
\end{table}

\subsection{Results using Straight-Through Gumbel-Softmax on Classic GNNs}
\label{sec:extended_EE}

In this section, we demonstrate that our early-exit (EE) module, based on the Straight-Through Gumbel-Softmax estimator, is modular and can be integrated with various GNN architectures, including GCN \citep{GCN}, GraphSAGE \citep{GraphSAGE}, GAT \citep{GAT}, and GIN \citep{WL1}. We first show how these models perform using the deep regime $L=20$. Then we perform a sensitivity analysis study, where we include also EEGNN, where we allow $L$ to vary across an interval, to understand how the number of parameters, exit distributions, and performance change. We already provided an example with the performance variation in the main text in Figure \ref{fig:early-exit_unstable}.

\textbf{Deep Regime experiment}. To integrate the EE module, we focus on message-passing schemes with weight sharing, which allows the shared functions $f_c$ and $f_{\nu}$ to operate across layers. We consider a deep regime with $L = 20$, enabling each node to leverage all shared weight matrices and allowing us to study the effect of deep processing on classic MPNNs. Concretely, we replace line~\ref{alg:message_passing_line} in Algorithm~\ref{alg:eegnn} with the message-passing function of the desired GNN. We denote these models as \texttt{EE+GNN-type}, where \texttt{GNN-type} $\in \{\text{GCN, SAGE, GAT, GIN}\}$.

All models use the same hyperparameters as the best-performing EEGNN configuration in Table~\ref{tab:result_heterophilic}, available in our code repository. We do not perform additional tuning, as our goal is to isolate the impact of the Neural Adaptive Step and end-to-end early-exit optimization in deep regimes. Note that these implementations differ from prior benchmarks~\citep{classicgnnsstrongbaselines}, as they are evaluated within our experimental framework.

We report results on the heterophilic graph benchmarks from \citet{criticallookatgnn}, which vary widely in density, size, and context, providing a more representative evaluation not biased by homophily assumptions. Table~\ref{tab:activation_comparison_grouped} shows results on \texttt{Minesweeper}, \texttt{Questions}, and \texttt{Tolokers}, while Table~\ref{tab:activation_comparison_amazon_roman} shows results on \texttt{Amazon Ratings} and \texttt{Roman Empire}. Bold indicates EE variants outperforming their base model, and asterisks mark cases where ReLU+Tanh generally improves over ReLU.

Across the tables, we observe that EE modules generally improve performance, especially with ReLU activations. Base models often struggle in deep regimes (e.g., GAT on \texttt{Minesweeper}, GIN on \texttt{Questions} or \texttt{Roman Empire}), likely due to over-processing in deep layers. In these cases, EE enables early termination, stabilizing performance. SAS-GNN, by contrast, naturally avoids over-smoothing and over-squashing, making it robust to deep configurations.

Some exceptions occur where EE slightly reduces performance, such as GIN on \texttt{Minesweeper} or GraphSAGE on \texttt{Tolokers}. This may result from applying weight sharing to architectures not originally designed for it, reducing expressivity. Generally, we do not expect EE to always increase performance, as the application of Neural Adaptive Step and exiting before is require only to stabilize the performance, it does not add additional information that potential can improve performance.  
We also observe from these experiments that GraphSAGE generally tolerates deep processing well, showing notable improvements with EE. Additionally, ReLU+Tanh activations consistently yield stable performance across both base and EE models, suggesting this combination effectively stabilizes layer-wise feature evolution in practice. However, we still suggest SAS-GNN or the use of more principled design as a backbone to perform early-exiting, since these are supported by theoretical proofs, and remain a safer solution to perform deep processing when needed.

\textbf{Instability of Early-Exit.}  
As discussed in Section~\ref{sec:experimental_setup}, the effectiveness of an early-exit head depends on the stability of node embeddings throughout their feature evolution. But what happens when deeper processing is required? In Figure~\ref{fig:early-exit_unstable}, we illustrate this issue on the \texttt{Questions} dataset: early-exit performance appears sensitive to changes in the budget $L$, which is undesirable.  

\begin{figure}[ht]
    \centering
    \includegraphics[width=0.48\linewidth]{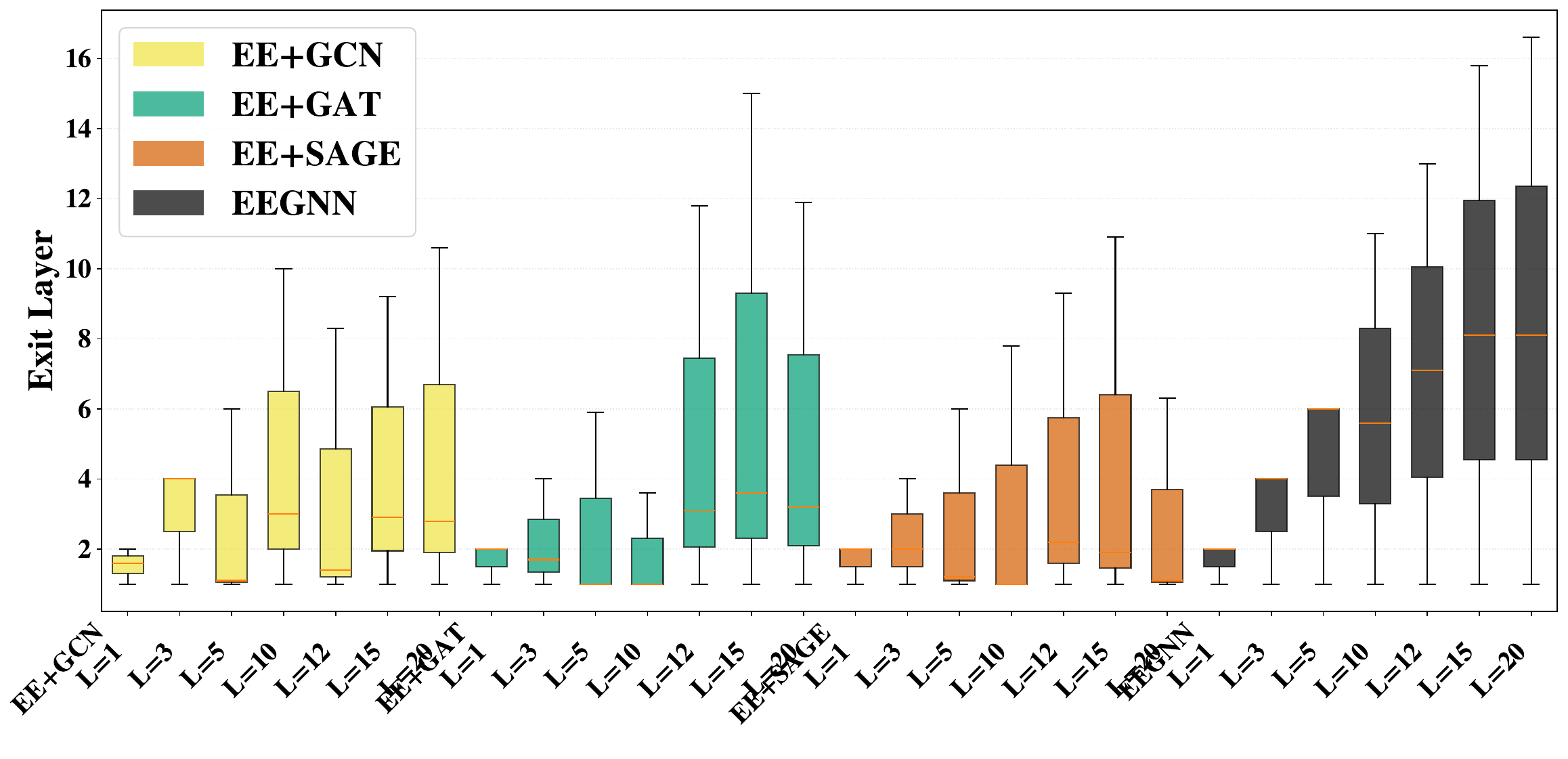}\hfill
    \includegraphics[width=0.48\linewidth]{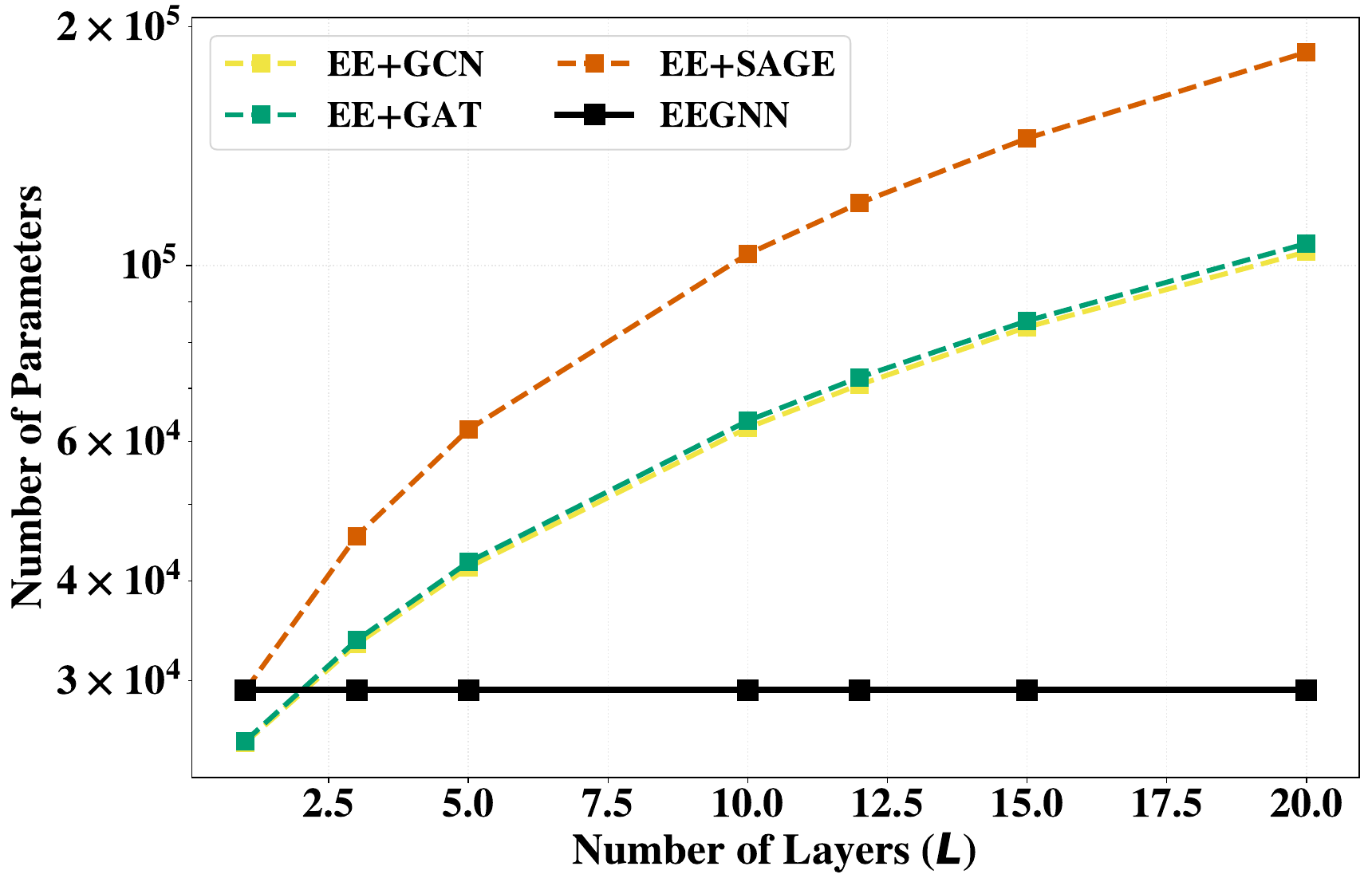}
    \caption{Exit distributions of EEGNN against the others (left) and Parameter evolution based on $L$ (right).}
    \label{fig:questions_exit_distribution_parameters}
\end{figure}

To further investigate, we compare EEGNN against baselines and provide additional evidence of its advantages. Figure~\ref{fig:questions_exit_distribution_parameters} reports both (i) parameter counts (without weight sharing for classic MPNNs, and with for EEGNN) and (ii) exit distributions across different budgets $L$. On the right, we observe that weight sharing keep the number of parameters for EEGNN self-contained. On the left, EEGNN uniquely shifts its exits toward higher depths, indicating that the model attempts to fully utilize its capacity. Although we cannot directly link this behavior to problem-radius prediction, since even with knowledge of the radius, it remains unclear whether sufficient feature extraction occurs within that many layers, we can conclude that EEGNN manage also to increase its effective depth on average when adopting a training with higher budget. Consistent with this, its performance improves at $L=20$, confirming the benefit of deeper processing.  

\begin{figure}[ht]
    \centering
    \includegraphics[width=0.4\linewidth]{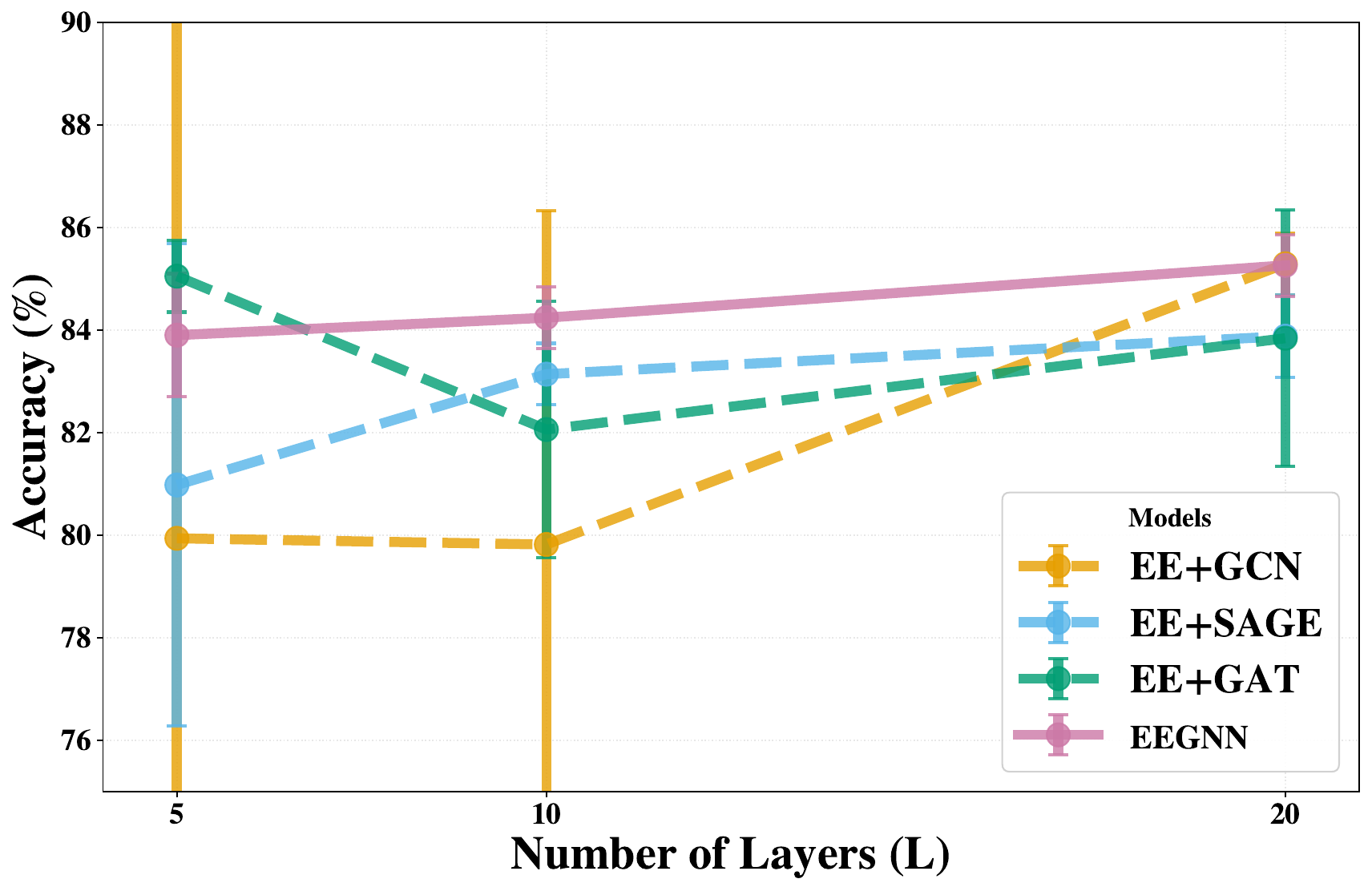}\hfill
    \includegraphics[width=0.55\linewidth]{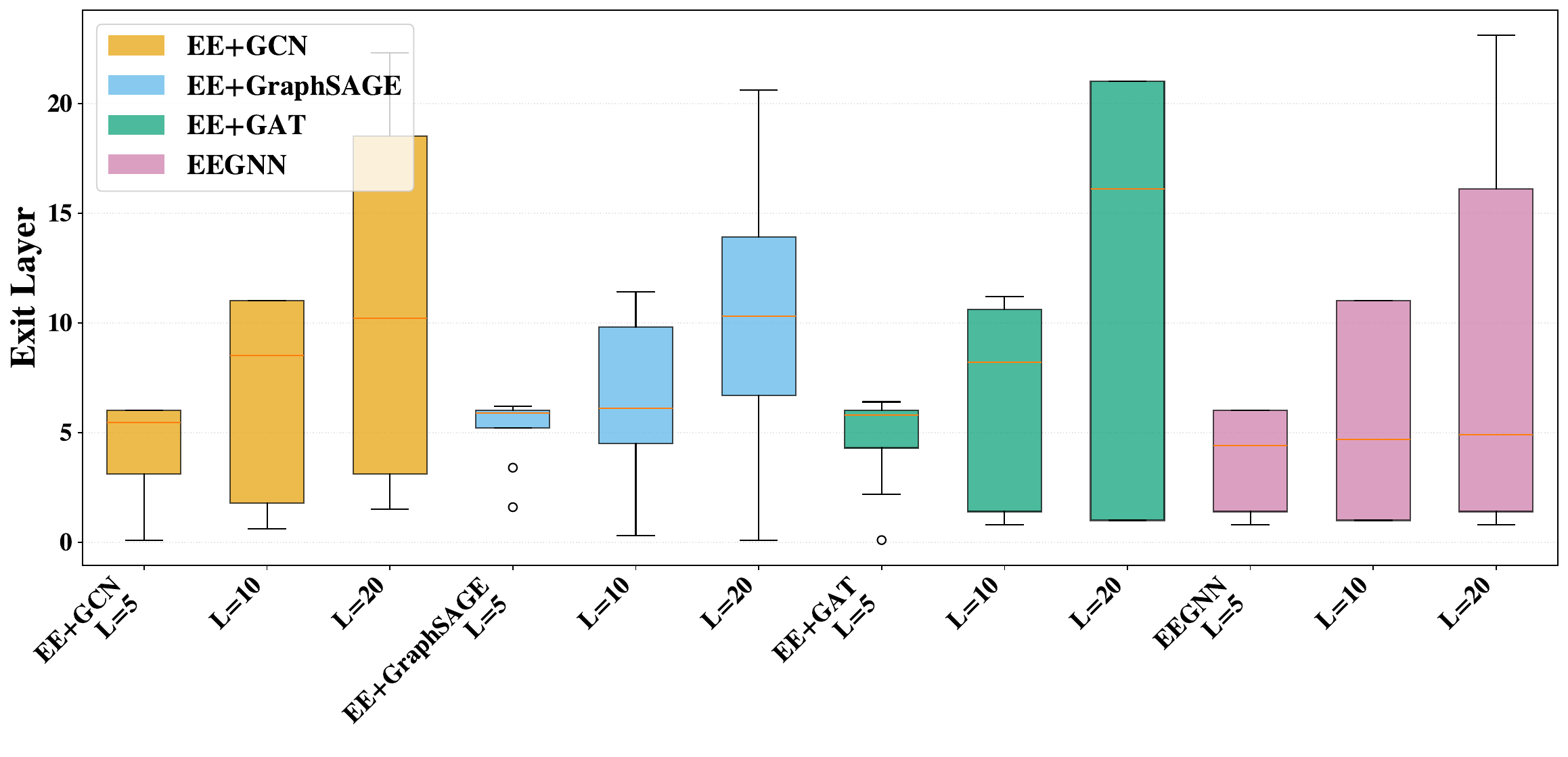}\hfill
    \includegraphics[width=0.4\linewidth]{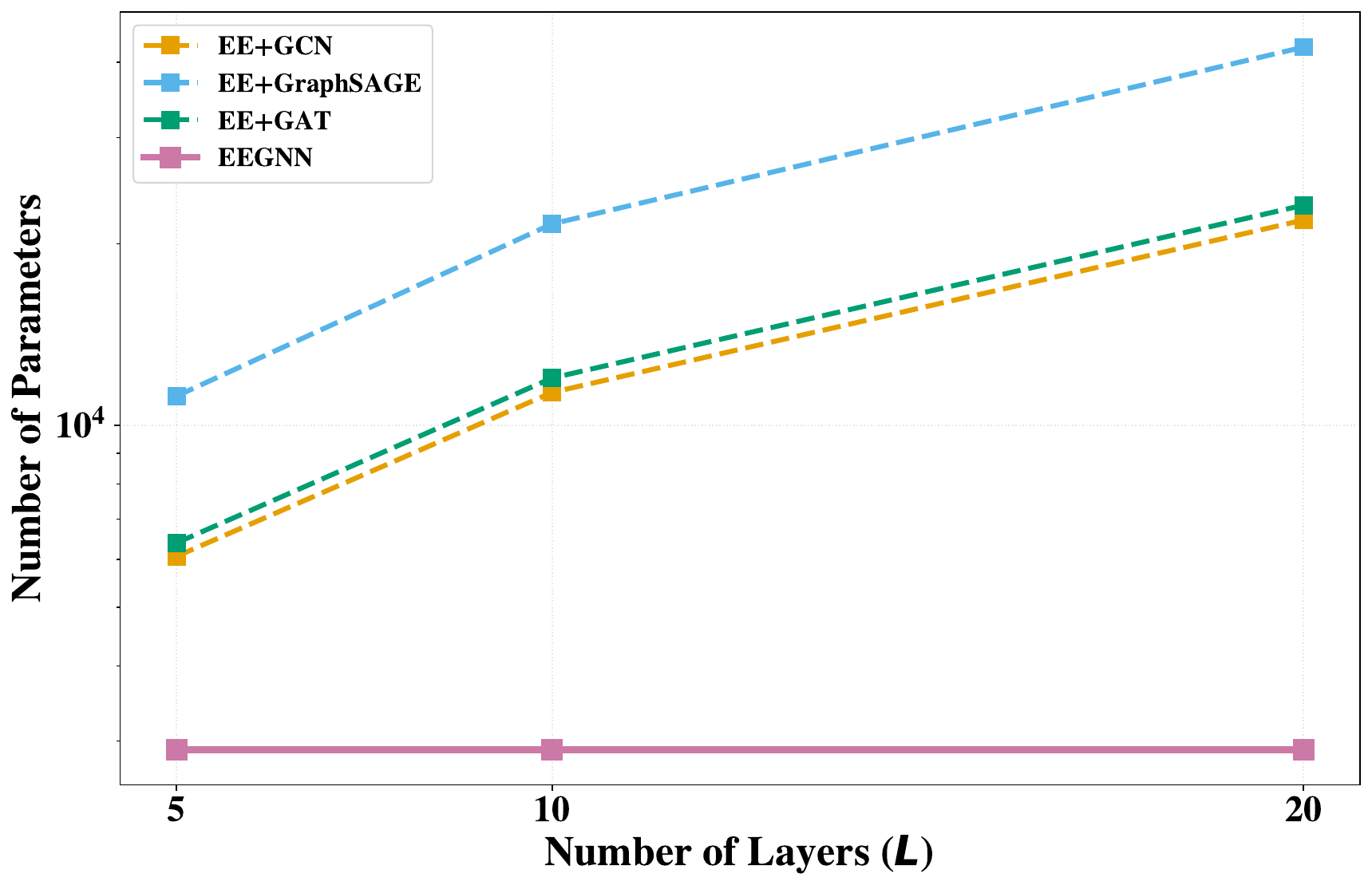}
    \caption{Performance variation based on budget $L$. Exit distributions of EEGNN against the others (right). Number of Parameters evolution based on $L$ (bottom).}
    \label{fig:tolokers_exit_distribution_parameters_performance}
\end{figure}

However, this trend does not always hold. For example, in the \texttt{Tolokers} dataset (Figure~\ref{fig:tolokers_exit_distribution_parameters_performance}), we repeat the same analysis over a restricted interval of $L$ values. Here, EEGNN again shows superior stability and reduced sensitivity to budget variation. At test time, the average exit shifts toward the maximum budget, suggesting that the model tends to fully exploit the available depth. In these case, all the baselines tend to use full model capacity, but without improving the overall performance, and sometimes degrading it. We don't want to affirm that EEGNN always improves, as $L$ increases, but rather show how it never degrades significantly. In this setting, efficiency is not the main objective; rather, the focus is on verifying the model’s ability to process up to a large $L$. Finally, thanks to weight sharing, the parameter count remains constant, enabling improved performance without increasing memory complexity.

\begin{table}[t]
\centering
\caption{Performance of classic GNNs using Gumbel Softmax as early-exit on \texttt{Minesweeper}, \texttt{Questions}, \texttt{Tolokers}. Models are also tested by changing the activation function type.}
\label{tab:activation_comparison_grouped}
\resizebox{\textwidth}{!}{
\begin{tabular}{lcccccc}
\toprule
\textbf{Model} & \multicolumn{2}{c}{\textbf{Minesweeper}} & \multicolumn{2}{c}{\textbf{Questions}} & \multicolumn{2}{c}{\textbf{Tolokers}} \\
\cmidrule(lr){2-3} \cmidrule(lr){4-5} \cmidrule(lr){6-7}
{} & \textbf{ReLU} & \textbf{ReLU + Tanh} & \textbf{ReLU} & \textbf{ReLU + Tanh} & \textbf{ReLU} & \textbf{ReLU + Tanh} \\
\midrule
GCN     & $91.29 \pm 0.5$ & $90.42 \pm 0.6$ & $74.92 \pm 1.4$ & $^*76.56 \pm 1.0$ & $75.58 \pm 3.7$ & $^*84.89 \pm 0.89$ \\
EE+GCN   & $91.07 \pm 0.7$ & $90.39 \pm 0.6$ & $\mathbf{75.56 \pm 1.1}$ & $^*76.33 \pm 1.3$ & $\mathbf{78.02 \pm 5.3}$ & $^*83.54 \pm 1.4$  \\ \hline
SAGE    & $95.61 \pm 0.6$ & $^*96.60 \pm 0.3$ & $75.03 \pm 0.4$ & $^*75.22 \pm 1.0$ & $84.54 \pm 0.65$ & $84.13 \pm 0.42$ \\
EE+SAGE  & $\mathbf{95.73 \pm 0.4}$ & $^*96.56 \pm 0.6$ & $71.36 \pm 1.2$ & $^*72.09 \pm 1.7$ & $76.04 \pm 3.7$ & $74.66 \pm 0.99$ \\ \hline
GAT     & $73.98 \pm 15$   & $^*91.08 \pm 0.5$ & $74.39 \pm 0.9$ & $^*75.53 \pm 1.3$ & $76.84 \pm 7.8$ & $^*84.09 \pm 0.89$ \\
EE+GAT   & $\mathbf{84.17 \pm 14}$  & $^*\mathbf{91.89 \pm 1.5}$ & $\mathbf{74.91 \pm 1.1}$ & $74.90 \pm 2.2$ & $\mathbf{79.02 \pm 3.8}$ & $80.44 \pm 4.0$  \\ \hline
GIN     & $61.92 \pm 1.9$ & $^*89.89 \pm 0.7$ & $50.0 \pm 0.0$   & $^*77.94 \pm 1.2$ & $50.0 \pm 0.0$   & $^*82.55 \pm 1.1$  \\
EE+GIN   & $51.11 \pm 0.3$ & $51.03 \pm 1.6$ & $\mathbf{71.27 \pm 0.8}$ & $70.99 \pm 1.4$ & $50.0 \pm 0.0$   & $^*78.87 \pm 4.3$  \\ \hline
SAS-GNN & - & $93.29 \pm 0.61$ & - & $79.60 \pm 1.15$ &- & $85.80 \pm 0.79$ \\
EEGNN   & - & $93.18 \pm 1.37$ & - & $78.90 \pm 1.15$ &- & $85.26 \pm 0.65$ \\
\bottomrule
\end{tabular}}

\end{table}

\begin{table}[t]
\centering
\caption{Performance of classic GNNs and SAS/EEGNN on Amazon Ratings and Roman Empire datasets. Bold indicates EE variants outperforming the base model.}
\label{tab:activation_comparison_amazon_roman}
\resizebox{0.7\textwidth}{!}{
\begin{tabular}{lcccccc}
\toprule
\textbf{Model} & \multicolumn{2}{c}{\textbf{Amazon Ratings}} & \multicolumn{2}{c}{\textbf{Roman Empire}} \\
\cmidrule(lr){2-3} \cmidrule(lr){4-5}
{} & \textbf{ReLU} & \textbf{ReLU + Tanh} & \textbf{ReLU} & \textbf{ReLU + Tanh} \\
\midrule
GCN     & 49.72 $\pm$ 0.3 & 48.53 $\pm$ 0.4 & 81.16 $\pm$ 0.6 & 76.75 $\pm$ 0.4 \\
\textbf{EE+GCN} & \textbf{51.13 $\pm$ 0.7} & \textbf{*51.25 $\pm$ 0.4} & 79.89 $\pm$ 0.7 & \textbf{79.76 $\pm$ 0.4} \\ \hline
GAT     & 48.40 $\pm$ 0.5 & *49.08 $\pm$ 0.7 & 42.26 $\pm$ 32.4 & *75.86 $\pm$ 1.1 \\
\textbf{EE+GAT} & \textbf{51.88 $\pm$ 0.5} & \textbf{51.85 $\pm$ 0.5} & \textbf{77.21 $\pm$ 0.9} & \textbf{*77.48 $\pm$ 0.8} \\ \hline
SAGE    & 52.06 $\pm$ 0.4 & *52.26 $\pm$ 0.6 & 84.28 $\pm$ 0.8 & *86.55 $\pm$ 0.6 \\
\textbf{EE+SAGE} & 50.51 $\pm$ 0.7 & *50.67 $\pm$ 0.6 & 78.94 $\pm$ 6.4 & *81.10 $\pm$ 1.2 \\ \hline
GIN     & 42.69 $\pm$ 4.3 & *49.50 $\pm$ 0.5 & 13.96 $\pm$ 0.0 & *83.27 $\pm$ 1.1 \\
\textbf{EE+GIN} & \textbf{51.12 $\pm$ 0.8} & \textbf{*51.26 $\pm$ 0.5} & \textbf{65.85 $\pm$ 0.8} & 65.75 $\pm$ 0.8 \\ \hline
SAS-GNN & - & 51.47 $\pm$ 0.7 & - & 83.46 $\pm$ 0.6 \\
EEGNN   & - & 51.54 $\pm$ 0.5 & - & 80.36 $\pm$ 0.4 \\
\bottomrule
\end{tabular}}
\end{table}

\subsection{Additional Results on Accuracy--Efficiency Trade-Off Analysis}
\label{sec:extended_tradeoff}

In this section, we extend our trade-off analysis to further compare EEGNN and SAS-GNN against classical MPNNs, Graph Transformers (GTs), and asynchronous MPNNs. While our models are not always state-of-the-art in terms of raw accuracy, partly due to the stronger regularization and normalization schemes employed by competing methods, we argue that they provide clear benefits in terms of space and time efficiency.  

This analysis differs from the runtime comparison in Table~\ref{tab:runtime_questions}, as here we evaluate models under their best-performing hyperparameter configurations, namely the hidden dimension $m'$ and number of layers $L$ (summarized in Table~\ref{tab:model_configs}). In the main paper, we showed that SAS-GNN and EEGNN achieve competitive accuracy on several benchmarks. Here, we complement those results with experiments on \texttt{Amazon Ratings} and \texttt{Roman Empire}, two datasets where our models underperform relative to GTs or heavily regularized MPNNs.  

\begin{figure}[t]
    \centering
    \begin{subfigure}{0.48\linewidth}
        \centering
        \includegraphics[width=\linewidth]{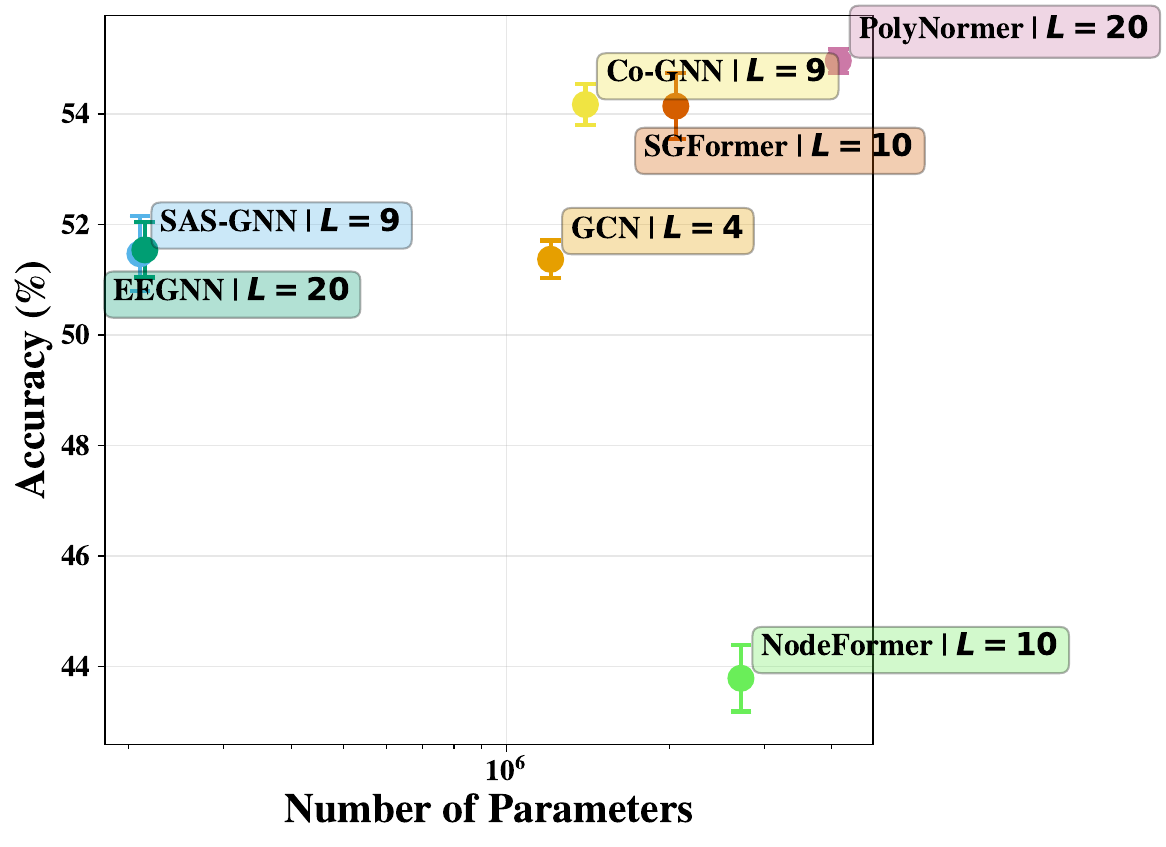}
        \caption{Accuracy vs Parameters}
    \end{subfigure}
    \hfill
    \begin{subfigure}{0.48\linewidth}
        \centering
        \includegraphics[width=\linewidth]{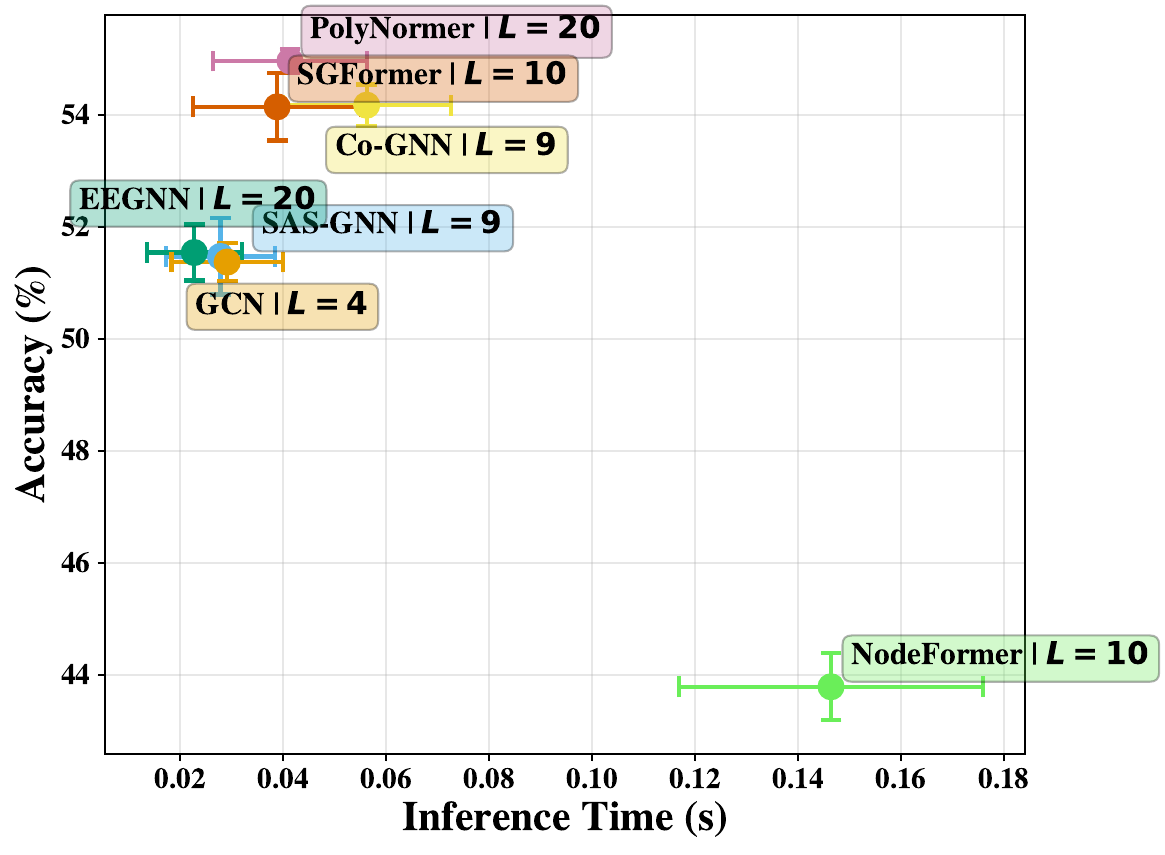}
        \caption{Accuracy vs Inference Time}
    \end{subfigure}
    \caption{Trade-off analysis on \texttt{Amazon Ratings}.}
    \label{fig:tradeoff_amazon_ratings}
\end{figure}

\begin{figure}[t]
    \centering
    \begin{subfigure}{0.48\linewidth}
        \centering
        \includegraphics[width=\linewidth]{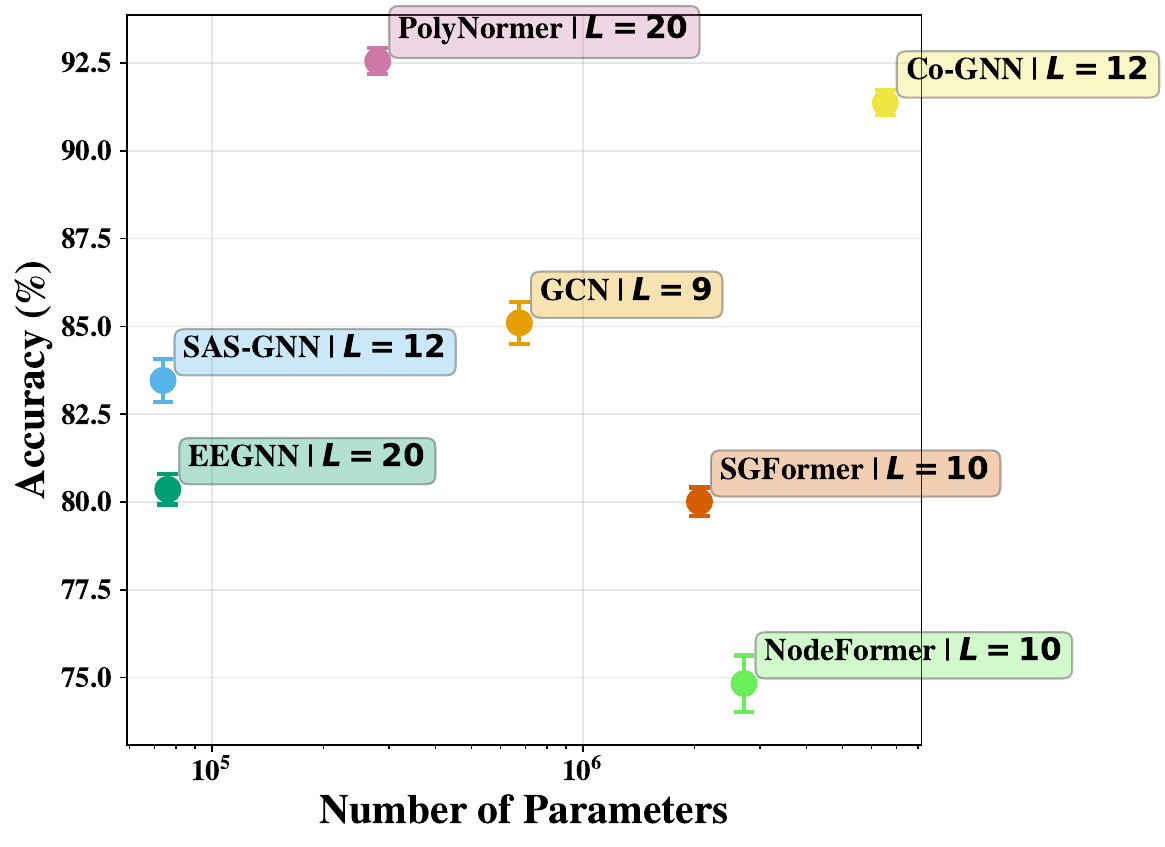}
        \caption{Accuracy vs Parameters}
    \end{subfigure}
    \hfill
    \begin{subfigure}{0.48\linewidth}
        \centering
        \includegraphics[width=\linewidth]{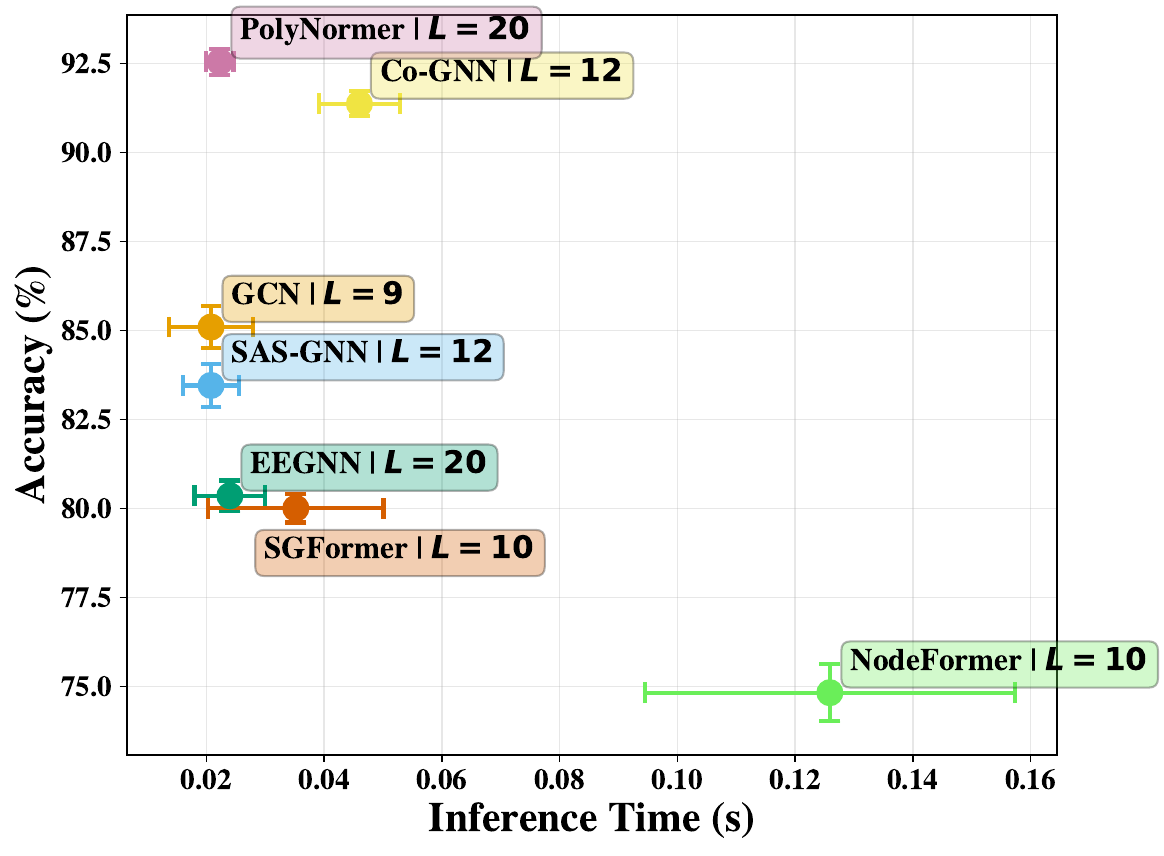}
        \caption{Accuracy vs Inference Time}
    \end{subfigure}
    \caption{Trade-off analysis on \texttt{Roman Empire}.}
    \label{fig:tradeoff_roman_empire}
\end{figure}

The results, illustrated in Figures~\ref{fig:tradeoff_amazon_ratings}--\ref{fig:tradeoff_roman_empire}, highlight the main advantage of our approach. Even when not achieving the highest accuracy, SAS-GNN and EEGNN consistently deliver superior space efficiency, requiring significantly fewer parameters. In terms of inference time, our models are also among the fastest across both datasets, with the sole exception of GCN on \texttt{Roman Empire}. These findings reinforce our claim that EEGNN and SAS-GNN strike a favorable balance between accuracy, scalability, and efficiency.  

\begin{table}[t]
\centering
\caption{Model configurations across \texttt{Questions}, \texttt{Roman Empire}, and \texttt{Amazon Ratings} during trade-off analysis. We report hidden dimension $m'$ and number of layers $L$.}
\label{tab:model_configs}
\resizebox{0.7\textwidth}{!}{%
\begin{tabular}{lccc}
\toprule
Model & \texttt{Questions} $(m',L)$ & \texttt{Roman Empire} $(m',L)$ & \texttt{Amazon Ratings} $(m',L)$ \\
\midrule
SAS-GNN    & (64, 9)   & (128, 12)  & (256, 9)  \\
EEGNN      & (64, 20)  & (128, 20)  & (256, 20) \\
Co-GNN     & (64, 9)   & (512, 12)  & (256, 9)  \\
PolyNormer & (64, 10)  & (64, 20)   & (256, 20) \\
SGFormer   & (64, 10)  & (256, 10)  & (64, 10)  \\
NodeFormer & (64, 10)  & (256, 10)  & (64, 10)  \\
GCN        & (512, 10) & (256, 9)   & (512, 4)  \\
\bottomrule
\end{tabular}}
\end{table}
\subsection{Additional Results on APGCN vs. EEGNN}
\label{sec:apgcnvseegnn}
\begin{figure}[t]
    \centering
    \includegraphics[width=0.7\linewidth]{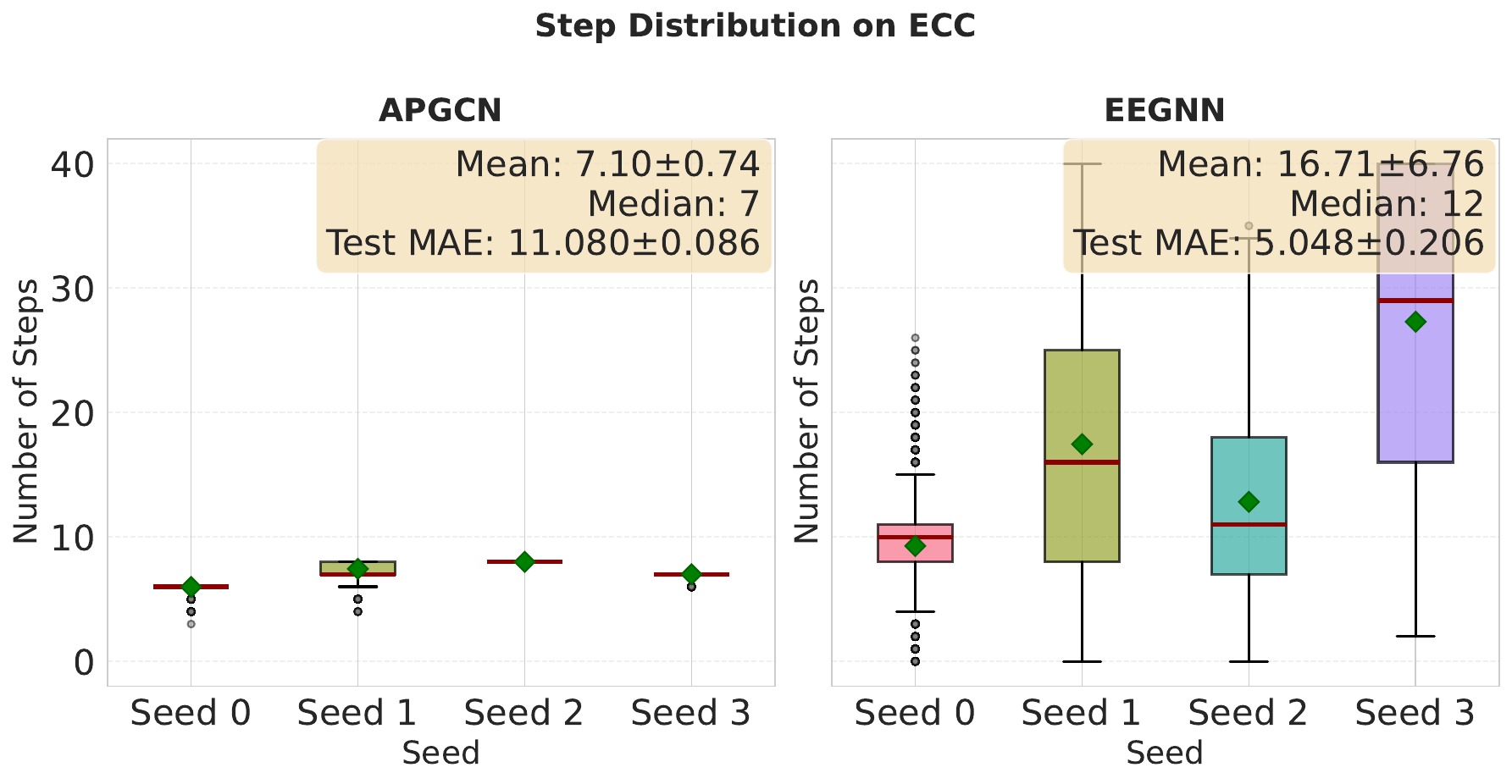}
    \caption{Distribution of exit steps on the \texttt{Eccentricity} dataset over 4 seeds. \textbf{Left:} APGCN exits prematurely (Mean $\approx 7$) leading to high error. \textbf{Right:} EEGNN adapts to the task's long-range requirements by sustaining deeper propagation (Mean $\approx 17$), significantly reducing the Test MAE.}
    \label{fig:ecc_step_distributions}
\end{figure}

As we have analysed and commented in the main text, the advantage of our EEGNN is that it correctly update the exit module based on the task, and this gives a potential to identify the correct exit point, as well as preserving feature informativeness when it exits. We now analyse such a behavior on the \texttt{Eccentricity} dataset, where a similar trend is observed. As illustrated in Figure~\ref{fig:ecc_step_distributions}, APGCN fails to propagate information sufficiently, exiting early with an average depth of only $7.10 \pm 0.74$, resulting in a high Test MAE of $11.08$. In contrast, EEGNN correctly identifies the complexity of the task, adapting to a significantly deeper regime with an average of $16.71 \pm 6.76$ steps, an average value even higher than that observed for \texttt{SSSP}. While the performance is not absolute state-of-the-art, the error is reduced by over 50\% compared to APGCN (MAE $5.05$). This confirms that when deep processing is required, but we don't want to manually tune the depth it is crucial to learn how to select the exit point, and having stable intermediate representations. EEGNN leverages these representations to sustain deeper propagation without succumbing to the premature exiting behavior typical of budget-constrained mechanisms.

\subsection{Results on Homophilic Node Classification}
\label{sec:extended_homophily}
To ensure completeness in terms of the graph types used for testing, we evaluate our models on standard homophilic node classification datasets. Our goal is to assess whether our models remain competitive, achieving performance comparable to classic GNNs. These benchmarks typically do not require deep architectures to achieve high accuracy, so we do not expect our methods to provide a clear advantage over shallow baselines in this setting. In fact, we propose SAS-GNN, and EEGNN as a solution when we want to extend MPNNs to a deep regime, which is not required to succeed in this set of datasets.

We evaluate our proposed models, \textbf{SAS-GNN} and \textbf{EEGNN}, on transductive homophilic node classification on the datasets \texttt{Computer}, \texttt{CS}, \texttt{Photo}, and \texttt{Physics}, introduced by~\citep{homophilicncdatasets}. We report the mean accuracy and standard deviation over 10 trials, with dataset statistics summarized in Table~\ref{tab:homophilicnc_statistics}.

The baselines that we use are both message-passing methods, specifically the revised version of \citet{classicgnnsstrongbaselines} with respective ablations of GCN, GraphSAGE, and GAT, as well as Graph Transformers (GT), such as NAGphormer~\cite{nagphormer}, Exphormer~\cite{Exphormer}, GOAT~\cite{GOAT}, Polynormer~\cite{polynormer}, and SGFormer~\cite{sgformer}.

We used the same hyperparameter search space proposed by~\citet{classicgnnsstrongbaselines}. Specifically, their configurations heavily rely on the use of normalization and dropout layers within the message-passing pipeline, techniques we deliberately avoid in favor of a principled design that emphasizes structural simplicity.

In Table~\ref{tab:results_homophilicnc}, we show our models' performance. We observe that they are never among the top-2 models. This is not surprising, since the other models in the benchmark leverage normalization and dropout layers, while our design is focused on efficiently handling long-range information propagation. Nevertheless, both \textbf{SAS-GNN} and \textbf{EEGNN} perform competitively, trailing the top models only marginally and outperforming several Transformer-based and classic baselines. These results further support the adaptability and robustness of our methods across diverse graph learning settings.
\begin{table}[t]
\centering
\caption{Statistics of the homophlic node classification datasets.}
\label{tab:homophilicnc_statistics}
\begin{tabular}{lcccc}
\toprule
& \texttt{Computer} & \texttt{Photo} & \texttt{CS} & \texttt{Physics} \\
\midrule
\# nodes        & 13,752   & 7,650    & 18,333  & 34,493  \\
\# edges        & 245,861  & 119,081  & 81,894  & 247,962 \\
\# features     & 767      & 745      & 6,805   & 8,415   \\
\# classes      & 10       & 8        & 15      & 5       \\
metric          & Accuracy & Accuracy & Accuracy & Accuracy \\
\bottomrule
\end{tabular}
\end{table}

\begin{table}[t]
\centering
\caption{Performance on Node classification datasets. These scores are taken from \cite{classicgnnsstrongbaselines}. The scores are marked in red for the \textcolor{red}{first}, blue for the \textcolor{blue}{second}.}
\label{tab:results_homophilicnc}
\begin{tabular}{lcccc}
\toprule
{} & \texttt{Computer} & \texttt{Photo} & \texttt{CS} & \texttt{Physics} \\
\midrule
NAGphormer & $91.69 \pm 0.30$ & $96.14 \pm 0.16$ & $95.85 \pm 0.16$ & \textcolor{blue}{$97.35 \pm 0.12$} \\
Exphormer & $91.47 \pm 0.17$ & $95.35 \pm 0.22$ & $94.93 \pm 0.01$ & $96.89 \pm 0.09$ \\
GOAT & $92.29 \pm 0.37$ & $94.33 \pm 0.21$ & $93.81 \pm 0.19$ & $96.47 \pm 0.16$ \\
GraphGPS & $91.79 \pm 0.63$ & $94.89 \pm 0.14$ & $94.04 \pm 0.21$ & $96.71 \pm 0.15$ \\
Polynormer & $93.78 \pm 0.10$ & $96.57 \pm 0.23$ & $95.42 \pm 0.19$ & $97.18 \pm 0.11$ \\
SGFormer & $91.99 \pm 0.76$ & $95.10 \pm 0.47$ & $94.78 \pm 0.20$ & $96.60 \pm 0.18$ \\
GCN* & \textcolor{blue}{$93.99 \pm 0.12$} & $96.10 \pm 0.46$ & $96.17 \pm 0.06$ & \textcolor{blue}{$97.46 \pm 0.10$} \\
\hspace{0.5em} -- Norm & $92.60 \pm 0.14$ & $95.48 \pm 0.36$ & $95.30 \pm 0.05$ & $97.16 \pm 0.11$ \\
\hspace{0.5em} -- Dropout & $93.78 \pm 0.26$ & $95.31 \pm 0.10$ & $95.95 \pm 0.13$ & $97.30 \pm 0.06$ \\
GraphSAGE* & $93.25 \pm 0.14$ & \textcolor{red}{$96.78 \pm 0.23$} & \textcolor{red}{$96.38 \pm 0.11$} & $97.19 \pm 0.05$ \\
\hspace{0.5em} -- Norm & $92.77 \pm 0.63$ & $95.51 \pm 0.14$ & $95.42 \pm 0.17$ & $96.97 \pm 0.07$ \\
\hspace{0.5em} -- Dropout & $92.02 \pm 0.35$ & $96.03 \pm 0.27$ & $96.11 \pm 0.17$ & $97.07 \pm 0.09$ \\
GAT* & \textcolor{red}{$94.09 \pm 0.37$} & \textcolor{blue}{$96.60 \pm 0.33$} & \textcolor{blue}{$96.21 \pm 0.14$} & $97.25 \pm 0.06$ \\ 
\hspace{0.5em} -- Norm & $93.22 \pm 1.27$ & $96.09 \pm 0.20$ & $95.13 \pm 0.35$ & $97.08 \pm 0.04$ \\
\hspace{0.5em} -- Dropout & $93.14 \pm 0.29$ & $96.36 \pm 0.25$ & $96.05 \pm 0.09$ & $97.01 \pm 0.05$ \\ \hline
SAS-GNN & $92.27 \pm 0.29$ & $96.47 \pm 0.33$ & $94.46 \pm 0.10$ & $96.49 \pm 0.10$ \\
EEGNN & $90.24 \pm 0.93$ & $95.23 \pm 0.35$ & $95.56 \pm 0.07$ & $96.22 \pm 0.09$ \\
\bottomrule
\end{tabular}
\end{table}

\subsection{Results on TUDataset benchmark for Graph Classification}
\label{sec:extended_graphclass}
To ensure completeness, we evaluate our models on standard graph classification datasets. Our goal is to assess whether our models remain competitive, achieving performance comparable to classic GNNs. These benchmarks typically do not require deep architectures to achieve high accuracy, so we do not expect our methods to provide a clear advantage over shallow baselines in this setting.

We evaluate our proposed models, SAS-GNN and EEGNN, on the TUDataset graph classification benchmark~\citep{tudataset}, following the protocol established by~\citet{coognns}. We report the mean accuracy and standard deviation across seven datasets, with dataset statistics summarized in Table~\ref{tab:tudataset_statistics}. These results were directly taken from \citet{coognns}.

Our models are compared against a broad range of baselines, including established GNNs such as DGCNN~\citep{dgcnn}, DiffPool~\citep{diffpool}, ECC~\citep{ecc}, CGMM~\citep{cgmm}, ICGMMf~\citep{icgmm}, SPN~\citep{spn}, and GSPN~\citep{gspn}. We also include results for Co-GNN~\citep{coognns} for direct comparison.

To ensure fairness, we adopt the same hyperparameter search space and evaluation pipeline used in previous work. Configurations that failed due to excessive memory or runtime demands are marked as \texttt{OOR} (Out of Resources) in the results tables.

As shown in Table~\ref{tab:results_tudataset}, both SAS-GNN and EEGNN achieve competitive or superior performance across datasets. For instance, our models match or outperform Co-GNN on datasets like \texttt{IMDB-B} and \texttt{PROTEINS}. Notably, despite using simple mean pooling, our models sometimes surpass approaches with more complex pooling strategies such as DiffPool. Furthermore, EEGNN improves upon SAS-GNN on the \texttt{ENZYMES} dataset, despite the high variance typically observed there. The exception is \texttt{NCI1}, where exiting before the full-depth lead to a significant performance degradation.
\begin{table}[t]
\centering
\caption{Dataset statistics for the TUDataset benchmark for graph classification.}
\label{tab:tudataset_statistics}
\resizebox{\textwidth}{!}{
\begin{tabular}{lccccccc}
\toprule
& \texttt{IMDB-B} & \texttt{IMDB-M} & \texttt{REDDIT-B} & \texttt{REDDIT-M} & \texttt{NCI1} & \texttt{PROTEINS} & ENZYMES \\
\midrule
\# graphs         & 1000   & 1500   & 2000& 4999   & 4110  & 1113     & 600    \\
\# avg. nodes     & 19.77  & 13.00  & 429.63 &508.51 & 29.87 & 39.06    & 32.63  \\
\# avg. edges     & 96.53  & 65.94  & 497.75 &1189.74 & 32.30 & 72.82    & 64.14  \\
\# classes        & 2      & 3      & 2   & 5   & 2     & 2        & 6      \\
metrics           & Accuracy    & Accuracy & Accuracy   & Accuracy     & Accuracy   & Accuracy      & Accuracy    \\
\bottomrule
\end{tabular}}
\end{table}
\begin{table}[t]
\centering
\caption{Graph classification results on the TUDataset benchmark. The scores are marked in red for the \textcolor{red}{first}, blue for the \textcolor{blue}{second}.}
\label{tab:results_tudataset}
\resizebox{\textwidth}{!}{
\begin{tabular}{l c c c c c c c}
\hline
\textbf{} & \texttt{IMDB-B} & \texttt{IMDB-M}& \texttt{REDDIT-B}& \texttt{REDDIT-M} & \texttt{NCI1} & \texttt{PROTEINS} & \texttt{ENZYMES} \\
\hline
DGCNN & 69.2 $\pm$ 3.0 & 45.6 $\pm$ 3.4 & 87.8 $\pm$ 2.5 & 49.2 $\pm$ 1.2 & 76.4 $\pm$ 1.7 & 72.9 $\pm$ 3.5 & 38.9 $\pm$ 5.7 \\
DiffPool & 68.4 $\pm$ 3.3 & 45.6 $\pm$ 3.4 & 89.1 $\pm$ 1.6 & 53.8 $\pm$ 1.4 & 76.9 $\pm$ 1.9 & \textcolor{blue}{73.7 $\pm$ 3.5} & 59.5 $\pm$ 5.6 \\
ECC & 67.7 $\pm$ 2.8 & 43.5 $\pm$ 3.1 & OOR & OOR & 76.2 $\pm$ 1.4 & 72.3 $\pm$ 3.4 & 29.5 $\pm$ 8.2 \\
GIN & 71.2 $\pm$ 3.9 & 48.5 $\pm$ 3.3 & 89.9 $\pm$ 1.9 & \textcolor{blue}{56.1 $\pm$ 1.7} & \textcolor{red}{80.0 $\pm$ 2.4} & \textcolor{red}{75.3 $\pm$ 2.5} & 59.6 $\pm$ 4.5 \\
GraphSAGE & 68.8 $\pm$ 4.5 & 47.6 $\pm$ 3.5 & 84.3 $\pm$ 1.9 & 50.0 $\pm$ 3.6 & 74.9 $\pm$ 1.9 & 70.5 $\pm$ 3.2 & 58.2 $\pm$ 6.0 \\
CGMM & - & - & 88.1 $\pm$ 1.9 & - & - & - & - \\
ICGMMf & 71.8 $\pm$ 4.4 & \textcolor{blue}{49.0 $\pm$ 3.8} & \textcolor{red}{91.6 $\pm$ 2.1} & 55.6 $\pm$ 1.7 & 76.4 $\pm$ 1.3 & 73.2 $\pm$ 3.9 & - \\
SPN($k = 5$) & - & - & - & - & 74.2 $\pm$ 2.7 & - & \textcolor{blue}{69.4 $\pm$ 6.2} \\
GSPN & - & - & \textcolor{blue}{90.5 $\pm$ 1.1} & 55.3 $\pm$ 2.0 & 76.6 $\pm$ 1.9 & - & - \\
Co-GNN & \textcolor{blue}{72.2 $\pm$ 4.1} & \textcolor{red}{49.9 $\pm$ 4.5} & \textcolor{blue}{90.5 $\pm$ 1.9} & \textcolor{red}{56.3 $\pm$ 2.1} & \textcolor{blue}{79.4 $\pm$ 0.7} & 71.3 $\pm$ 2.0 & 68.3 $\pm$ 5.7 \\ \hline
SAS-GNN & \textcolor{red}{72.3 $\pm$ 2.9} & 48.0 $\pm$ 4.5 & 88.3 $\pm$ 2.3 & 55.1 $\pm$ 2.2 & 77.9 $\pm$ 1.8& 72.0 $\pm$ 3.0 & 66.5 $\pm$ 5.9 \\
EEGNN & 71.2 $\pm$ 3.9 & 46.8 $\pm$ 4.4 & 86.5 $\pm$ 2.8 & 53.9 $\pm$ 2.4 & 62.53 $\pm$ 4.3& 70.8 $\pm$ 2.7 & \textcolor{red}{70.3 $\pm$ 7.3} \\
\hline
\end{tabular}}

\label{tab:graph_classification}
\end{table}

\subsection{Experiments on Open Graph Benchmark datasets for Node and Graph Classification}
\label{sec:ogb_results}

To further address reviewer concerns regarding scalability and evaluation on larger benchmarks, we complement our previous experiments with results on two widely used datasets from the Open Graph Benchmark (OGB) \citep{ogb} suite: \texttt{ogbn-arxiv} (node classification) and \texttt{ogbg-molhiv} (graph classification). These datasets are representative of two key scaling regimes: a single large graph with over 169k nodes and 1.1M edges (\texttt{ogbn-arxiv}), and a large collection of over 41k molecular graphs with an average of 25.5 nodes and 27.5 edges per graph (\texttt{ogbg-molhiv}). 

While there are currently no publicly available OGB datasets that explicitly address the scenarios that we want to scrutinizeize to test long-range dependency or heterophilic graphs, these two tasks allow us to test model robustness with respect to graph size and number of graphs. 

We evaluate both SAS-GNN and EEGNN, comparing them against standard GNN baselines and their ablated versions without dropout or normalization layers, following \citet{classicgnnsstrongbaselines}. Similar to Tables \ref{tab:extended_result_heterophilic} and \ref{tab:result_lrgb_extended}, we show that high performance is typically due to the use of dropout and normalization layers. When removing these components in the ogb dataset, the other baselines lag behind our model, which still manages to perform when learning on large-scale datasets.

\paragraph{Node classification on \texttt{ogbn-arxiv}.} 
Results are reported in Table~\ref{tab:ogb_arxiv_results}. Our methods achieve competitive accuracy compared to GCN*, GraphSAGE*, and GAT*, when they are ablated by normalization and dropout. This is not the case with Polynormer, that still outperform the other models. However, we manage to improve over two transformer baselines, namely GraphGPS and NAGphormer. This highlights the scalability of our approach to graphs exceeding one million edges. 

\begin{table}[t]
\centering
\caption{Results on \texttt{ogbn-arxiv} (node classification). Dataset: 169,343 nodes, 1,166,243 edges. Baselines and ablations are reported from \citet{classicgnnsstrongbaselines}.}
\label{tab:ogb_arxiv_results}
\begin{tabular}{lcc}
\toprule
Model & Accuracy $\uparrow$ \\
\midrule
Polynormer & $73.46 \pm 0.16$ \\
GCN* & $73.53 \pm 0.12$ \\
\hspace{0.5em} -- Dropout & $72.06 \pm 0.13$ \\
\hspace{0.5em} -- Normalization & $71.53 \pm 0.14$ \\
GraphSAGE* & $73.00 \pm 0.28$ \\
\hspace{0.5em} -- Dropout & $71.30 \pm 0.21$ \\
\hspace{0.5em} -- Normalization & $71.13 \pm 0.27$ \\
GAT* & $73.30 \pm 0.18$ \\
\hspace{0.5em} -- Dropout & $71.68 \pm 0.32$ \\
\hspace{0.5em} -- Normalization & $71.33 \pm 0.29$ \\
GraphGPS & $70.97 \pm 0.41$ \\
NAGphormer & $70.13 \pm 0.55$ \\
\midrule
SAS-GNN & $71.83 \pm 0.20$ \\
EEGNN & $71.47 \pm 0.40$ \\
\bottomrule
\end{tabular}
\end{table}

\paragraph{Graph classification on \texttt{ogbg-molhiv}.} 
Results are reported in Table~\ref{tab:ogb_molhiv_results}. This dataset provides molecular edge features, which we incorporate following the methodology formalized in Theorem~\ref{theo:edge_features}, and we distinguish between models with and without these features (denoted as \texttt{edge} vs. \texttt{ours}). More specifically: SAS-GNN$_{noedge}$ ($f_e(\mathbf{E}) = 0$), SAS-GNN$_{edge}$ ($f_e(\mathbf{E}) = \mathbf{BEW}_e$), and SAS-GNN/EEGNN$_{ours}$ ($f_e(\mathbf{E}) = -\text{ReLU}(\mathbf{BEW}_e)$).

We note that \texttt{GCN$^+$} and \texttt{GatedGCN$^+$} are not simply the classical models, but rather stronger variants that explicitly add dropout, normalization, and additional feedforward networks. For this reason, we regard them as distinct baselines. Through ablation, it becomes clear that the performance gap between these variants and our methods can largely be attributed to such auxiliary components rather than fundamental architectural differences, which is what we are rather more interested.  

Relative to the standard GCN and GatedGCN (without enhancements), both SAS-GNN and EEGNN achieve higher AUROC scores, confirming the benefit of our design. When including the edge features validated by our theoretical results, our methods further improve, aligning with the theoretical justification that our edge-aware message passing better leverages structural information. While we trail slightly behind the heavily tuned \texttt{GCN$^+$} and \texttt{GatedGCN$^+$}, our models remain competitive without requiring other components.

\begin{table}[t]
\centering
\caption{Results on \texttt{ogbg-molhiv} (graph classification). Dataset: 41,127 graphs, avg. 25.5 nodes, 27.5 edges. Baselines and ablations are reported from \citet{classicgnnsstrongbaselines}.}
\label{tab:ogb_molhiv_results}
\begin{tabular}{lcc}
\toprule
Model & AUROC $\uparrow$ \\
\midrule
GraphGPS & $78.80 \pm 1.01$ \\
GCN	&76.06 ± 0.97 \\
GatedGCN	& 76.87 ± 1.36 \\
GCN$^+$ & $80.12 \pm 1.24$ \\
\hspace{0.5em} -- Dropout & $74.31 \pm 1.85$ \\
\hspace{0.5em} -- Normalization & $77.53 \pm 0.49$ \\
GatedGCN$^+$ & $80.40 \pm 1.64$ \\
\hspace{0.5em} -- Dropout & - \\
\hspace{0.5em} -- Normalization & $78.79 \pm 1.78$ \\
\midrule
SAS-GNN$_{edge}$ & $77.26 \pm 1.60$ \\
EEGNN$_{edge}$ & $77.39 \pm 1.00$ \\
SAS-GNN$_{ours}$ & $78.09 \pm 0.50$ \\
EEGNN$_{ours}$ & $78.05 \pm 0.80$ \\
\bottomrule
\end{tabular}
\end{table}

\paragraph{Discussion.} 
Across both benchmarks, our methods remain competitive with state-of-the-art baselines and scale effectively to graphs with over a million edges or tens of thousands of graphs. While tuned variants of classical GNNs with dropout and normalization achieve the strongest results, our models perform well without these components, supporting the robustness of our principled design. We note that publicly available OGB datasets do not yet include large-scale heterophilic graphs or verified long-range benchmarks, and we regard this as an important direction for future evaluation.

\subsection{Extended Runtime Analysis} \label{appendix:run_time_analysis}
In Table \ref{tab:runtime-datasets}, we show our extended runtime analysis, and in Table \ref{tab:parameter-datasets-models}.

\begin{table}[t]
    \centering
    \caption{Runtime Analysis Across Datasets and Layers. }
    \resizebox{\textwidth}{!}{
    \begin{tabular}{l|c|c|c|c|c|c|c|c}
        \hline
        \multirow{2}{*}{\textbf{Model}} 
        & \multicolumn{2}{c|}{\textbf{Roman Empire}} 
        & \multicolumn{2}{c|}{\textbf{Minesweeper}} 
        & \multicolumn{2}{c|}{\textbf{Tolokers}} 
        & \multicolumn{2}{c}{\textbf{Amazon Ratings}} \\
        \cline{2-9}
        & \textbf{10 Layers} & \textbf{20 Layers} 
        & \textbf{10 Layers} & \textbf{20 Layers} 
        & \textbf{10 Layers} & \textbf{20 Layers} 
        & \textbf{10 Layers} & \textbf{20 Layers} \\
        \hline
        \textbf{GCN}     
        & $0.0266 \pm 0.0108$ & $0.0411 \pm 0.0147$ 
        & $0.0139 \pm 0.0082$ & $0.0273 \pm 0.0119$ 
        & $0.0286 \pm 0.0099$ & $0.0428 \pm 0.0124$ 
        & $0.0269 \pm 0.0113$ & $0.0370 \pm 0.0140$ \\
        \textbf{Co-GNN}  
        & $0.0553 \pm 0.0167$ & $0.0739 \pm 0.0253$ 
        & $0.0315 \pm 0.0091$ & $0.0665 \pm 0.0221$ 
        & $0.0498 \pm 0.0156$ & $0.0874 \pm 0.0225$ 
        & $0.0562 \pm 0.0163$ & $0.0838 \pm 0.0262$ \\
        \textbf{Polynormer}  
        & $ 0.0192 \pm 0.0031$ & $0.0312 \pm 0.0045$ 
        & $ 0.01509\pm0.0022 $ & $ 0.0274\pm0.0030$ 
        & $ 0.0183 \pm  0.0025$ & $ 0.0322 \pm 0.0045$ 
        & $0.0191 \pm 0.0037$ & $0.0310 \pm 0.0046$  \\
        \textbf{SAS-GNN} 
        & $0.0323 \pm 0.0126$ & $0.0510 \pm 0.0156$ 
        & $0.0177 \pm 0.0080$ & $0.0371 \pm 0.0123$ 
        & $0.0268 \pm 0.0098$ & $0.0437 \pm 0.0131$ 
        & $0.0278 \pm 0.0106$ & $0.0442 \pm 0.0150$ \\
        \textbf{EEGNN}   
        & $0.0288 \pm 0.0107$ & $0.0290 \pm 0.0117$ 
        & $0.0153 \pm 0.0108$ & $0.0209 \pm 0.0111$ 
        & $0.0227 \pm 0.0120$ & $0.0283 \pm 0.0126$ 
        & $0.0267 \pm 0.0128$ & $0.0227 \pm 0.0092$ \\
        
               \hline
    \end{tabular}}
    
    \label{tab:runtime-datasets}
\end{table}

\begin{table}[t]
    \centering
    \caption{Number of Parameters Across Datasets and Layers.}
    \label{tab:parameter-datasets-models}
    \resizebox{\textwidth}{!}{
    \begin{tabular}{l|c|c|c|c|c|c|c|c}
        \hline
        \multirow{2}{*}{\textbf{Model}} 
        & \multicolumn{2}{c|}{\textbf{Roman Empire}} 
        & \multicolumn{2}{c|}{\textbf{Minesweeper}} 
        & \multicolumn{2}{c|}{\textbf{Tolokers}} 
        & \multicolumn{2}{c}{\textbf{Amazon Ratings}} \\
        \cline{2-9}
        & \textbf{10 Layers} & \textbf{20 Layers} 
        & \textbf{10 Layers} & \textbf{20 Layers} 
        & \textbf{10 Layers} & \textbf{20 Layers} 
        & \textbf{10 Layers} & \textbf{20 Layers} \\
        \hline
        \textbf{GCN}     
        & 20,768 & 31,328  
        & 10,880 & 21,440  
        & 10,976 & 21,536  
        & 20,352 & 30,912  \\
        \textbf{Co-GNN}  
        & 35,478 & 56,278  
        & 25,574 & 46,374  
        & 25,670 & 46,470  
        & 35,049 & 55,849  \\
        \textbf{Polynormer}  
        & 48,164  &  81,124 
        &  37,732 & 70,692
        &  37,828 & 70,788
        &  47,306 & 80,266 \\
        \textbf{SAS-GNN} 
        & 12,320 & 12,320  
        & 2,432  & 2,432   
        & 2,528  & 2,528   
        & 11,904 & 11,904  \\
        \textbf{EEGNN}   
        & 14,562 & 14,562  
        & 4,674  & 4,674   
        & 4,770  & 4,770   
        & 14,146 & 14,146  \\
        \hline
    \end{tabular}}
    
\end{table}

We confirm our claims in Section \ref{sec:impact_earlyexit}, where we show that not only are SAS-GNN and EEGNN parameter-efficient, but EEGNN preserves its time complexity even when allowing it to go deeper, implementing it with a higher $L$ budget (i.e., from $L=10$ to $L=20$). This analysis can be seen as a way to say that EEGNN has constant time complexity w.r.t. the number of layers. This is simply a result of the learning algorithm that does not let the nodes/graph exit distribution change when the budget has increased. However, whenever we record an increased inference time by EEGNN, it could mean that the nodes/graphs require could benefit from more processing time, and increasing $L$ allows the model to test deeper configurations.

\subsection{Promising Directions for Early-Exit and GNNs}
\label{sec:oracle}
In addition to the results presented in the main paper, we aimed to explore whether implementing early-exit mechanisms in GNNs has space for accuracy improvements. \cite{AdaProp} provides evidence of accuracy enhancements within their experiments, but their experimental setup differs from ours; in particular, we consider heterophilic graphs and do not use any additional losses. However, due to the depth robustness of SAS-GNN, we hypothesize that exiting early or late may offer no significant accuracy gains, as the network's performance is stable across varying depths.
To assess this hypothesis, we designed an oracle-like evaluation for SAS-GNN. In this setup, the network `knows' the optimal exit point for each node, defined as the point where its latent representation would be classified correctly. Our goal is to compare this oracle-based performance with that of a standard SAS-GNN. If the accuracy remains the same, we conclude that EEGNN does not provide any inherent accuracy advantage.
More specifically, for each node, we track the logits produced at each possible exit point and check whether they yield a correct prediction. If no intermediate layer produces a correct output, we default to the logits from the final layer. The results of this analysis are summarized in Table \ref{tab:oracle_like}.
\begin{table}[t]
\centering
\caption{Metrics computed at the Optimal Exit.}
\label{tab:oracle_like}
\begin{tabular}{l c c c}
\toprule
\textbf{Metrics}            & \textbf{No Exit} &\textbf{Optimal Exit} \\
\midrule
\texttt{Amazon Ratings}     &    51.47 ± 0.68 & 68.36 ± 0.86          \\
\texttt{Tolokers}            &  85.80 ± 0.79 & 89.21 ± 1.37       \\
\texttt{Questions}            &  79.60 ± 1.15 & 80.35 ± 1.29          \\
\texttt{Roman Empire} & 83.46 ± 0.61& 89.45 ± 0.37 \\ 
\texttt{Minesweeper} & 93.29 ± 0.61 & 96.93 ± 0.58 \\
\bottomrule
\end{tabular}
\end{table}
As seen in the Table, individual nodes often arrive at correct predictions earlier in the forward pass, but when forced to process through all layers, the overall performance drops compared to the optimal exit scenario. This aligns with the `over-thinking' phenomenon frequently observed in early-exit neural networks \citep{overthinking}, where intermediate layers produce accurate predictions, but later layers overwrite them with incorrect outputs.
Now that GNNs have made progress in mitigating simple message-passing flaws and deepening architectures, these results highlight the potential of EEGNNs to further enhance performance by integrating early-exit strategies. We believe that this promising direction opens new possibilities for more efficient and accurate GNNs. So far, we observed that our current framework tends only to reconstruct the performance of SAS-GNN, but the results presented in Table \ref{tab:oracle_like} let us believe that EEGNN can impact more than what is currently achieved.
\begin{table}[t]
\centering
\caption{Comparison of the fixed number of layers selected by SAS-GNN (i.e., \textbf{Fixed \# Layers}) with the minimum, median, and maximum number of layers selected by EEGNN (\textbf{Min. Layers}, \textbf{Median Layers}, \textbf{Max. Layers}). The values are averaged across 10 test splits and are taken from the discrete exit point distributions.}
\begin{tabular}{l c c c c}
\hline
\textbf{Datasets}              & \textbf{Fixed \# Layers} & \textbf{Min. Layers} & \textbf{Median Layers} & \textbf{Max. Layers} \\ \hline
\texttt{Amazon Ratings} &        10           &  5.1 $\pm$ 2.64                       &         13.8 $\pm$ 1.39                 &   21.0 $\pm$ 0.00                    \\ 
\texttt{Roman Empire}   &         12          &                  1.7 $\pm$ 0.48     &              3.3 $\pm$ 0.48            &  10.2 $\pm$ 4.84                     \\ 
\texttt{Minesweeper}    &      15           &                     11.4 $\pm$ 10.13   &         15.6 $\pm$ 8.69                 & 15.7 $\pm$ 8.53                        \\ 
\texttt{Questions}      &     9              & 1.0 $\pm$ 0.00                       &     7.0 $\pm$ 0.66                     &  18.8 $\pm$ 2.78                     \\ 
\texttt{Tolokers}       &         10          &  1.7 $\pm$ 0.82        &   8.3 $\pm$ 2.00         &     21.0 $\pm$ 0.00        \\ \hline
\end{tabular}
\label{tab:node_distributions}
\end{table}
\begin{figure}
    \centering
    \includegraphics[width=0.5\linewidth]{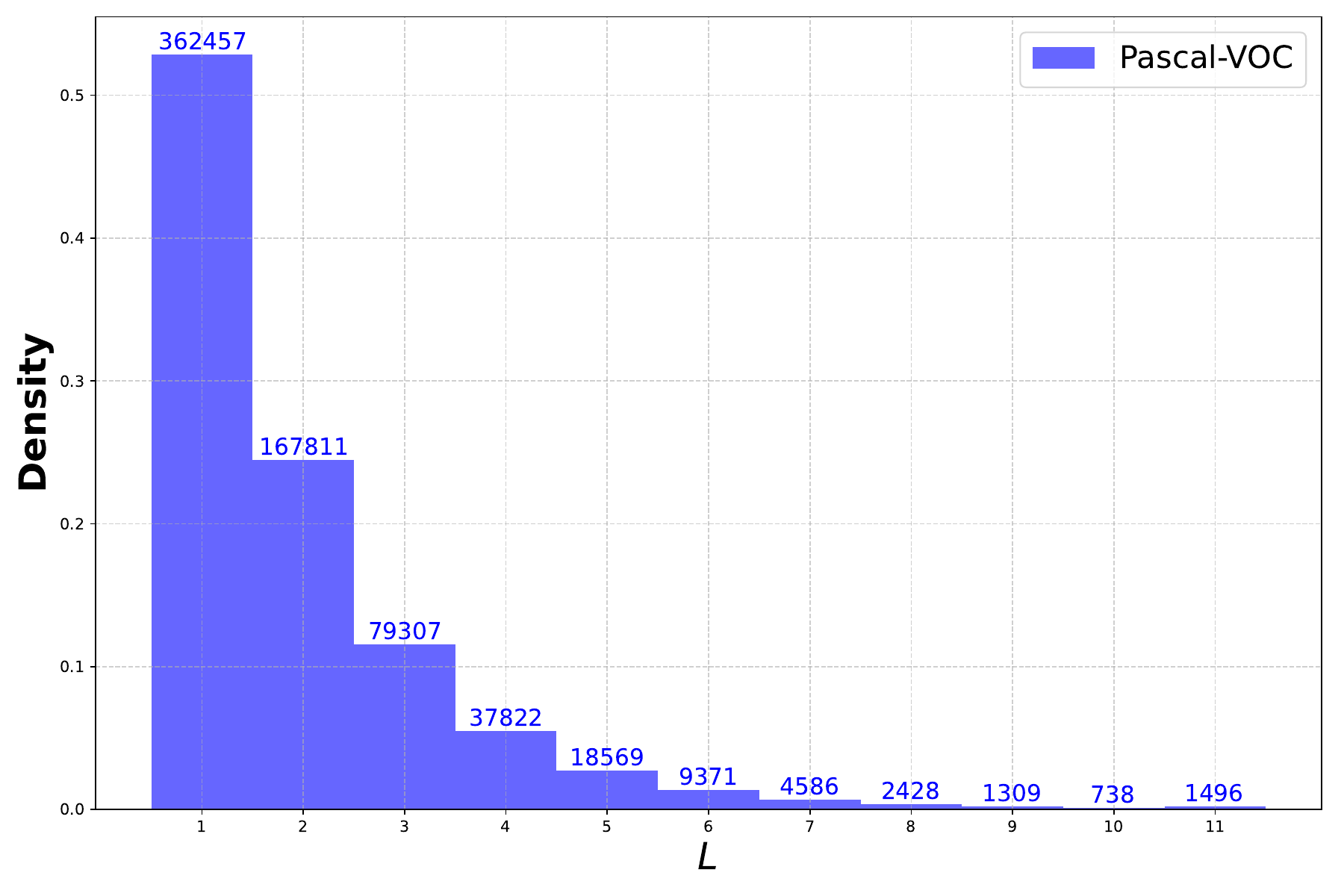}
    \caption{Node Exit per Layer discrete distribution in \texttt{Pascal-VOC}.}
    \label{fig:pascal_exit}
\end{figure}
\begin{figure}[t]
    \centering
    \begin{subfigure}{0.49\linewidth}
        \centering
        \includegraphics[width=1\textwidth]{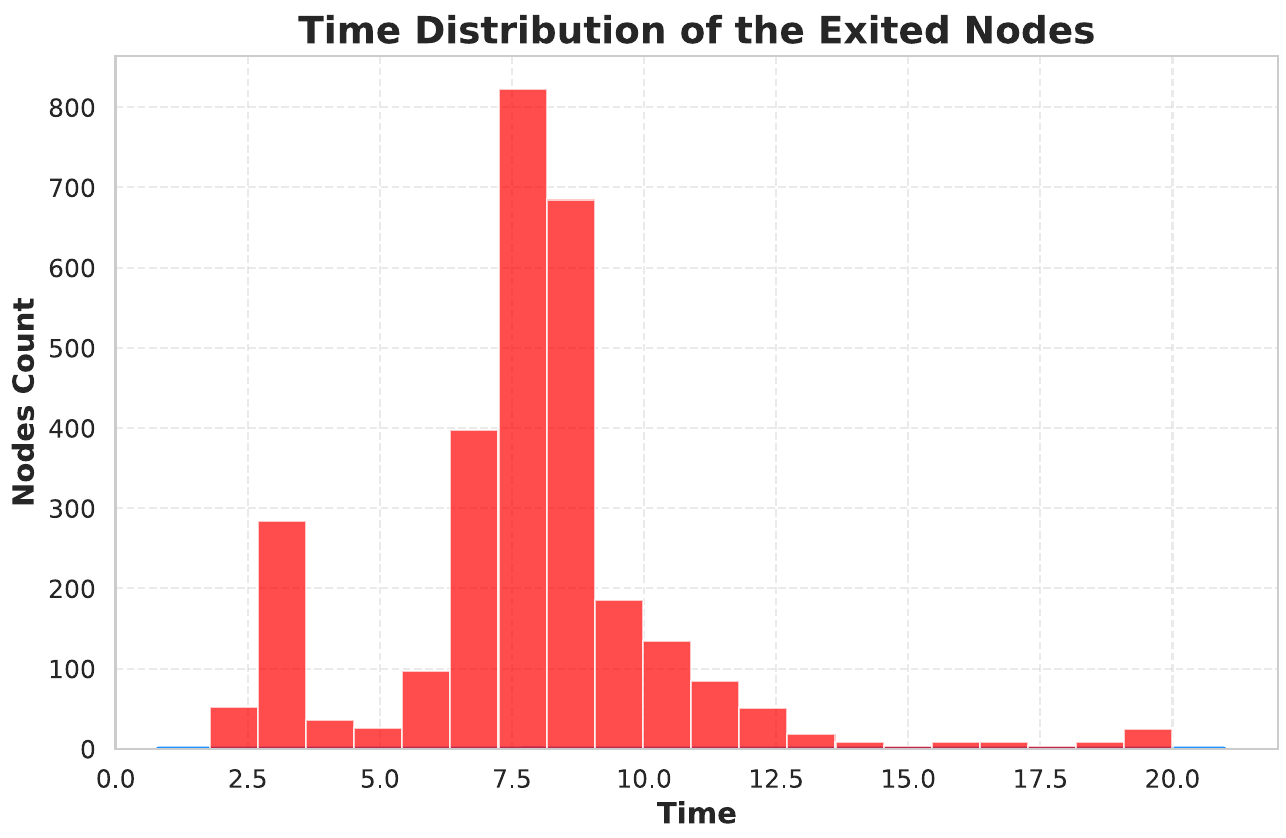}
        \caption{\vspace{4mm}}
        \label{fig:tolokers_continuo}
    \end{subfigure}
    \hfill
    \begin{subfigure}{0.49\linewidth}
        \centering
        \includegraphics[width=1\textwidth]{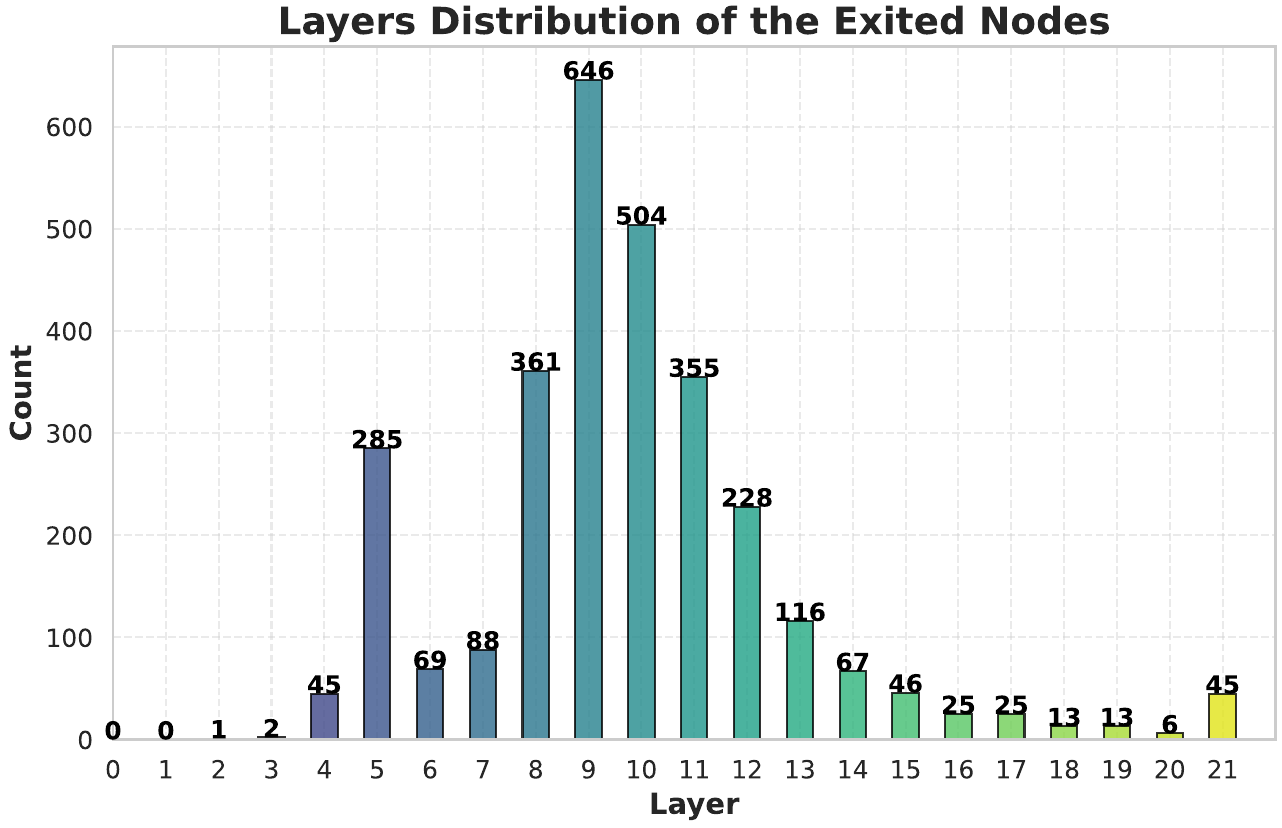}
        \caption{}
        \label{fig:tolokers_discreto}
    \end{subfigure}
    \caption{\texttt{Tolokers}: Exit point of the nodes in the test set. EEGNNs let the nodes choose their exit in the continuous domain (Left). We can visualize their exit in the discrete domain as well (Right).}
    \label{fig:node_distributions}
\end{figure}

\begin{figure}[t]
    \centering
    \begin{subfigure}[b]{0.46\textwidth}
        \centering
        \includegraphics[width=\textwidth]{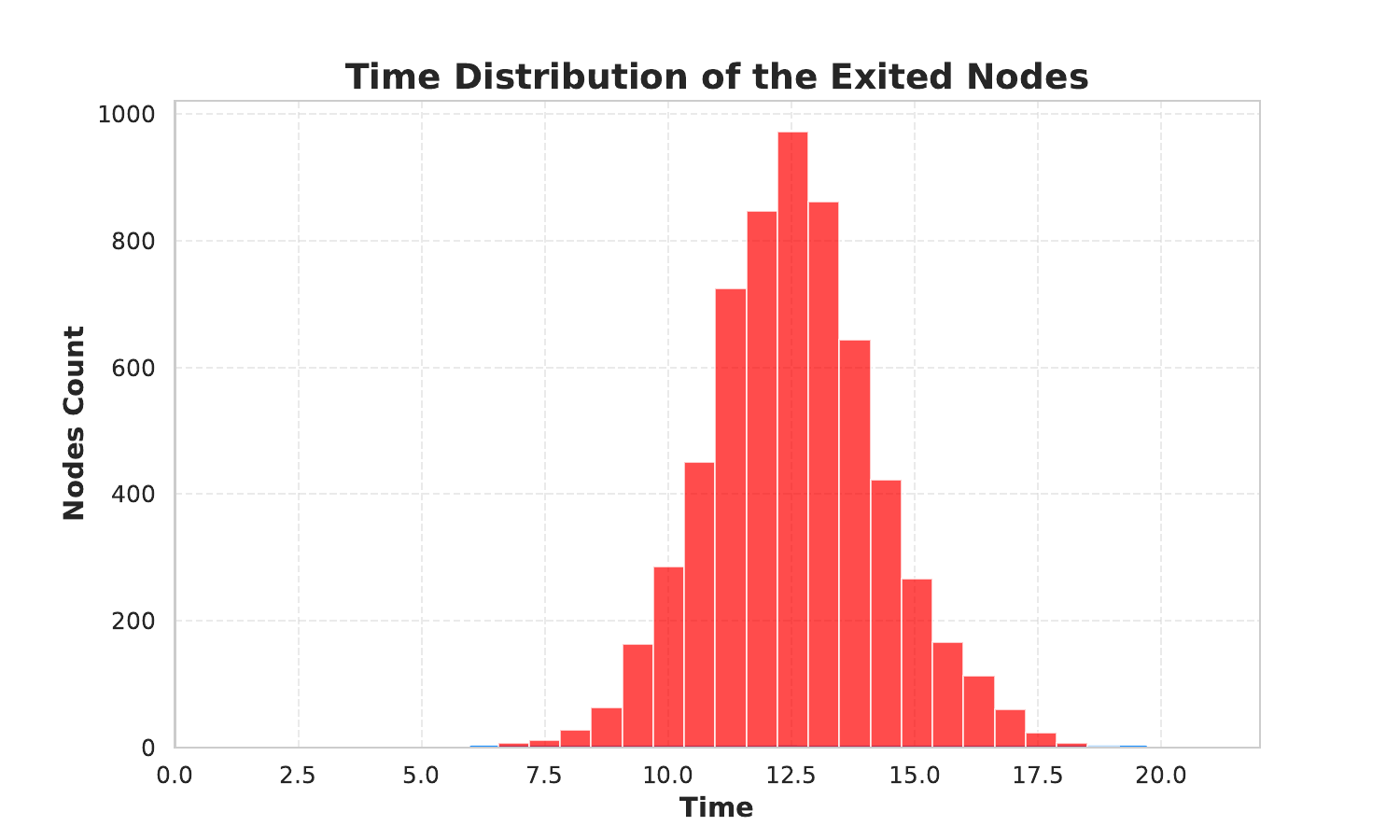}
        \caption{Node Exit in the time domain.}
        \label{fig:amazon_step_continuous}
    \end{subfigure}
    \hfill
    \begin{subfigure}[b]{0.46\textwidth}
        \centering
        \includegraphics[width=\textwidth]{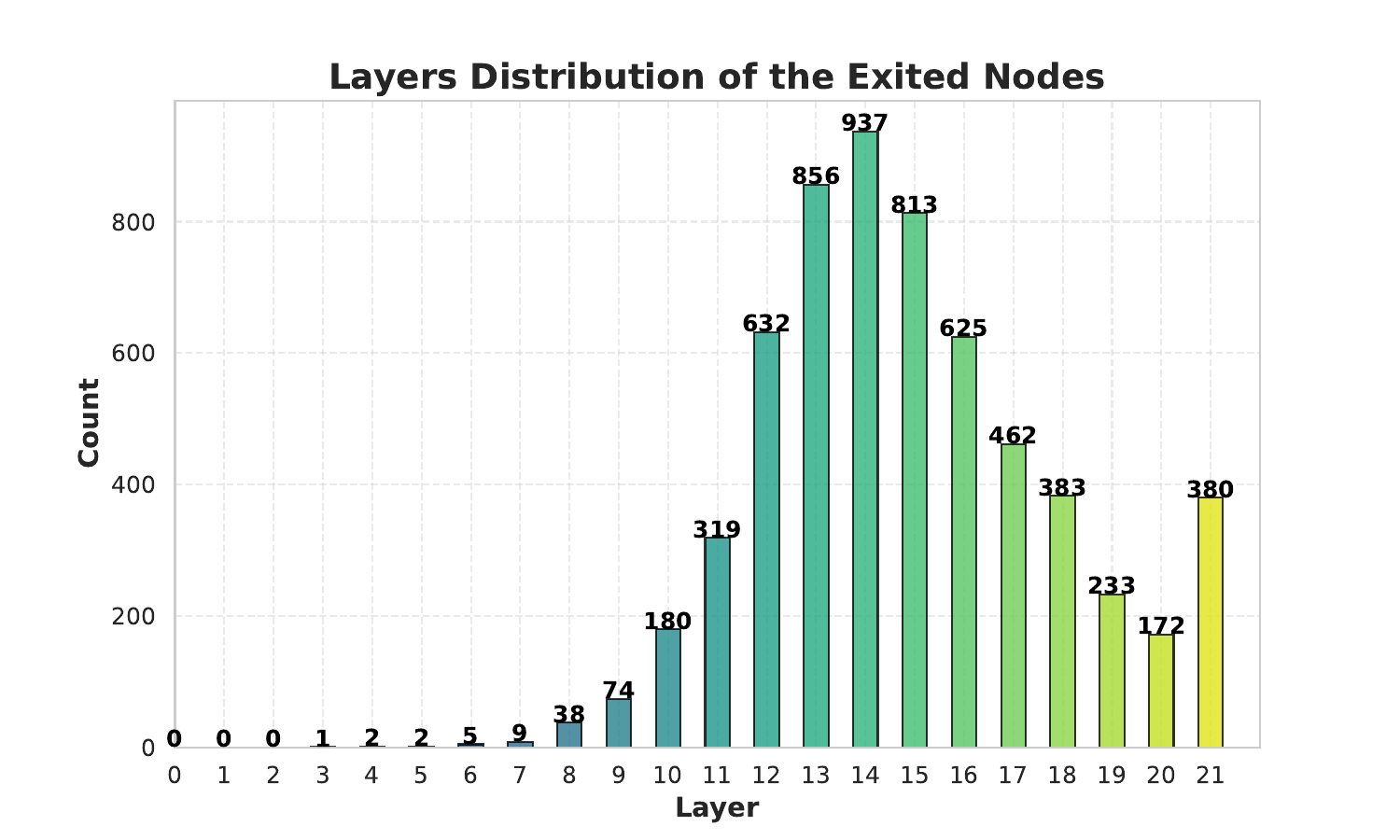}
        \caption{Node exits at each layer.}
        \label{fig:amazon_ratings_step_disccrete}
    \end{subfigure}
    \caption{\texttt{Amazon Ratings}: Exit point of the nodes in the test set. We have the discrete case (Right) and the continuous case (Left), thanks to our neural Adaptive-step mechanism.}
    \label{fig:amazon_node_distributions}
\end{figure}

\begin{figure}[t]
    \centering
    \begin{subfigure}[b]{0.46\textwidth}
        \centering
        \includegraphics[width=\textwidth]{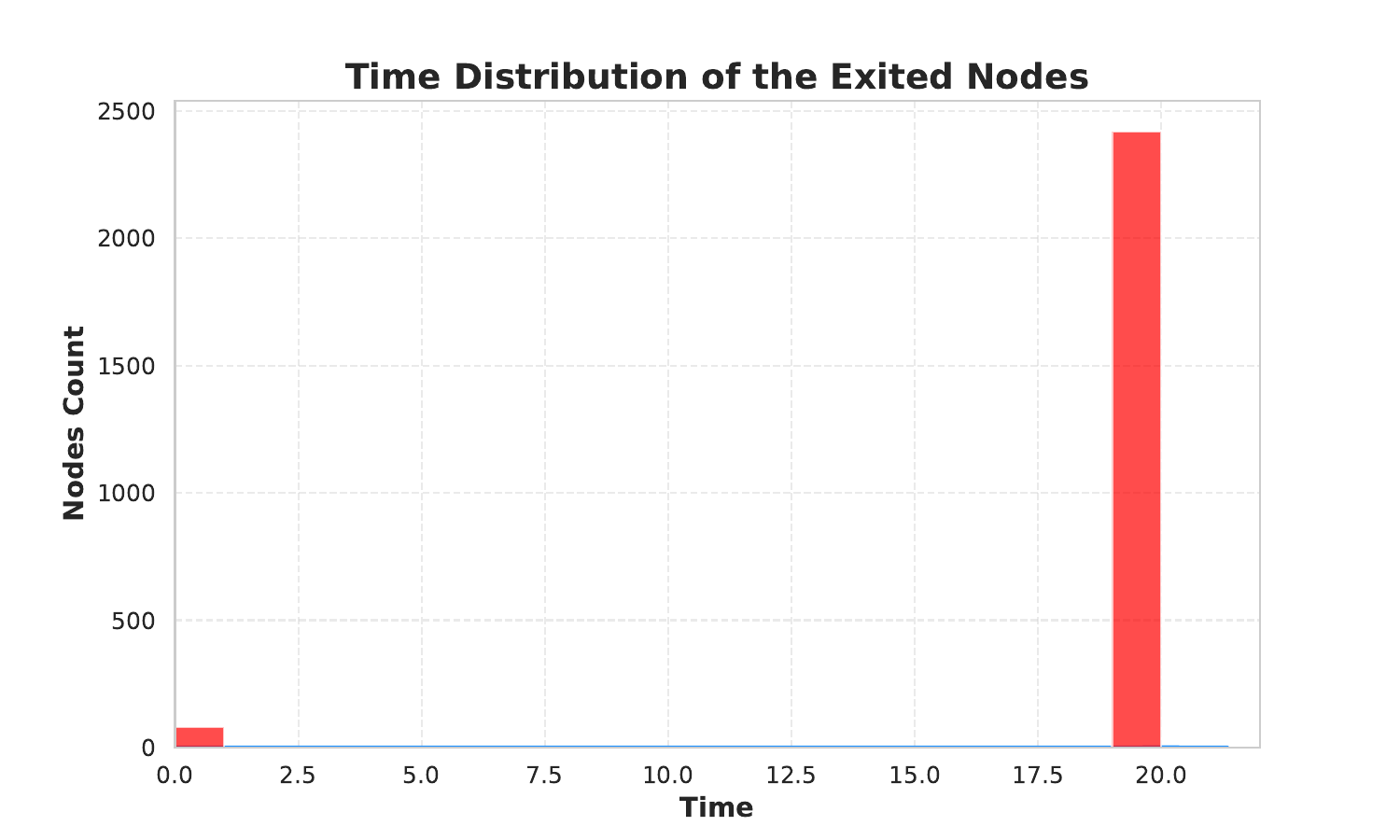}
        \caption{Node Exit in the time domain.}
        \label{fig:minesweeper_step_continuous}
    \end{subfigure}
    \hfill
    \begin{subfigure}[b]{0.46\textwidth}
        \centering
        \includegraphics[width=\textwidth]{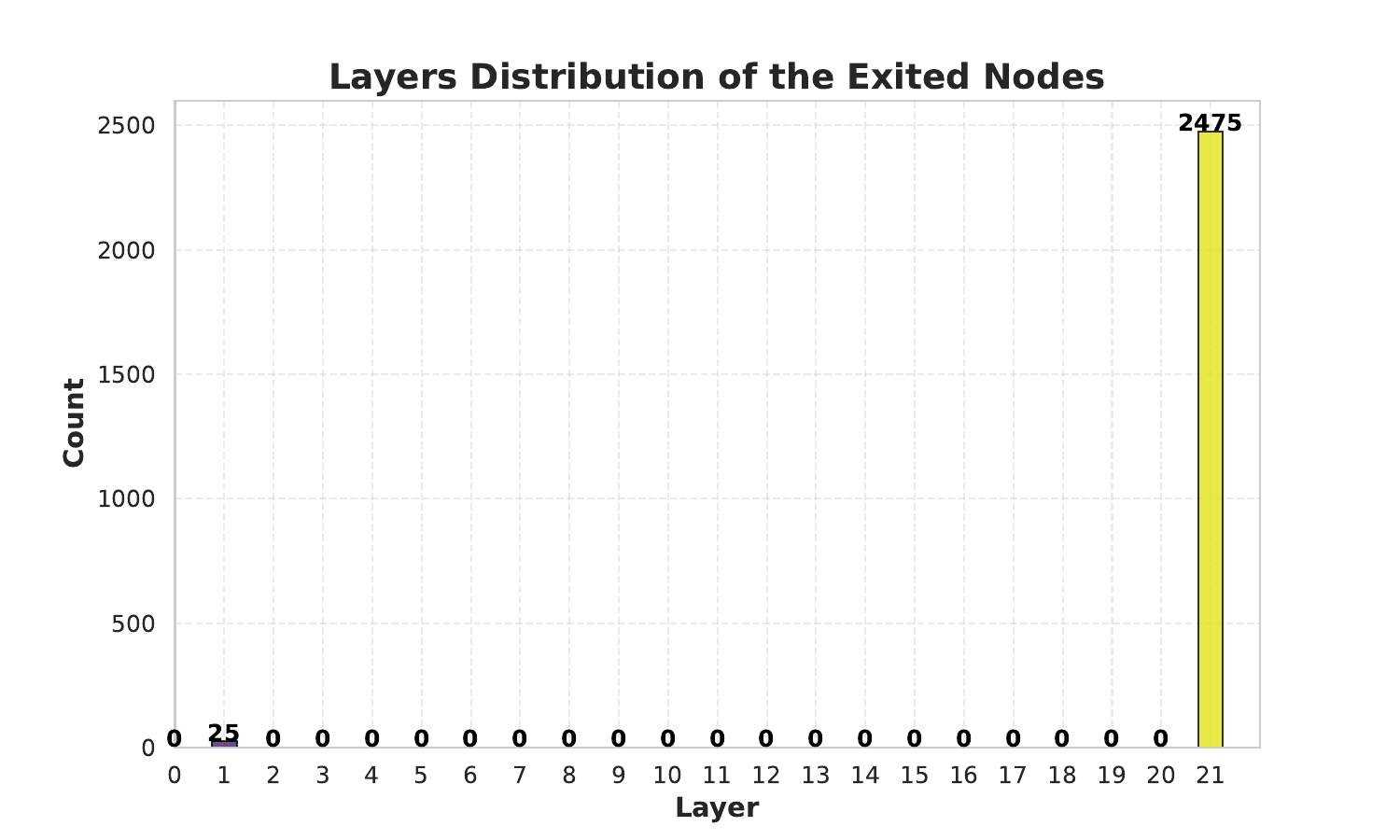}
        \caption{Node exits at each layer.}
        \label{fig:minesweeper_step_discrete}
    \end{subfigure}
    \caption{\texttt{Minesweeper}: Exit point of the nodes in the test set. We have the discrete case (Right) and the continuous case (Left), thanks to our neural Adaptive-step mechanism. At the first fold.}
    \label{fig:mines_node_distributions1}
\end{figure}
\begin{figure}[t]
    \centering
    \begin{subfigure}[b]{0.46\textwidth}
        \centering
        \includegraphics[width=\textwidth]{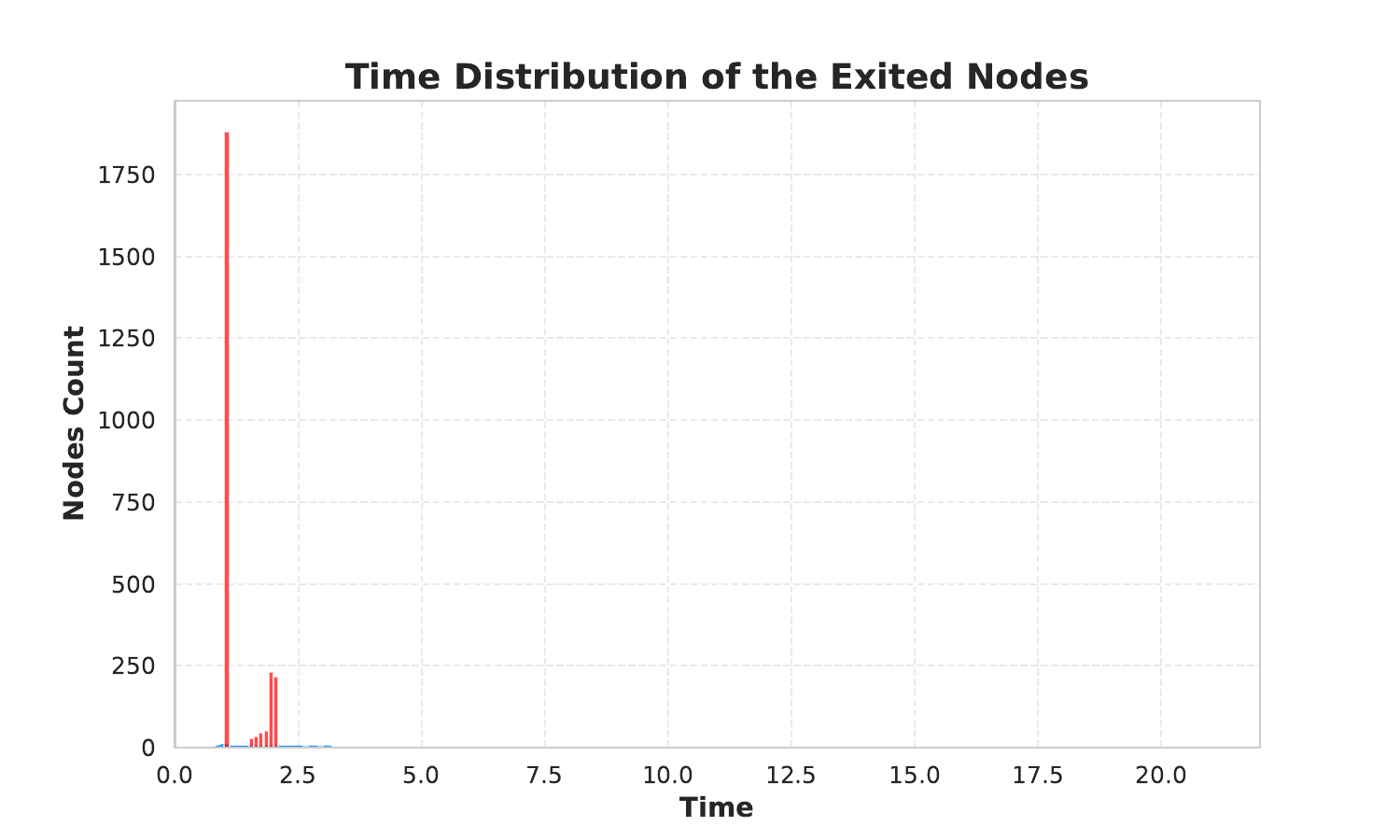}
        \caption{Node Exit in the time domain.}
        \label{fig:minesweeper_step_continuous2}
    \end{subfigure}
    \hfill
    \begin{subfigure}[b]{0.46\textwidth}
        \centering
        \includegraphics[width=\textwidth]{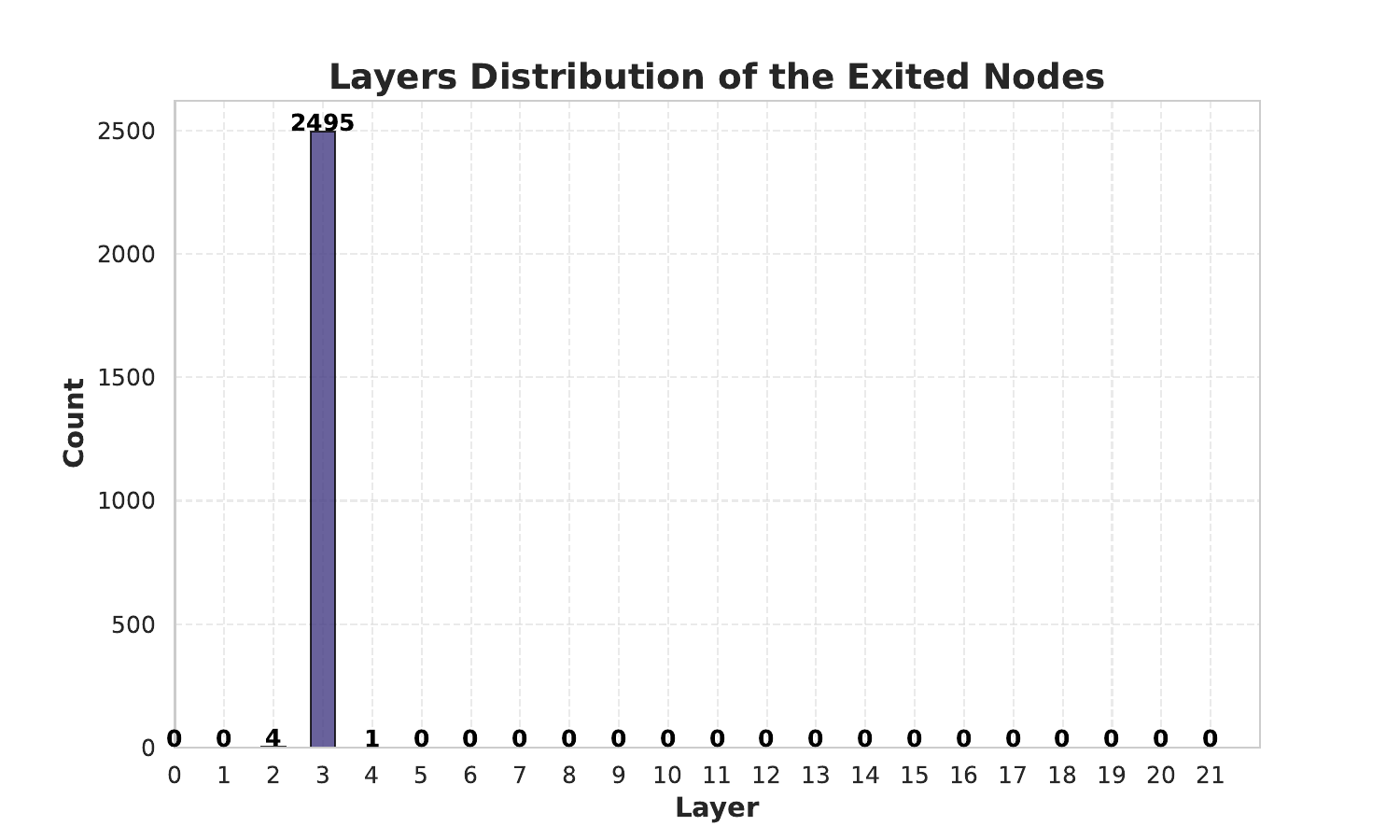}
        \caption{Node exits at each layer.}
        \label{fig:minesweeper_step_discrete2}
    \end{subfigure}
    \caption{\texttt{Minesweeper}: Exit point of the nodes in the test set. We have the discrete case (Right) and the continuous case (Left), thanks to our neural Adaptive-step mechanism. At the second fold.}
    \label{fig:mines_node_distributions2}
\end{figure}
\begin{figure}[t]
    \centering
    \begin{subfigure}[b]{0.46\textwidth}
        \centering
        \includegraphics[width=\textwidth]{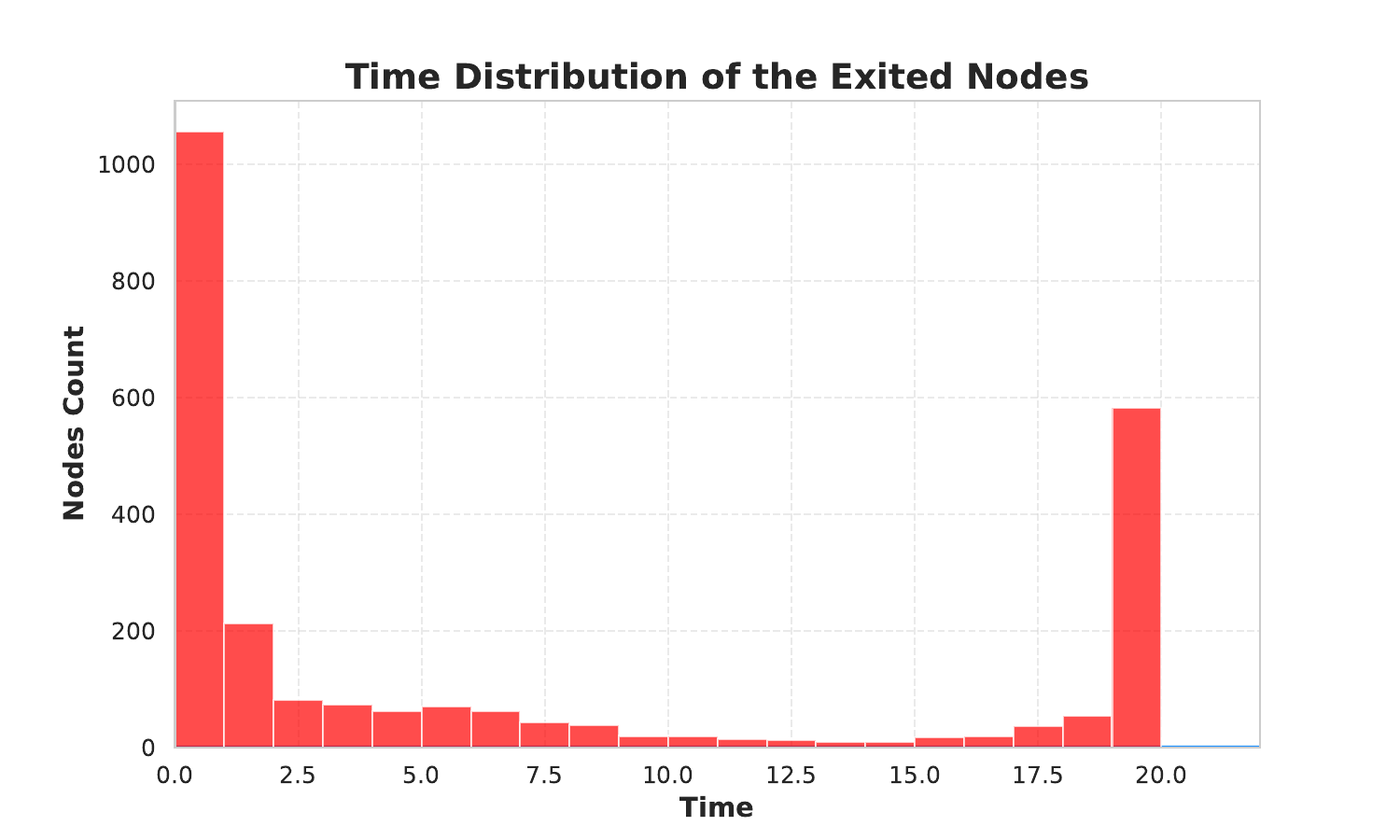}
        \caption{Node Exit in the time domain.}
        \label{fig:minesweeper_step_continuous3}
    \end{subfigure}
    \hfill
    \begin{subfigure}[b]{0.46\textwidth}
        \centering
        \includegraphics[width=\textwidth]{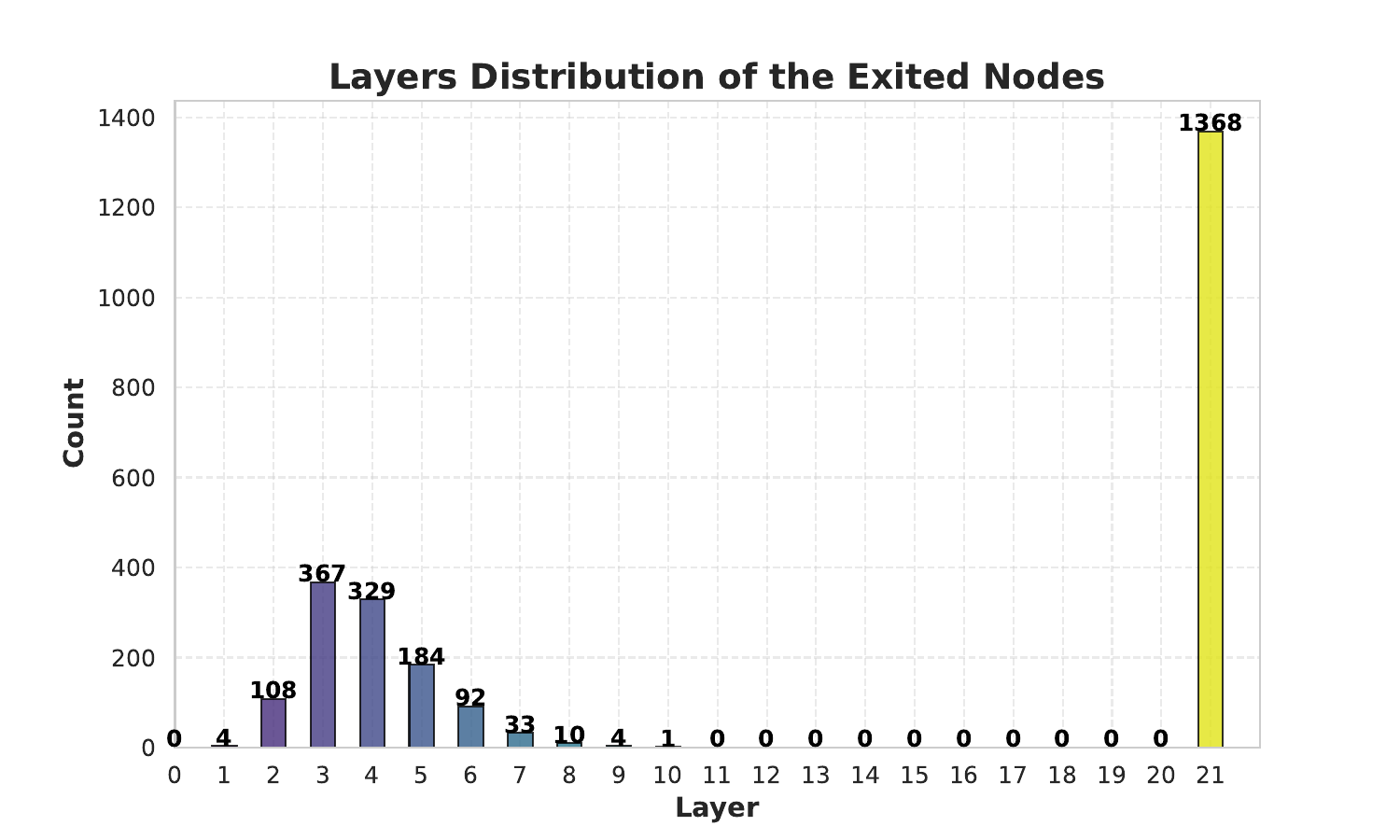}
        \caption{Node exits at each layer.}
        \label{fig:minesweeper_step_discrete3}
    \end{subfigure}
    \caption{\texttt{Minesweeper}: Exit point of the nodes in the test set. We have the discrete case (Right), and the continuous case (Left), thanks to our neural Adaptive-step mechanism. At the fifth fold.}
    \label{fig:mines_node_distributions3}
\end{figure}
\begin{figure}[t]
    \centering
    \begin{subfigure}[b]{0.46\textwidth}
        \centering
        \includegraphics[width=\textwidth]{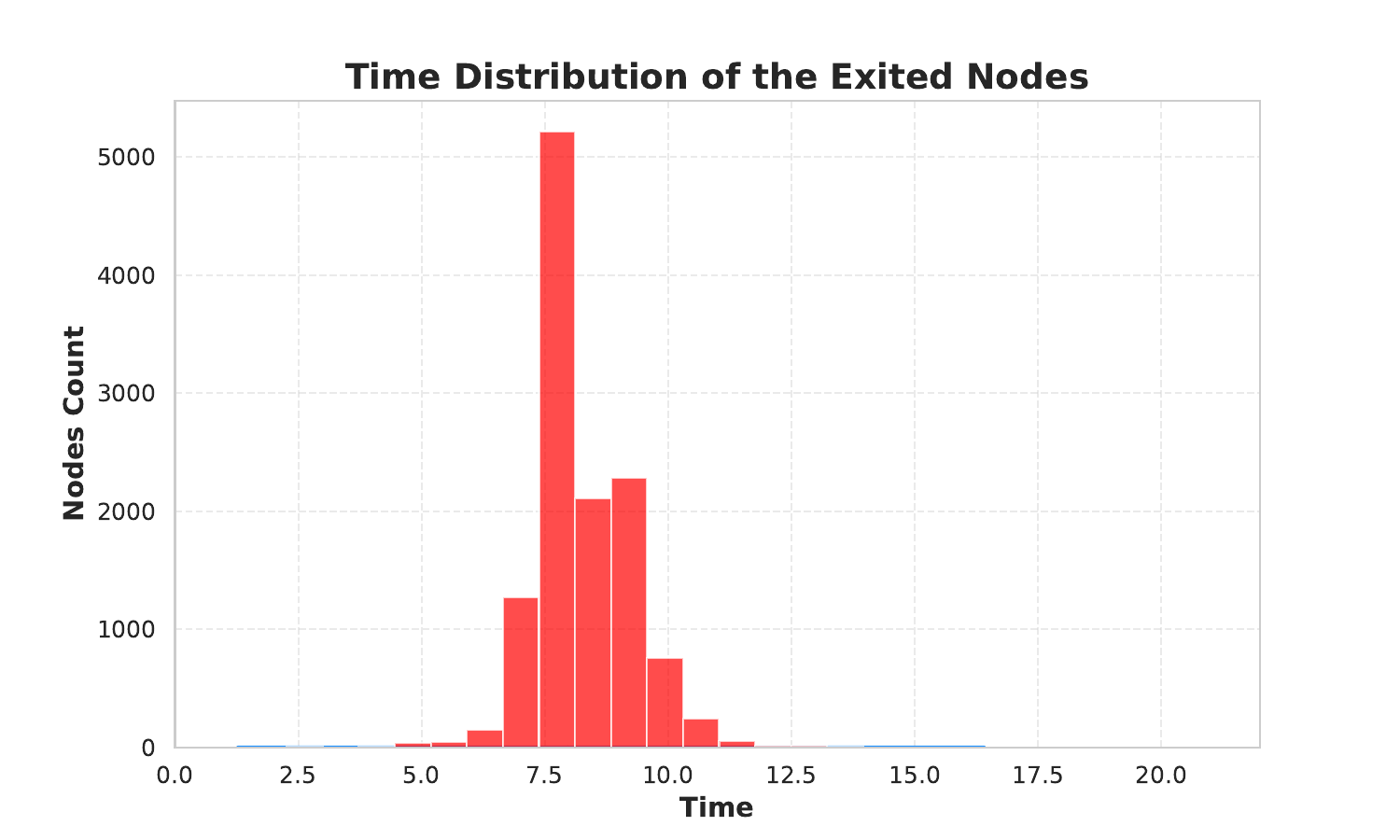}
        \caption{Node Exit in the time domain.}
        \label{fig:questions_step_continuous}
    \end{subfigure}
    \hfill
    \begin{subfigure}[b]{0.46\textwidth}
        \centering
        \includegraphics[width=\textwidth]{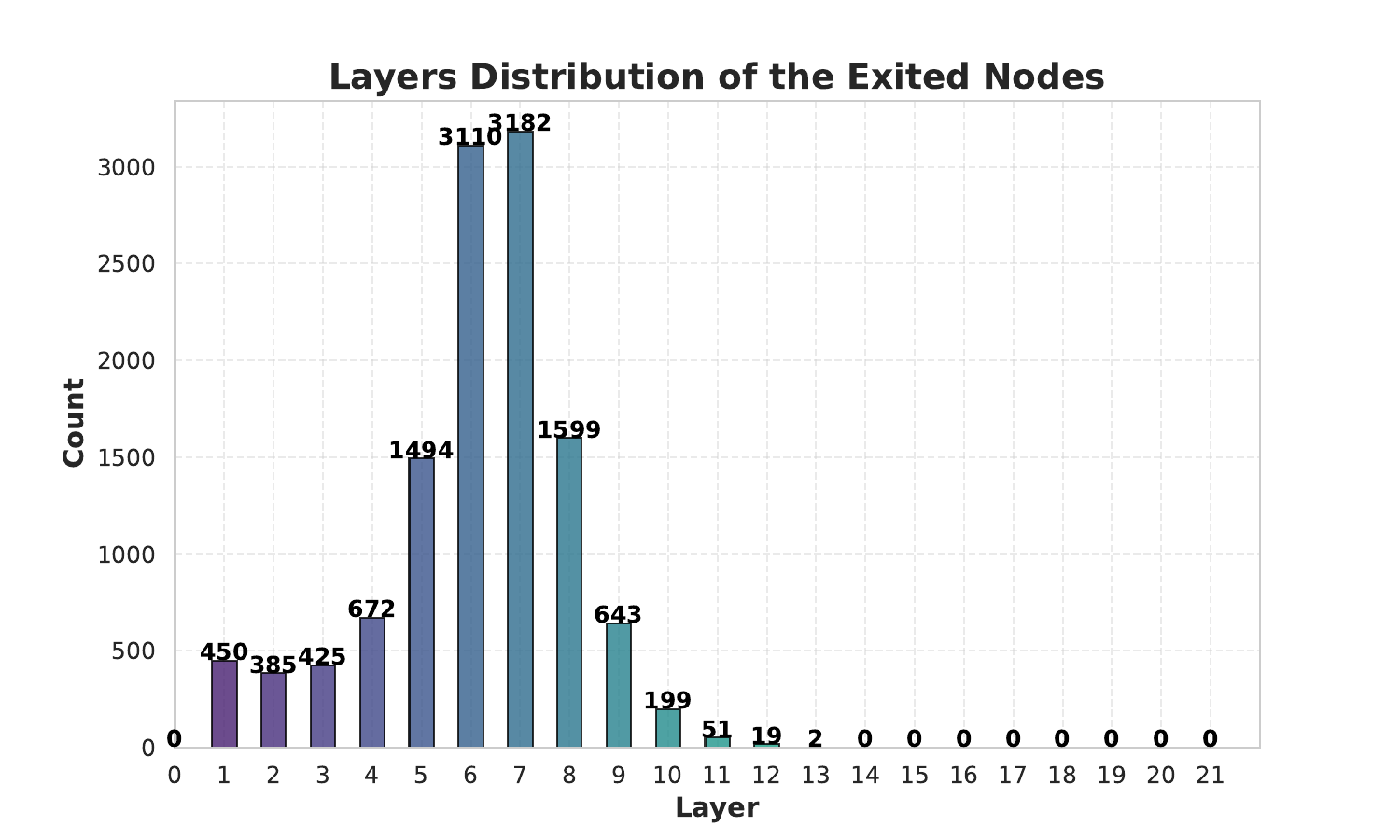}
        \caption{Node exits at each layer.}
        \label{fig:questions_step_discrete}
    \end{subfigure}
    \caption{\texttt{Questions}: Exit point of the nodes in the test set. We have the discrete case (Right) and the continuous case (Left), thanks to our neural Adaptive-step mechanism.}
    \label{fig:questions_node_distributions}
\end{figure}

\begin{figure}[b]
    \centering
    \begin{subfigure}{0.46\textwidth}
        \centering
        \includegraphics[width=\textwidth]{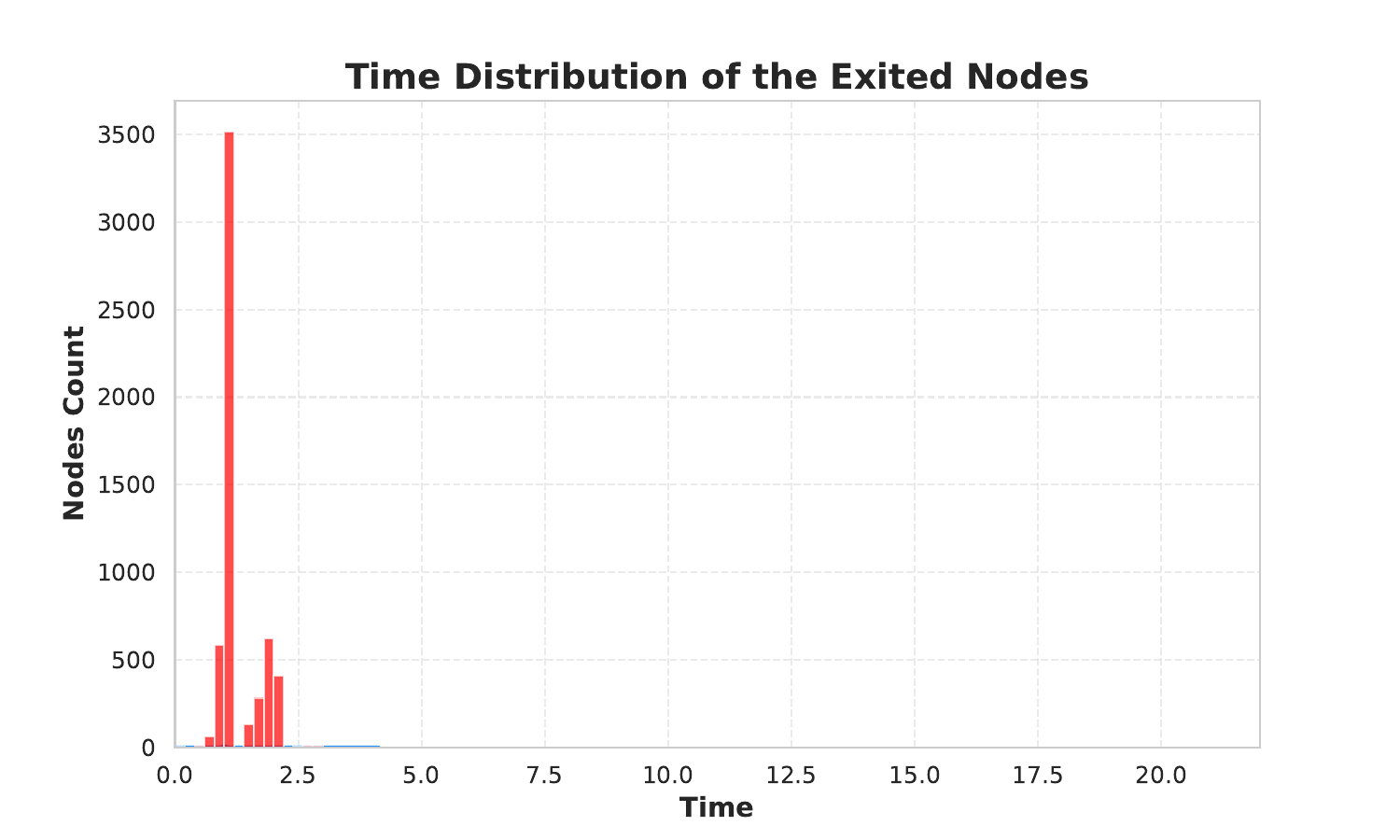}
        \caption{Node Exit in the time domain.}
        \label{fig:roman_empire_step_continuous}
    \end{subfigure}
    \hfill
    \begin{subfigure}{0.46\textwidth}
        \centering
        \includegraphics[width=\textwidth]{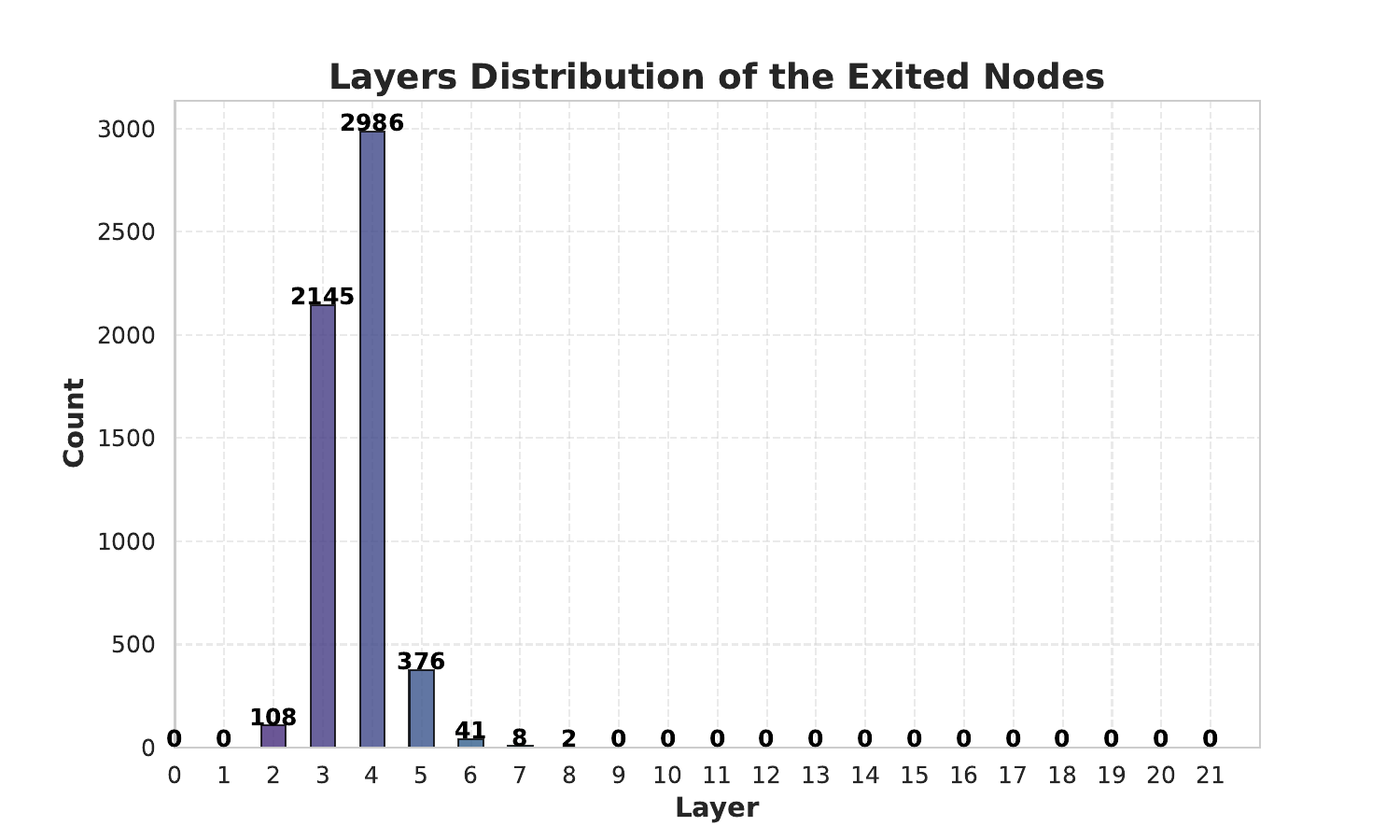}
        \caption{Node exits at each layer.}
        \label{fig:roman_empire_step_discrete}
    \end{subfigure}
    \caption{\texttt{Roman Empire}: Exit point of the nodes in the test set. We have the discrete case (Right) and the continuous case (Left), thanks to our neural Adaptive-step mechanism.}
    \label{fig:roman_empire_node_distributions}
\end{figure}
\subsection{Additional Examples of Exit Distributions}
\label{sec:node_distributions}

In the main paper, we presented an example of graph exit distributions on the test set of \texttt{Peptides-Func} and \texttt{Peptides-struct}. Here, we extend this analysis to \texttt{Tolokers}, \texttt{Amazon Ratings}, \texttt{Minesweeper}, \texttt{Roman Empire}, and \texttt{Questions}. In Table~\ref{tab:node_distributions}, we compare the fixed depth chosen in SAS-GNN with discrete statistics of the learned exit distributions (minimum, maximum, median), averaged across splits. For EEGNN, we set a budget of $L=20$ layers, which we found sufficient with respect to the median of most distributions. Interestingly, there are datasets where the median exit depth exceeds the fixed baseline (e.g., \texttt{Amazon Ratings}, see Figure~\ref{fig:amazon_node_distributions}), as well as cases where nodes tend to exit much earlier than the fixed number of layers (e.g., \texttt{Roman Empire} and \texttt{Questions}, see Figures~\ref{fig:roman_empire_node_distributions} and \ref{fig:questions_node_distributions}). For \texttt{Tolokers}, Figure~\ref{fig:node_distributions} shows a clear contrast: many nodes exit before $l=10$, while others go significantly deeper, whereas SAS-GNN enforces a rigid $L=10$, constraining the model to a fixed behavior.  
Figure~\ref{fig:pascal_exit} illustrates the case of \texttt{Pascal-VOC}. These results also highlight variability across splits. The \texttt{Minesweeper} dataset, in particular, exhibits a high standard deviation in exit statistics. Examining the distribution plots clarifies this phenomenon: in some splits, all nodes exit very early (Figure~\ref{fig:mines_node_distributions2}); in others, nearly all nodes use the full budget (Figure~\ref{fig:mines_node_distributions1}); and in yet others, exits are smoothly distributed (Figure~\ref{fig:mines_node_distributions3}). Thanks to EEGNN, such diverse behaviors can be expressed naturally, since exits are determined adaptively at the node level. Future works can also build upon this capabilities to enable more personalized and adaptive behavior when the confidence network decides whether to exit or not. 

\begin{figure}[t]
    \centering
    \includegraphics[width=0.48\linewidth]{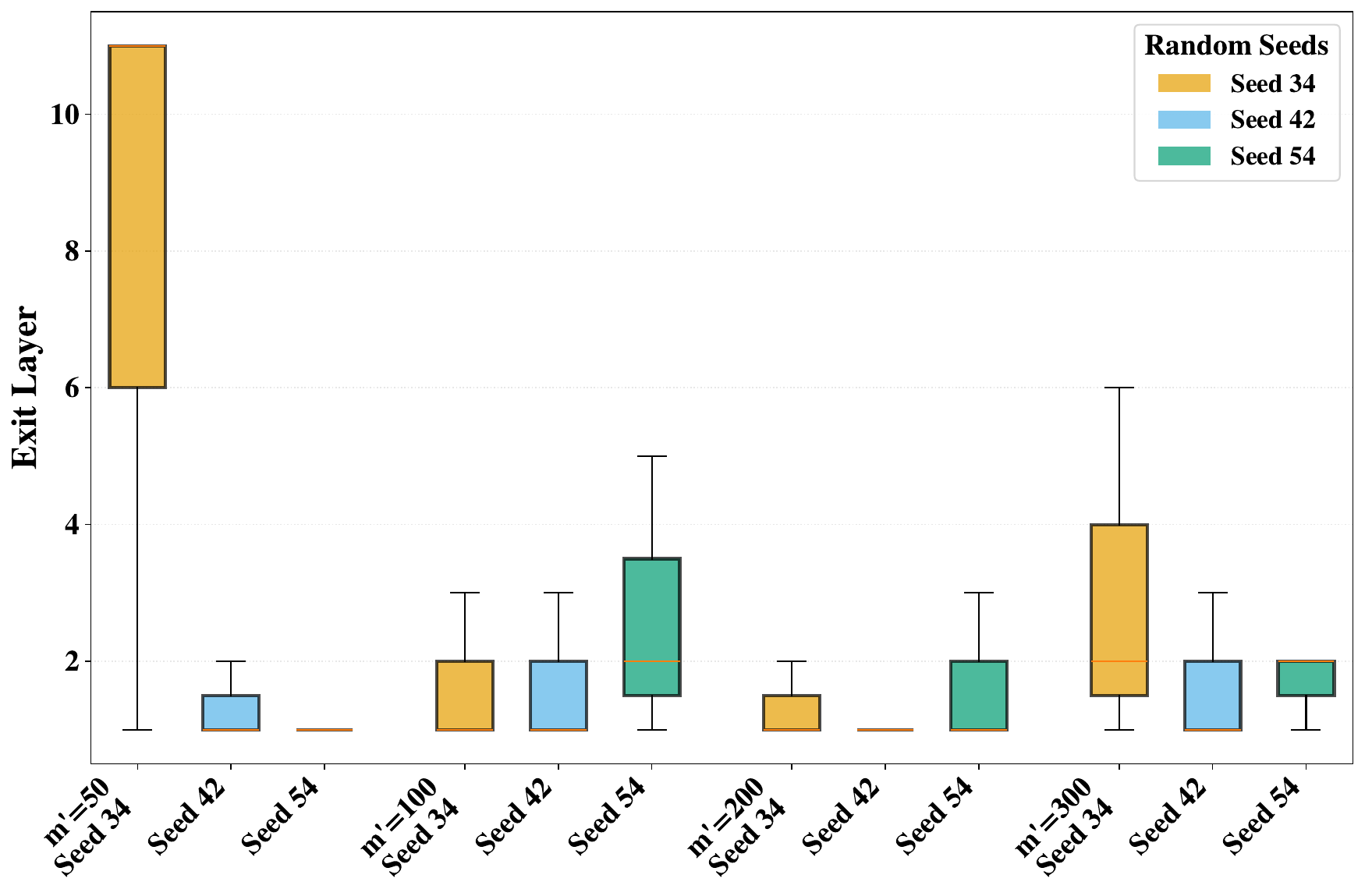}
    \hfill
    \includegraphics[width=0.48\linewidth]{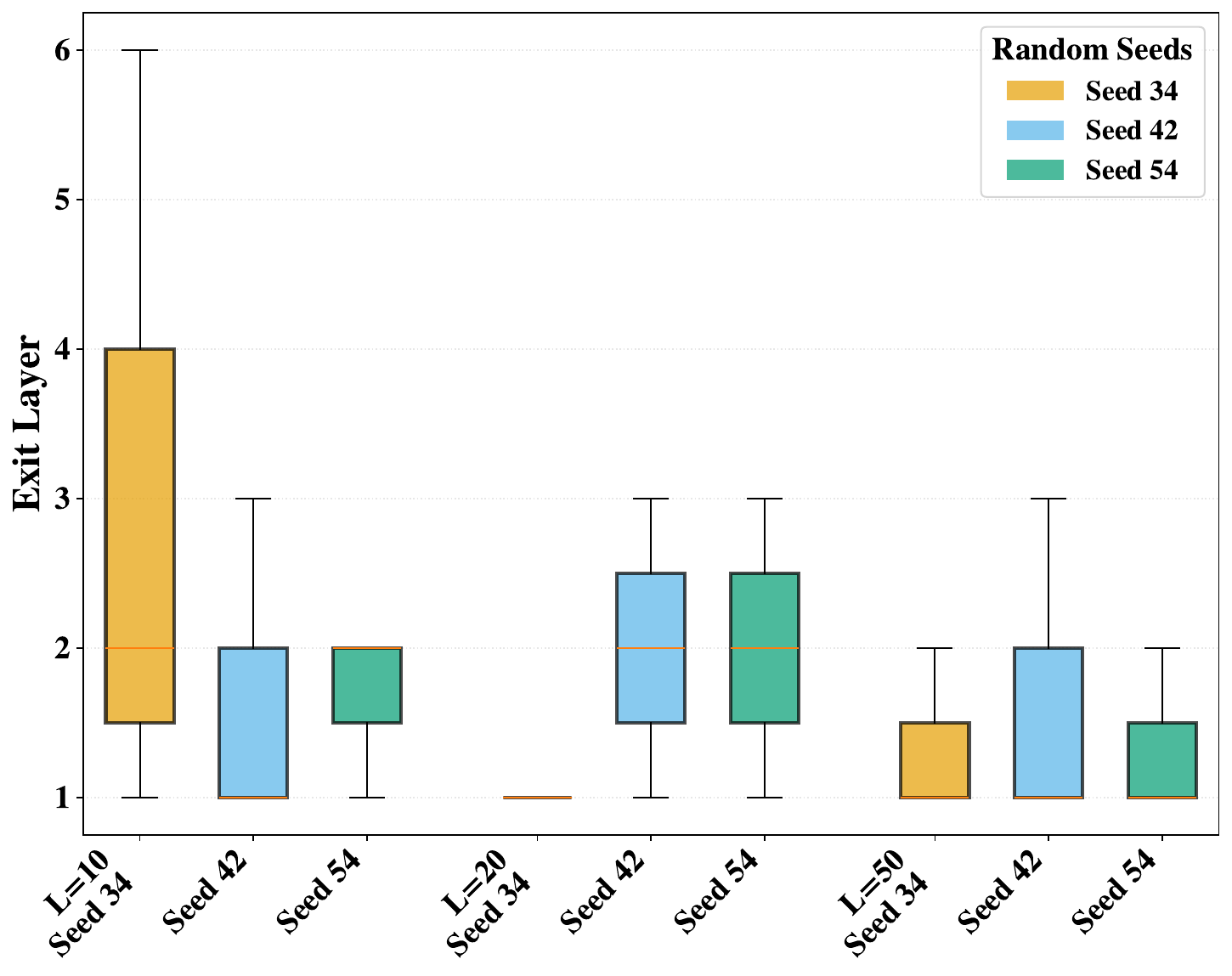}
    \caption{Exit distributions of EEGNN on \texttt{Peptides-struct}. \textbf{Left:} varying hidden dimensions ($m'$) across random seeds. \textbf{Right:} varying depth budgets ($L$) across random seeds.}
    \label{fig:peptides_exit_analysis}
\end{figure}

\paragraph{Robustness to Training Conditions.}  
The analyses above focused on split-level variability. To further examine robustness, we conducted additional experiments on the \texttt{Peptides-struct} dataset from the LRGB benchmark, where results are averaged by varying random seeds. Here we assess the exit distributions according to these, the hidden dimensions $m'$, and the depth budget $L$. Exit distributions under these settings are reported in Figure~\ref{fig:peptides_exit_analysis}, while Table~\ref{tab:peptides_exit} summarizes the quantitative results.  

\begin{table}[ht]
\centering
\caption{Robustness analysis of EEGNN on \texttt{Peptides-struct} under varying hidden dimensions and depth budgets. Reported values are accuracy (mean $\pm$ std) and parameter counts. Exit distributions are shown in the figures.}
\label{tab:peptides_exit}
\begin{tabular}{lcc}
\toprule
Setting & Accuracy $\pm$ Std & Parameters \\
\midrule
$\{m'=50$, $L=10\}$   & 0.2783 $\pm$ 0.0120 & 21,599 \\
$\{m'=100$, $L=10\}$  & 0.2668 $\pm$ 0.0065 & 63,649 \\
$\{m'=200$, $L=10\}$  & 0.2607 $\pm$ 0.0049 & 207,749 \\
$\{m'=300$, $L=10\}$  & 0.2532 $\pm$ 0.0050 & 431,849 \\
\midrule
$\{m'=300$, $L=10\}$  & 0.2532 $\pm$ 0.0050 & 431,849 \\
$\{m'=300$, $L=20\}$  & 0.2548 $\pm$ 0.0057 & 431,849 \\
$\{m'=300$, $L=50\}$  & 0.2572 $\pm$ 0.0066 & 431,849 \\
\bottomrule
\end{tabular}
\end{table}

From these results, two consistent trends emerge:  
(i) increasing the hidden dimension improves performance while leaving exit behavior largely unaffected—most graphs exit after only 1–2 layers regardless of width, with deeper exits occurring only in lower-performance regimes; and  
(ii) increasing the depth budget $L$ up to 50 layers does not shift the distributions, with the model consistently exiting early. Importantly, these exit patterns remain robust across random seeds, demonstrating that EEGNN is stable not only to dataset splits but also to training stochasticity. In this particular case, \texttt{Peptides-struct} is known to be a short-range dataset \citep{measuringlongrange}, so the insensitivity to $L$ becomes appropriate, even though our model is not designed to be aware of the problem radius.

\section{Extended Related Work}
\label{sec:extended_related_works}

This section reviews the literature most relevant to our work, encompassing Graph Neural Networks (GNNs), advanced message-passing strategies, early-exit mechanisms for GNNs, and early-exit techniques in general neural architectures.\\

\textbf{Graph Neural Networks (GNNs). }Most GNNs follow the message-passing paradigm~\citep{messagepassing}, where node features are iteratively updated by aggregating information from neighboring nodes via convolutions, attention, or neural networks~\citep{everythingisconnected}. These Message-Passing Neural Networks (MPNNs), including GCN~\citep{GCN}, GraphSAGE~\citep{GraphSAGE}, and GAT~\citep{GAT}, achieve strong empirical performance but are inherently limited by the number of layers: a $k$-layer MPNN can only propagate information along paths of length at most $k$. Deep MPNNs often suffer from \emph{over-smoothing} (OST: node features become indistinguishable)~\citep{oversmoothing, oversmoothing2}, \emph{over-squashing} (OSQ: long-range information compressed through narrow bottlenecks)~\citep{oversquashing1}, and \emph{under-reaching} (insufficient depth to capture necessary dependencies)~\citep{underreaching}. 
Other limitations stem from restricted expressivity, as classical MPNNs are at most as powerful as the 1-WL (Weisfeiler-Lehman) test~\citep{WL1}.  

\textbf{Enhancing expressivity. }Several strategies aim to address these issues. Higher-order message passing~\citep{TDL} improves substructure expressivity beyond the 1-WL test. Graph Transformers (GTs)~\citep{GT1} leverage self-attention to allow unrestricted node interactions, avoiding reliance on depth to capture long-range dependencies. However, GTs generally scale quadratically with the number of nodes, limiting their scalability, and \citet{MPNN_is_transformer} shows that they are not inherently more expressive than MPNNs equipped with a virtual node. Moreover, in practice, GTs do not consistently outperform classical MPNNs such as GCN on benchmarks like LRGB~\citep{MPNN_good_lrgb}, casting doubts on their suitability for long-range tasks. Both GTs and MPNNs also typically rely on normalization and dropout~\citep{classicgnnsstrongbaselines}, which can obscure model dynamics.  


\textbf{Neural ODE-inspired GNNs. }GraphNODEs~\citep{GraphNODE} are a class of GNNs that model node features propagation through a continuous-time dynamical system defined by a differential equation. This paradigm offers a way of interpreting GNNs as discretisations of ODEs and PDEs and provides a principled framework for model design, as the differential equation formulation enables an easier theoretical analysis of stability and expressiveness of neural networks.
Among these we have examples of GNNs which simulate the heat equation diffusion process \cite{GRAND, grand++}. However, even if these models present the opportunity for a more principled-then-interpretable design, MPNNs are prone to common learning flaws, as they are typically encompassed in the message-passing paradigm. In order to overcome this, the design of GraphNODE have been biased towards the mitigation of well known problems such as over-smoothing (OST) or over-squashing (OSQ).
For OST we have examples of GNNs interpreted as gradient flaw, as GRAFF \cite{GRAFF2}, here node message-passing simulate a dynamics where nodes induce attraction-repulsion behavior edge-wise. In \cite{graphcon}, an example of second-order differential equation for GraphNODE, oscillator equations are used to model the way messages are exchanged across adjacent nodes. We also have another class of attraction-repulsion mechanism, based on allen-cahn dynamics \cite{wang2025acmpallencahnmessagepassing}, or Graph Neural Diffusion-Reaction equation \cite{GREAD}. More recently, as the phenomenon of over-squashing (OSQ) has attracted increasing attention, GraphNODE-based approaches have also been explored in this context. From the GraphNODE point of view, the problem of OSQ has been studied as a way of increasing the long-range capabilities of these networks, which unfortunately leave unsolved the more recent problem of short-range over-quashing \cite{mishayev2025shortrangeoversquashing, blayney2025glstmmitigatingoversquashingincreasing}. However, in terms of long-range over-squashing, the use of physics-inspired architectures have seen its development encoding the stability and non-dissipativity properties both locally as well as globally within the message-passing~\cite{ADGN, swan}. Similarly, these properties can be preserved through the use of GNNs behaving as port-Hamiltonian ~\cite{porthamiltonian_gnn}. Second-order differential equations found place also for GNNs who preserves long-range propagation~\cite{trenta2025sonar}.

In this work, we draw inspiration from prior approaches addressing both OST and OSQ, and we propose EEGNN, the first GraphNODE who natively supports an early-exit mechanism.
Recently, have been shown that over-smoothing and over-squashing are related to the vanishing gradient phenomenon, and a state-space modelling have been proposed to mitigate both \cite{vanishinggradientsoversmoothingoversquashing}.

Despite these advances, most approaches either increase architectural complexity or incur scalability issues. Our work departs from this by enabling dynamic, per-node (or per-graph) depth selection without altering the topology. Unlike many methods, we do not consider the expressivity limitations of the 1-WL test, leaving this for future work.  

\textbf{Asynchronous Message Passing. }Asynchronous MPNNs~\citep{coognns, AdaptiveMP, gwac} dynamically alter graph connectivity at each layer to improve long-range communication. For instance, Co-GNN~\citep{coognns} introduces node-level interaction modes (e.g., isolate, broadcast, listen) to address OST and OSQ. Inspired by this, we allow discrete node decisions—but instead of modifying connectivity, nodes decide whether to \emph{exit} or \emph{continue}, preserving topology and ensuring efficiency. Unlike Co-GNNs, which require manual depth tuning and higher inference cost, our approach is lightweight and scalable. Similarly, AMP~\citep{AdaptiveMP} learns depth and a message-filtering strategy, but applies a fixed depth at inference. In contrast, we adapt depth both at training and testing time while using a fixed topology.  

Overall, compared to asynchronous message-passing, our synchronous design is more efficient in time and space complexity while still mitigating OST and OSQ. 

\textit{Note on Short-Range Over-squashing:} In this work, we focus on long-range information propagation. We do not explicitly address "computational over-squashing" in the short-range scenario~\citep{mishayev2025shortrangeoversquashing}, a recently observed phenomenon where capacity limits hinder local processing. As shown by \citet{blayney2025glstmmitigatingoversquashingincreasing}, this can be mitigated by increasing model width/capacity, which is complementary to our depth-adaptive approach. Generally graph transformers are a better fit than MPNNs in these settings. 

\textbf{Early-Exit Mechanisms in GNNs. }Few works have explored early-exit in GNNs. \citet{AdaProp} introduced an exit mechanism allowing nodes to stop updating during message passing. However, their method relies on auxiliary loss terms, is limited to addressing OST, and does not handle OSQ or deep architectures. Other works~\citep{learningtopropagatemess, turningacurse, ADMP-GNN} also propose exit mechanisms, but remain restricted to node classification and do not analyze OST/OSQ theoretically. In contrast, we design an early-exit mechanism that is fully differentiable, trained end-to-end with only the task loss, and applicable to both node-level and graph-level tasks. Nodes remain active as long as necessary for feature extraction, addressing deep MPNN flaws by design.  

\textbf{Early-Exit in Neural Networks (General). }Early-exit has been widely explored in standard deep learning to reduce inference cost by attaching auxiliary classifiers to intermediate layers. Training paradigms include deeply supervised nets~\citep{deeplysupervisednets}, layer-wise pretraining~\citep{layer-wise}, and independent exit training~\citep{separatedtraining}. Inference termination typically depends on confidence thresholds~\citep{overthinking, deebert, confidence_method, berxit} or threshold-free approaches~\citep{threshold_agnosticEE, contextual_bandits}. These works inspire our threshold-free, jointly trainable exit mechanism, tailored to graph-structured data and obviating the need for hand-tuned thresholds.

\end{document}